\documentclass{article}

\usepackage{amsmath}
\usepackage{amssymb}
\usepackage{mathtools}
\usepackage{amsthm}
\usepackage{listings} 
\usepackage{xspace}
\usepackage{multirow}
\usepackage{subcaption}
\usepackage[dvipsnames]{xcolor}
\usepackage{booktabs}
\usepackage{wrapfig}   
\newcommand{\RR}{\mathbb{R}} 
 
\newcommand{\CC}{\mathbb{C}} 
\newcommand{\bC}{\mathbf{C}} 
\newcommand{\bO}{\mathbf{O}}

\newcommand{\bW}{\mathbf{W}}

\newcommand{\bZ}{\mathbf{Z}} 
\newcommand{\cE}{\mathcal{E}} 

\newcommand{\numtoken}{n} 
\newcommand{\latd}{d}

\newcommand{\core}{R} 
\newcommand{\lat}{\bZ} 
\newcommand{\chunk}{L} 

\newcommand{\obs}{\bO} 
\newcommand{\LRM}{LRM} 
\newcommand{\Proj}{\mathbf{Proj}} 
\newcommand{\figref}[1]{Figure~\ref{#1}}

\usepackage{pifont} 
\newcommand{\cmark}{\ding{51}}
\newcommand{\xmark}{\ding{55}}
\usepackage{tikz}
\usepackage{subcaption}
\usepackage{stfloats}
\usepackage{float}
\usepackage{tabularx}
\newcolumntype{Y}{>{\raggedright\arraybackslash}X}

\usepackage[most]{tcolorbox}
\newtcblisting{promptbox}[2][]{%
  listing only,
  \ifblank{#2}{notitle}{title={#2},fonttitle=\bfseries},
  colback=gray!4,
  colframe=gray!60,
  boxrule=0.4pt,
  arc=1mm,
  left=1.2mm,
  right=1.2mm,
  top=0.5mm,
  bottom=0.5mm,
  listing options={
    basicstyle=\ttfamily\tiny,
    breaklines=true,
    columns=fullflexible,
    keepspaces=true,
    showstringspaces=false,
    upquote=true
  },
  #1
}

\newcommand{\method}{FSRM}
\newcommand{\Method}{Fast--Slow recurrent model}

\newcommand{\akorn}{AKOrN\xspace}
\newcommand{\Localmaze}{Local-Maze\xspace}
\newcommand{\Minigrid}{Minigrid\xspace}

\newcommand{\maso}[1]{\textcolor{red}{\textbf{}#1}}
\newcommand{\kohei}[1]{\textcolor{black}{\textbf{}#1}}
\newcommand{\takashiro}[1]{\textcolor{black}{\textbf{}#1}}

\newcommand{\rev}[1]{\textcolor{blue}{\textbf{}#1}}

\usepackage[preprint]{styles/neuripsfiles/neurips_2026}

\usepackage[utf8]{inputenc} 
\usepackage[T1]{fontenc}    
\usepackage{hyperref}       
\hypersetup{hidelinks}
\usepackage{url}            
\usepackage{booktabs}       
\usepackage{colortbl}       
\usepackage{amsfonts}       
\usepackage{nicefrac}       
\usepackage{microtype}      
\usepackage{xcolor}         

\usepackage{enumitem}
\newenvironment{tight_itemize}[2][1.5em]
  {\begin{itemize}[
    topsep=0pt,
    itemsep=#2,
    parsep=0pt,
    partopsep=0pt,
    leftmargin=#1
  ]}
  {\end{itemize}}

\title{Exploration of Fast-Slow Latent Recurrence for Train-Short, Test-Long Generalization}

\author{%
\begin{tabular}{c}
Shota Takashiro$^{1,*}$ \quad
Masanori Koyama$^{1}$ \quad
Takeru Miyato$^{2}$ \\
Yusuke Iwasawa$^{1}$ \quad
Yutaka Matsuo$^{1}$ \quad
Kohei Hayashi$^{1}$ \\[0.8em]
{\small $^{1}$The University of Tokyo, Tokyo, Japan} \\
{\small $^{2}$University of Tübingen, Tübingen, Germany} \\[0.8em]
{\small $^{*}$Correspondence: \texttt{takashiro@weblab.t.u-tokyo.ac.jp}}
\end{tabular}
}

\begin{document}

\maketitle


\begin{abstract}

We study out-of-distribution generalization in streaming tasks where models are trained on short sequences but must operate over much longer, unknown horizons under bounded memory. Our focus is on a persistent fast–slow recurrent formulation in which a latent state is maintained across observations rather than reset at each stream step. For each incoming observation, the model performs multiple weight-shared latent updates with a recurrent core and then carries the resulting state forward to the next observation. This allows the model to maintain and refine a compact stream-level state without reprocessing a growing context. We evaluate this formulation across symbolic sequence prediction, supervised navigation, and partially observable reinforcement learning tasks. Across these settings, persistent latent recurrence improves OOD generalization over recurrent, state-space, and Transformer baselines. Through recurrent-core ablations, we identify architectural ingredients that are consistently associated with strong OOD performance, including state-dependent transitions and feature-wise nonlinear mixing. Together, these results highlight the value of revisiting persistent recurrence as an architectural bias for more generalizable sequence prediction.

\end{abstract}

\section{Introduction}


Humans can often extrapolate symbolic streams generated by a fixed rule far beyond the lengths observed during learning, or solve navigation puzzles of previously unseen size, after seeing only small, bounded examples.
How can we endow machines with this capability?
We posit that this is not merely a question of memory maintenance: 
a finite-size memory cannot indefinitely store or attend to the raw history of a stream whose horizon is not known in advance.
Instead, the model must learn, from short observations, a compact task-relevant rule of the stream, such as its generative rule in symbolic streams, or a chain of subtasks in a puzzle that exists irrespective of stream length or environment size \citep{dehaene2022symbols, amalric2017language}.

Recent advances in machine learning suggest a possible architectural route toward this kind of extrapolation: rather than storing the entire history, a model may learn to infer a compact latent procedure. 
Latent recurrent reasoning, which iterates a weight-shared block in 
latent space, has recently emerged as a way to build related capabilities 
directly into model architectures 
\citep{Geiping2025LatentReasoning, HRM, TRM, ouro2025, mcleish2025retrofit}.
This mechanism, \emph{especially with recurrent blocks based on attention}, 
has been shown to represent programs and universal computation 
\citep{giannou2023programmable, xu2025expressive} and, most relevantly
to the present work, to learn and execute iterative algorithms at test 
time in the form of in-context regression, gradient-descent-like updates, 
and other pattern-driven procedures 
\citep{yang2024looped, Akorn, saunshi2025latent}.
Yet the in-depth study of these methods has so far focused on 
\emph{static input}: the loop iterates over a fixed input, and a naive 
application to unknown-horizon streaming tasks of the kind described above
would require a context that grows with stream length. \\

We hypothesize that, by allowing the latent loop to persist alongside 
the stream, we can extend the ability of latent reasoning to capture, 
from a short sequence, the rules governing the input stream \textit{and thereby enable OOD generalization at horizons far beyond training}.
We therefore combine latent recurrent modeling with
the following persistent latent recurrence, which updates
the latent $\lat$ using a recurrent core block $\core$:
\begin{equation}
\begin{aligned}
\lat_s(t+1) = \core\left(\lat_s(t),\, \bC_s \right), 
\qquad \lat_{s+1}(0) = \lat_s(T),
\qquad t = 0, \ldots, T-1, 
\end{aligned}
\label{eq:fastslow} 
\end{equation}
where $s \in \{1, 2, \ldots\}$ indexes observations along the stream, 
$t$ indexes the inner latent recurrence, and $\bC_s = \mathrm{Encoder}(\obs_s) $ is the feature of the 
observation $\obs_s$ at step $s$.
The core $\core$ is applied $T$ times for each observation, and the latent is carried
forward without reset.
This describes a dynamical system on the latent variable that is not reset at every observation step, and it reduces to an RNN when $T=1$; when $T>1$, 
the same recurrent core is applied multiple times per observation, yielding a \textbf{weight-shared}, deep-transition type update between consecutive stream steps in the sense of~\citep{pascanu2014deep}.
Indeed, related structures have appeared in broader forms: adaptive computation methods such as ACT~\citep{ACT} can in principle be applied to recurrent systems of this kind to modulate the amount of computation, while \citet{schone2025implicit} explore analogous persistent updates with Mamba cores in pursuit of a balance between parallelism and representation power.
The question still remains as to which designs of $\core$ support OOD generalization on streaming tasks, and whether the benefits of recent findings in latent recurrent reasoning can be extended to such tasks.

Our contribution is not the introduction of recurrence per se, but the investigation of latent recurrent reasoning combined with this persistent streaming formulation, identifying effective designs of $\core$ for OOD generalization on sequential tasks requiring reusable rules.
Empirically, we make two valuable observations: 
\begin{tight_itemize}{5pt}
\item The transformer-based cores, as used in recent
latent reasoning methods, yield strong OOD generalization, practically extending the merit of static counterparts. 
We find that our recurrent core family (i.e., multiple recurrence) plays an important role: models without this persistent fast--slow mechanism, including looped Transformers, do not exhibit OOD generalization.
\item In designing $\core$, the following two structures are particularly important: (i) \textit{state-dependent} mixing of the tokens and (ii) nonlinear dimensional mixing. 
For example, an LSTM+FFN core also performs competitively with transformer-based cores.   
\end{tight_itemize}

Concretely, on the Dyck language \citep{hewitt2020rnns, merrill2024illusion},
a canonical state-tracking benchmark, our model of this family, trained on sequences of
length $\leq 40$, maintains $\approx 80\%$ token accuracy up to length
$10^5$, three orders of magnitude beyond training.
Our models also outperform baselines such as 
Looped Transformer~\citep{Looped} on a limited-vision maze navigation task,
showing that rules learned in small environments can transfer to 
larger ones.
Examining the learned latents, we further find that they are highly 
structured: they form a low-dimensional manifold indexed by task-relevant discrete  
states (e.g., stack depth in Dyck), 
and the dynamics reuse
representations, in that states sharing the same task-relevant features cluster together in latent space.

\section{Related Work}


Out-of-distribution (OOD) generalization along axes such as length extrapolation (small-to-large) is a long-standing challenge.
Many approaches have been proposed to address this issue, including improved positional encodings (PEs)~\citep{pe1su2021roformer, pe2press2022train, pe3chen2023extending, peng2024yarn} and looped Transformers~\citep{Looped}.
More recently, looping and recurrence have been identified as particularly effective mechanisms for improving generalization in these settings.
These approaches repeatedly apply weight-shared layers in a loop on a static input~\citep{UT, Geiping2025LatentReasoning, Akorn, blayney2026mechanistic}, and \citet{giannou2023programmable, xu2025expressive, yang2024looped, Looped} report the ability of these models to construct explicitly repeatable programs or rules.
Unlike language-space reasoning approaches~\citep{openai2024o1, deepseekai2025r1}, they "require no specialized trace data, can work with small context windows, and can capture the reasoning not easily expressed in the discrete output space of words and symbols"~\citep{Geiping2025LatentReasoning}.
However, all of these approaches are designed for static inputs, without an explicit mechanism for capturing reasoning patterns \textit{that exist across the stream.}

Biological systems handle an analogous problem by integrating ongoing input across a hierarchy of intrinsic timescales that persist on top of incoming stimuli, with longer integration windows in higher-order areas~\citep{bio0, bio1, bio2}.
This combines latent recurrent processing with persistence across the stream under bounded neural resources. 
The mechanism~\eqref{eq:fastslow} that we investigate is \textit{a} minimal architectural counterpart of this combination, motivated by OOD generalization on streaming tasks from short training sequences, a setting in which neither the training horizon nor a growing context is available at test time. 
Two recent lines of work share part of the combination but stop short. HRM~\citep{HRM} and CTM~\citep{CTM} draw on similar biological motivation and use multi-timescale latent loops, but again apply them to static inputs. Coconut~\citep{hao2024training} maintains latent thoughts persistently, but does so as part of a growing conditional context, so its memory and computation scale with sequence length rather than remaining bounded.

Broadly, our investigation also belongs to the family of nonlinear RNNs
with inner loops between observations.
The mechanism \eqref{eq:fastslow} is itself agnostic to the choice of $\core$
and can in principle also be combined with adaptive halting~\citep{ACT} to make
$T$ observation-dependent as well.
With $T>1$, it can also be viewed as weight-tied deep-transition~\citep{pascanu2014deep}.
Most relevantly, \citet{schone2025implicit} and \citet{siems2025deltaproduct} use
related inner-loop mechanisms with $\core$ based on Mamba and DeltaNet, respectively.
While such works have been studied through the lens of expressivity limits
and parallelism-expressivity trade-offs
\citep{merrill2024illusion, merrill2023parallelism},
we tackle in this paper a question orthogonal to whether,
for each input length $n$, there exists \textit{some} setting of weights
in the recurrent model that solves the problem at length $n$.
That is, we instead tackle the following question:
``is it possible to train a fixed-weight network on short sequences such that
it learns languages or patterns that extrapolate to horizons far beyond training?''
We argue that this is a more difficult problem than expected.
For example, the $\mathrm{Dyck}(k,m)$ language with $k$ bracket types and fixed-depth $m$ admits constructions by
fixed-depth self-attention networks with modest per-token memory
\citep{yao2021bounded}, and log-precision transformers lie within
$\mathsf{TC}^0$ \citep{merrill2023parallelism};
expressivity therefore does not appear to be the binding constraint.
Yet, as we show in Section~\ref{sec:results}, transformers trained on short sequences
still fail to extrapolate to long horizons.
This is consistent with \citet{anil2022exploring}: the relevant gap here
is one of learnability from short training data rather than of expressive power.
By combining recurrence with the mechanism \eqref{eq:fastslow} that increases
the number of persistent loops, we explore the mechanisms that can help close this gap.
We also show in Section~\ref{sec:latent-analysis} that, through this combination, we inherit the ability of latent recurrence to exhibit metastable clustering and cyclical stability \citep{blayney2026mechanistic, geshkovski2024dynamic, geshkovski2025mathematical}.

\section{Latent Recurrent Model}\label{sec:recurrence_structure}

Given a static observation $\obs$ of fixed length, the Latent Recurrent Model (\LRM) \citep{EBT, Geiping2025LatentReasoning, Akorn, Looped, yang2024looped} first processes $\obs$ through a \textit{prelude} encoder.
Starting from a randomly initialized latent $\lat^{\rm init} \in \RR^{\numtoken \times \latd}$ of $\numtoken$ tokens in $\latd$ dimension, the model sets $\lat(0) = \lat^{\rm init}$ and iteratively updates the latent state by injecting the fixed encoded input at every step over $t = 0, 1, \ldots, T-1$:
\begin{align}
\lat(t+1) = \core \left(\lat(t), \bC\right) \label{eq:LRM}
\end{align}
where $\bC = \mathrm{Encoder}(\obs)$.  
The recurrence of this produces $\lat(T)$, a latent conclusion of depth $T$, which is passed through a \textit{coda} decoder, $\mathrm{Decoder}(\lat(T))$, for inference.
\citet{Geiping2025LatentReasoning} instantiate $\core$ with a Transformer block, for example,
\begin{align}
    \core_{TF}(\lat(t), \bC )  = n_4(\tilde{\lat} + \mathbf{FFN}(n_3(\tilde{\lat}))  ~~~\textrm{where} ~~~  \tilde{\lat}  = n_2(\lat(t) +  \textrm{Attn}(n_1(h[\lat(t), \bC])))) \label{eq:lrm-transformer-core}
\end{align}
where $\textrm{Attn}$ denotes the usual attention mechanism used in prior latent-reasoning models \citep{blayney2026mechanistic, Geiping2025LatentReasoning}, 
$h$ is an adapter that merges $\lat$ with the observation, $n_k$s are layer norms of choice, and \textbf{FFN} acts as a token-wise MLP map $\RR^d \to \RR^d$.
Another recently reported model that belongs to this line is \akorn \citep{Akorn}, which constrains $\lat$ to be a set of oscillating tokens on the hypersphere $S^{\latd-1}$ by projecting the update $\lat(t+1) - \lat(t)$ onto the tangent space of $S^{\latd-1}$ and by adding rotational momentum.
We mention \akorn because we found it to be a stably performing choice of $\core$ throughout our investigation. See Appendix~\ref{app:impl-J} for details. 


\section{Fast--Slow Coupling of Latent Recurrence with the Observational Process}
\label{sec:fastslow}

Since the rule we wish to extract governs the evolution of the entire stream, the latent recurrence must be allowed to run across observations rather than reset between them.
The family of models we investigate in this study implements this by injecting observations into a recurrence that is never reset.
Given a stream $\obs_1, \obs_2, \ldots, \obs_s, \ldots$ and an initial latent $\lat^{\rm init}$, we set $\lat_1(0) = \lat^{\rm init}$ and evolve the latent jointly with the stream using the persistent fast--slow update
for observation indices $s \in \{1, 2, \ldots\}$.
The latent $\lat$ evolves $T$ times per observation, so the observation clock $s$ progresses $T$ times more slowly than the latent clock.
Inference is made from the decoded latent at each $s$.
We refer to this mechanism as \method{} (\Method) for short.

We investigate this mechanism under various choices of the recurrent core $\core$.
Although ACT~\citep{ACT} can be applied to this loop in general to make $T$ adaptive, we will focus our study on a fixed $T$ to isolate the design choice of $\core$ as well as the fast--slow formulation itself.
This mechanism can also be used in a hierarchical form for better stability and stronger representation; see Figure~\ref{fig:code-meta}.

\subsection{Recurrent core with state-dependent transition and feature-wise dimensional mixing} \label{sec:strong_core}

We will empirically show that choices of $\core$ equipped with
(i) \emph{state-dependent transition} and (ii) \emph{feature-wise dimensional mixing}
are particularly successful at OOD generalization on sequential tasks that
require learning a rule governing the stream.
We make these two properties precise below.
Because the latent $\lat$ evolves with the injected stream $\bC_s$, we reserve the word \textit{input} for $\bC_s$ rather than $\lat$.
See Table~\ref{tab:core-classification} for the classification of the cores in this study with respect to (i) and (ii).
 
\paragraph{(i) State-dependent transition.}
Let $\lat \in \RR^{\numtoken \times \latd}$ be the latent, with $\numtoken$ tokens of $\latd$-dimensional features, and let $\bC$ denote the encoded observation.
We say the transition $\core(\,\cdot\,, \bC) : \RR^{\numtoken \times \latd} \to \RR^{\numtoken \times \latd}$ has a \emph{state-dependent transition} if it contains a linear operator (linear in some part of $\lat$) whose coefficients themselves depend on $\lat$.
Equivalently, at the Jacobian level, this condition requires that $\partial_\lat \core(\lat, \bC)$ does not admit a factorization into fixed (state-independent) linear maps composed with diagonal matrices arising from pointwise nonlinearities.
This excludes vanilla RNN-style cores $\sigma(A\lat + B\bC)$, in which all linear operations have coefficients independent of $\lat$ and all $\lat$-dependence is mediated by pointwise nonlinearities.
The canonical example of state-dependent transition is attention: the linear operator $\mathrm{AttnMat}_\theta(\lat) \in \RR^{\numtoken \times \numtoken}$ acting on $\lat W_V$ is itself a function of $\lat$.
This distinction from state-independent transition has been made in prior work on dynamic/adaptive mixing \citep{adapt1_wang2022dynamixer, adapt2_wei2023active, adapt3_wei2023active}, where an update of the form $A(\lat) \lat$ is used.

\paragraph{(ii) Feature-wise dimensional mixing.}
Treating $\lat \in \RR^{\numtoken \times \latd}$ as $\numtoken$ rows $\lat_n \in \RR^{\latd}$, dimensional mixing is the family of maps $\RR^d \to \RR^d$ that act \emph{within} each row by mixing its $\latd$ feature coordinates.
The canonical instance is the per-token MLP.
Note that mixing across the \emph{token} axis $\numtoken$ (e.g.\ attention) is a different operation and does not by itself realize (ii).
\citet{tolstikhin2021mlpmixer} make this distinction explicit: MLP-Mixer combines a state-independent token mixer with a channel mixer.

\begin{table}[t]
\centering
\caption{Classification of recurrent cores $\core$ with respect to (i) state-dependent transition and (ii) feature-wise dimensional mixing; \cmark{} indicates the property is satisfied.
For our LSTM/Mamba2 cores, we use multiple internal updates per $t$, treating $\lat + \bC$ as the external input so the latent-dependent gates, projections, and channel mixing are applied multiple times when producing $\lat(t+1)$.
See Appendix~\ref{app:mamba_core_details} for details.
}
\setlength{\tabcolsep}{4pt}
\renewcommand{\arraystretch}{1.18}

\begin{tabularx}{\linewidth}{@{}l c c Y@{}}
\toprule
Core & (i) & (ii) & Mechanism / remark \\
\midrule
Conv
& \xmark & \xmark
& State-independent token/pixel mixing with linear channel mixing. \\

Attention
& \cmark & \xmark
& $\mathrm{AttnMat}_\theta(\lat)$ depends on $\lat$; no per-token FFN. \\

MLP-Mixer
& \xmark & \cmark
& Mixer weights are independent of $\lat$; MLP-style depth-2 mixing. \\


LSTM (loops per $t$)
& \cmark & \cmark
& Gated update over $\lat+\bC$, repeated multiple times per $t$. \\

Mamba2 (loops per $t$)
& \cmark & \cmark
& Selective/gated SSM over $\lat+\bC$, repeated multiple times per $t$. \\

Transformer
& \cmark & \cmark
& Attention gives (i); FFN gives (ii). \\

Linear Transformer
& \cmark & \cmark
& Transformer with linear attention. \\

AKOrN + FFN
& \cmark & \cmark
& Attention-derived couplings dependent on $\lat$. \\

\bottomrule
\end{tabularx}
\label{tab:core-classification}
\end{table}

\section{Experimental Settings}\label{sec:exp}

We conduct experiments to verify out-of-distribution (OOD) generalization 
across three settings: symbolic stream prediction on the Dyck language 
\citep{dyckpaper}, navigation of a maze under egocentric observation 
(local maze), and reinforcement learning on \Minigrid
\citep{10.5555/3666122.3669331}.
Unless otherwise specified, we instantiate \method{} with an AKOrN + FFN core with $T=5$ inner loops as the default FSRM configuration. Implementation details are provided in Appendix~\ref{app:impl-J}.
Results report means and standard deviations over $3$ random seeds ($5$ for \Minigrid{}), shown as shaded regions or error bars.

\newcommand{\DyckExampleGraphic}[1]{%
  \begin{tikzpicture}
    \node[inner sep=0pt,anchor=south west] (dyckexample) at (0,0) {%
      \includegraphics[width=\linewidth,trim=18 14 24 11,clip]{#1}%
    };
    \begin{scope}[x={(dyckexample.south east)},y={(dyckexample.north west)}]
      \fill[white] (0.000,0.61) rectangle (0.225,0.95);
      \fill[white] (0.000,0.14) rectangle (0.225,0.48);
      \node[anchor=east,font=\small] at (0.205,0.80) {input};
      \node[anchor=east,font=\small] at (0.205,0.34) {target};
    \end{scope}
  \end{tikzpicture}%
}

\begin{figure*}[t]
    \centering
    \captionsetup{skip=5pt}
    \begin{minipage}[t]{0.48\textwidth}
        \vspace{0pt}
        \centering
        \captionsetup[sub]{skip=0pt}
        \setcounter{subfigure}{0}

        \begin{minipage}[t][0.60\linewidth][t]{\linewidth}
            \centering
            \begin{subfigure}[t]{0.8\linewidth}
              \centering
              \DyckExampleGraphic{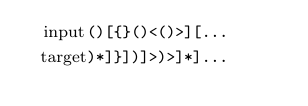}
              \caption{ID (random pattern)}
            \end{subfigure}
            \par\vspace{1.2em}
            \begin{subfigure}[t]{0.8\linewidth}
              \centering
              \DyckExampleGraphic{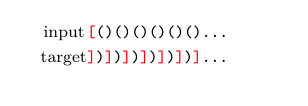}
              \caption{OOD (1-regular run)}
            \end{subfigure}
        \end{minipage}
        \caption{Dyck-$(30,5)$ examples. (a) The target is the closing token for the current stack top, or ``$*$'' when the stack is empty. (b) A $1$-regular run requires retaining the first unresolved bracket over a long repetition.}
        \label{fig:dyckexample}
    \end{minipage}
    \hfill
    \begin{minipage}[t]{0.48\textwidth}
        \vspace{0pt}
        \centering
        \captionsetup[sub]{skip=6pt}
        \setcounter{subfigure}{0}
        \begin{minipage}[t][0.60\linewidth][t]{\linewidth}
            \centering
            \begin{subfigure}[t]{0.48\linewidth}
                \centering
                \includegraphics[width=\linewidth]{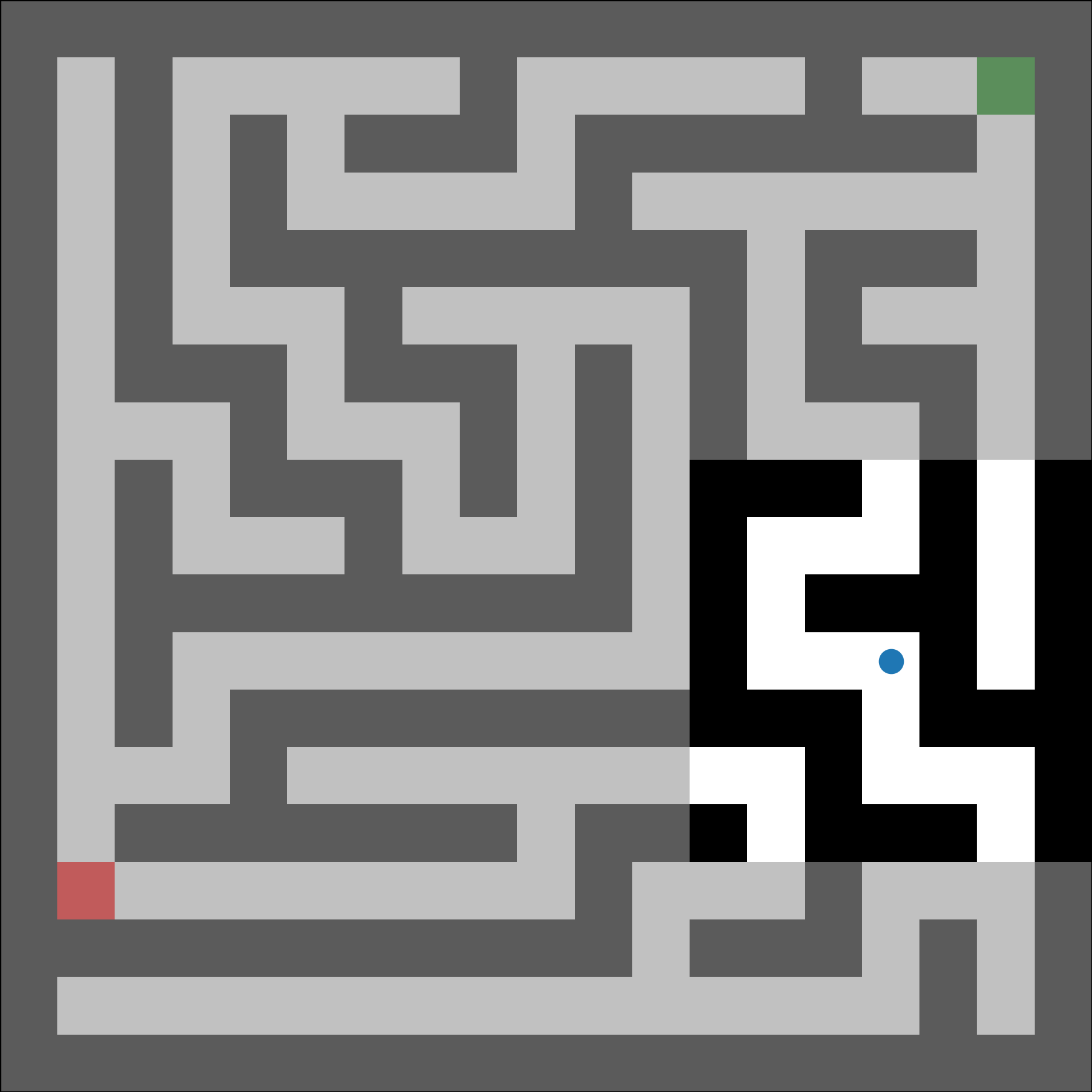}
                \caption{ID ($19\times 19$)}
            \end{subfigure}
            \hfill
            \begin{subfigure}[t]{0.48\linewidth}
                \centering
                \includegraphics[width=\linewidth]{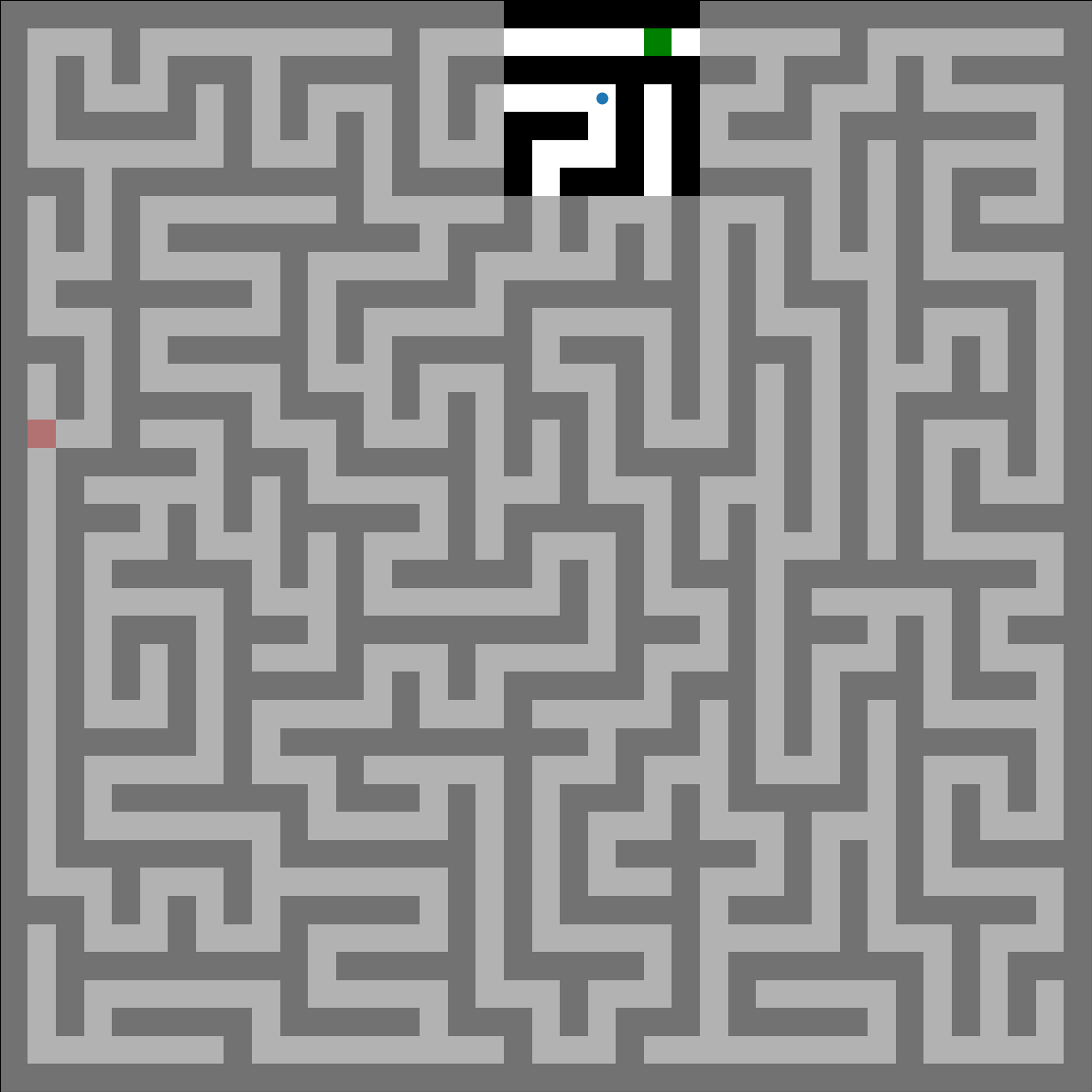}
                \caption{OOD ($39\times 39$)}
            \end{subfigure}
        \end{minipage}
        \caption{\Localmaze examples. Models are trained on small mazes (a) and evaluated on larger ones (b). The green and red cells denote the start and the goal. The observation is limited to a $7\times 7$ region centered on the current position.}
        \label{fig:maze-sample}
    \end{minipage}
    \vspace{-0.5\baselineskip}
\end{figure*}

\textbf{Dyck-$(30,5)$.} 
We use Dyck-$(30,5)$~\citep{hewitt2020rnns}, where $30$ is the number of bracket types and $5$ is the maximum stack depth, to test whether a model can infer a stack rule from short streams of length at most $40$ and apply it to much longer horizons.
At each position, the model predicts the closing bracket for the current stack top, or ``\verb|*|'' when the stack is empty.
Training and ID evaluation use randomly generated Dyck-$(30,5)$ strings; the ID curves extend this same random-pattern distribution beyond the training length, testing length generalization.
OOD evaluation additionally changes the data pattern to $1$-regular runs with repeated open--close blocks (Figure~\ref{fig:dyckexample}), so the Dyck setting tests both longer horizons and generalization from random patterns to regular-run patterns; Appendix~\ref{app:dyck-details} gives the full protocol.

\textbf{\Localmaze.}
\Localmaze extends the navigation task of \citet{CTM} and tests whether a geometric rule learned on $19\times19$ mazes transfers to larger $39\times39$ mazes.
The model observes a stream of egocentric $7\times7$ views collected by a fixed right-hand-rule navigator and predicts the shortest start-to-goal action sequence.
Because the global position is not directly observable, the task requires integrating local observations into a maze-level plan; Appendix~\ref{app:maze-details} gives the full protocol and supervision details.

\textbf{\Minigrid.}
We use three partially observable RL tasks from \Minigrid~\citep{10.5555/3666122.3669331}: DoorKey, MultiRoom, and LavaCrossing.
Agents are trained on a simple configuration and evaluated zero-shot on larger or more complex OOD environments, using sparse rewards and egocentric $7\times7$ observations.
Unlike Dyck-$(30,5)$ and \Localmaze, no ground-truth action sequence is provided; Appendix~\ref{app:rl-details} gives the task definitions, action space, and ID/OOD splits.

\textbf{Baselines.}
We choose baselines to isolate the two ingredients of \method{}: carrying a bounded stream-level state and refining it through multiple weight-shared latent updates before the next observation.
Recurrent, state-space, and cached-memory sequence models (LSTM~\citep{LSTM}, Mamba2/S5~\citep{gu2023mamba,dao2024transformers,smith2022simplified}, DeltaProduct~\citep{siems2025deltaproduct}, and TransformerXL~\citep{dai2019transformer}) all test whether sequence memory alone is sufficient.
Transformer variants~\citep{vaswani2017attention} and LLM/Qwen baselines test context-based alternatives based on attention, prompting, pretraining, scale, or fine-tuning.
Looped TF~\citep{Looped} and CTM~\citep{CTM} test whether latent or recurrent computation without the same persistent fast--slow streaming formulation is enough.

Baselines are task-matched: Dyck-$(30,5)$ uses LSTM, Mamba2, Transformer variants, prompted frontier LLMs, and fine-tuned Qwen3-4B-Base; \Localmaze uses LSTM, Mamba2, Transformer (TF), Looped TF, and DeltaProduct, with S5 and CTM in the appendix; and \Minigrid uses LSTM, Mamba2, and TransformerXL under the same PPO setup.
Checkpoints are selected by in-distribution validation after a shared learning-rate sweep, with architecture-specific hyperparameters reported in \ref{app:hyperparams} and Tables~\ref{tab:hparams-merged}--\ref{tab:model-hparams-maze-minigrid}.
\ref{app:baseline-details} gives the full training protocol.

\textbf{Investigated Questions:} Altogether, we tackle the following four questions:
\textbf{RQ1.} Does the persistent fast--slow loop improve OOD generalization over baseline approaches, including methods designed for static inputs?
\textbf{RQ2.} How does the number of inner-loop iterations per observation affect OOD performance?
\textbf{RQ3.} Is the gain due to weight-shared recurrence rather than simply adding heterogeneous, unshared depth?
\textbf{RQ4.} Which recurrent-core designs preserve the OOD gains?
The subsections answer these questions in order: we first compare OOD performance across tasks, then isolate the roles of inner-loop iterations and weight sharing, and finally ablate the recurrent core.

\section{Results and Ablations}\label{sec:results}

Across the streaming tasks, extending latent recurrence with the \method{} fast--slow loop improves OOD generalization over the corresponding one-step sequence-model baselines.
The strongest results come from cores that combine state-dependent transition with sufficient feature-wise mixing, although the value of adding a separate FFN depends on the task and on the core's native inductive bias.

\subsection{OOD generalization across tasks}
\begin{wrapfigure}[15]{r}{0.45\textwidth}
  \vspace{-1.9\baselineskip}
  \centering
  \captionsetup{skip=2pt}
  \includegraphics[width=\linewidth]{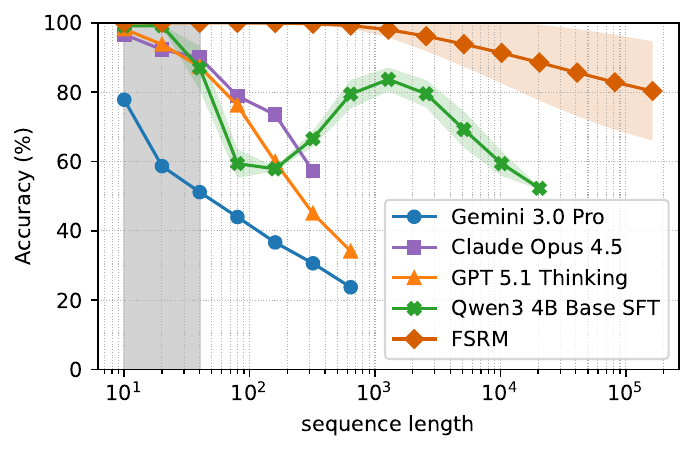}
  \caption{Dyck-$(30,5)$ accuracy vs.\ length. LLMs use algorithmic prompts. \method{} and Qwen3-4B-Base are trained only up to length $40$ (shaded area); \method{} generalizes to horizons far beyond training.}
  \label{fig:dyck-llms}
\vspace{-0.8\baselineskip}
\end{wrapfigure}

In all settings (Dyck-$(30,5)$, \Localmaze and \Minigrid), \method{} with an attention-based recurrent core (\ref{app:impl-J}) substantially improves OOD generalization over the main sequence-model baselines.

\begin{figure}[t]
  \centering
  \captionsetup{skip=6pt}
  \captionsetup[sub]{skip=3pt}
  \begin{subfigure}[t]{0.47\textwidth}
    \centering
    \includegraphics[width=\linewidth]{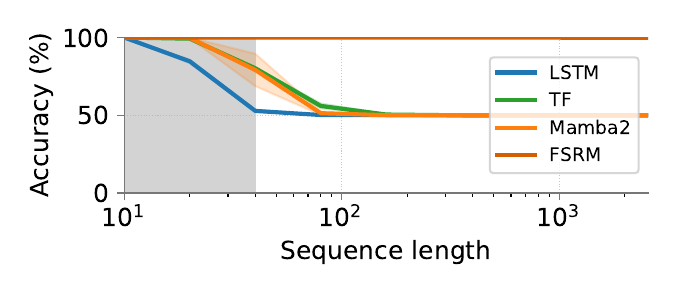}
    \caption{Dyck-$(30,5)$ OOD accuracy.}
    \label{fig:main-dyck-acc}
  \end{subfigure}
  \hfill
  \begin{subfigure}[t]{0.47\textwidth}
    \centering
    \includegraphics[width=\linewidth]{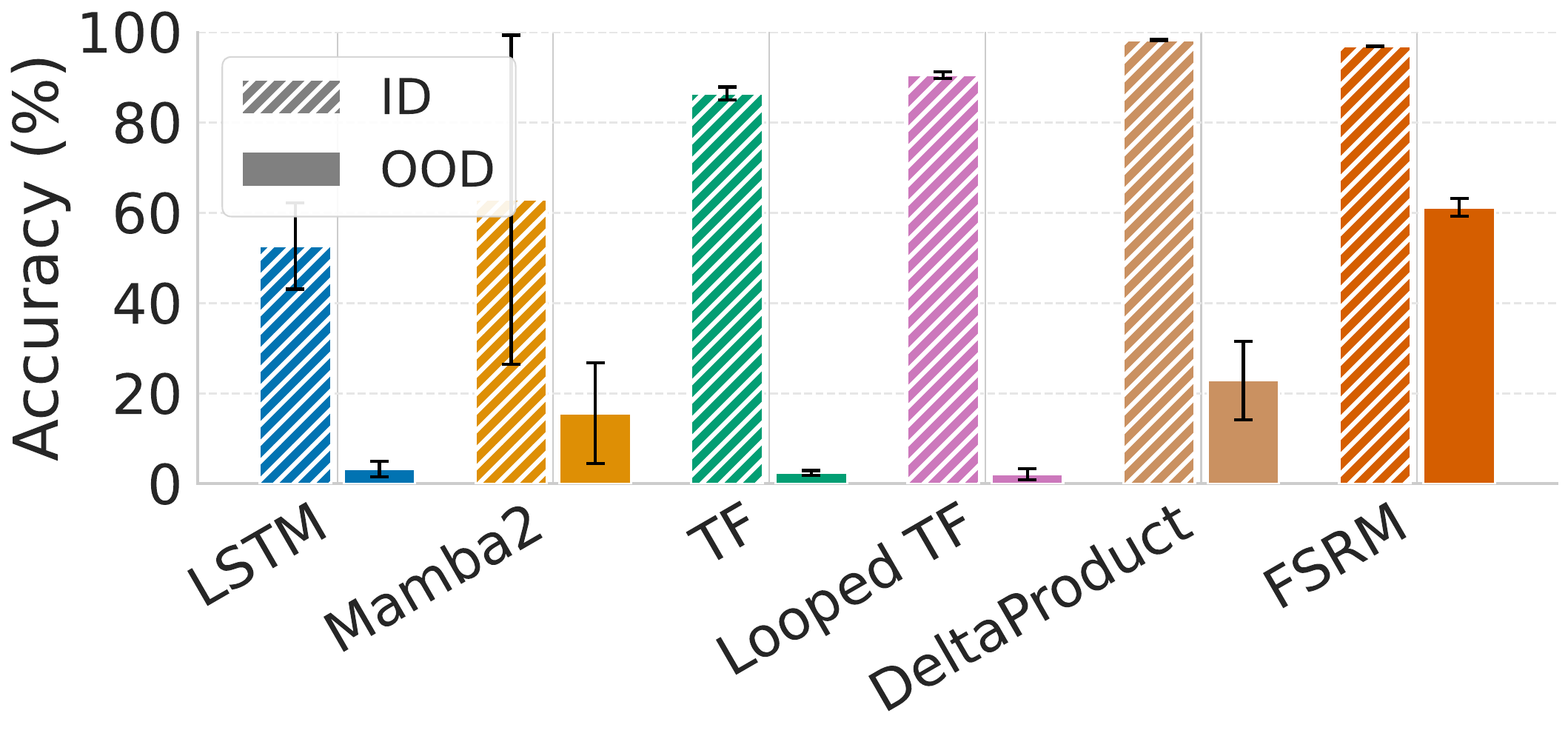}
    \caption{\Localmaze ID and OOD accuracy}
    \label{fig:main-maze-id-ood}
  \end{subfigure}

  \vspace{0.5em}
  \begin{subfigure}[t]{0.31\textwidth}
    \centering
    \includegraphics[width=\linewidth,trim=8 6 8 6,clip]{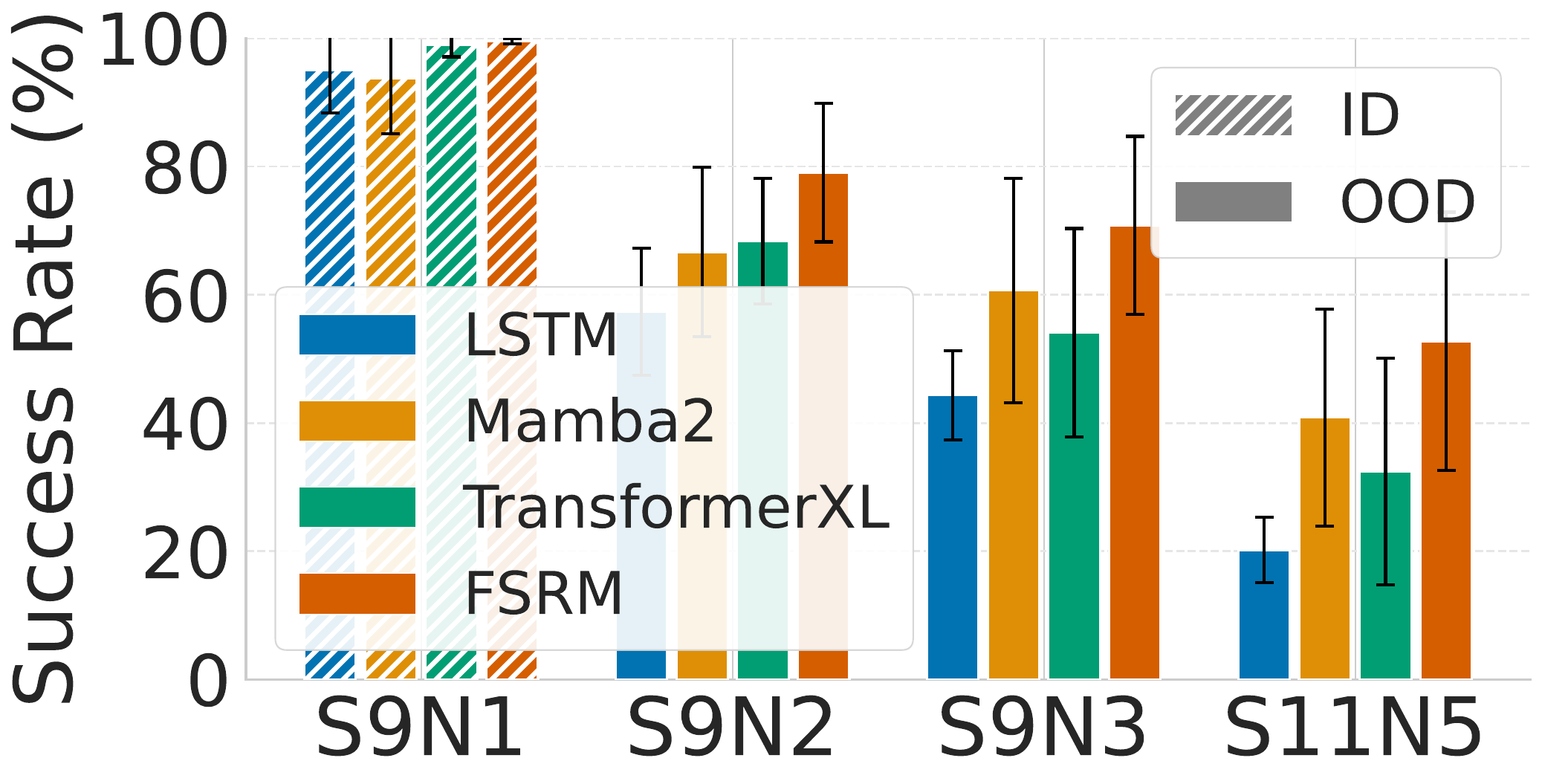}
    \caption{LavaCrossing}
    \label{fig:main-lava}   
  \end{subfigure}
  \hfill
  \begin{subfigure}[t]{0.31\textwidth}
    \centering
    \includegraphics[width=\linewidth,trim=8 6 8 6,clip]{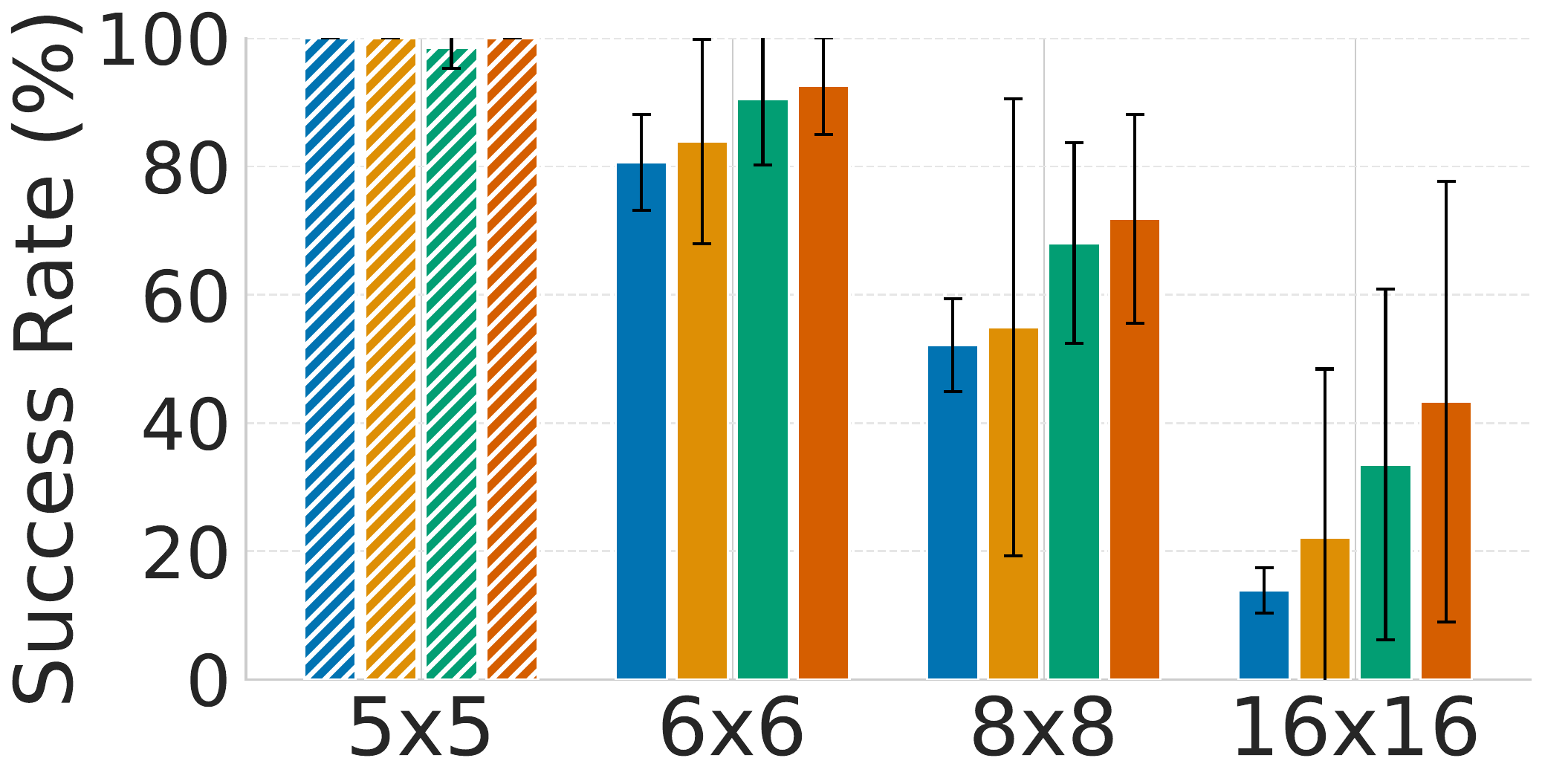}
    \caption{DoorKey}
    \label{fig:main-doorkey}
  \end{subfigure}
  \hfill
  \begin{subfigure}[t]{0.31\textwidth}
    \centering
    \includegraphics[width=\linewidth,trim=8 6 8 6,clip]{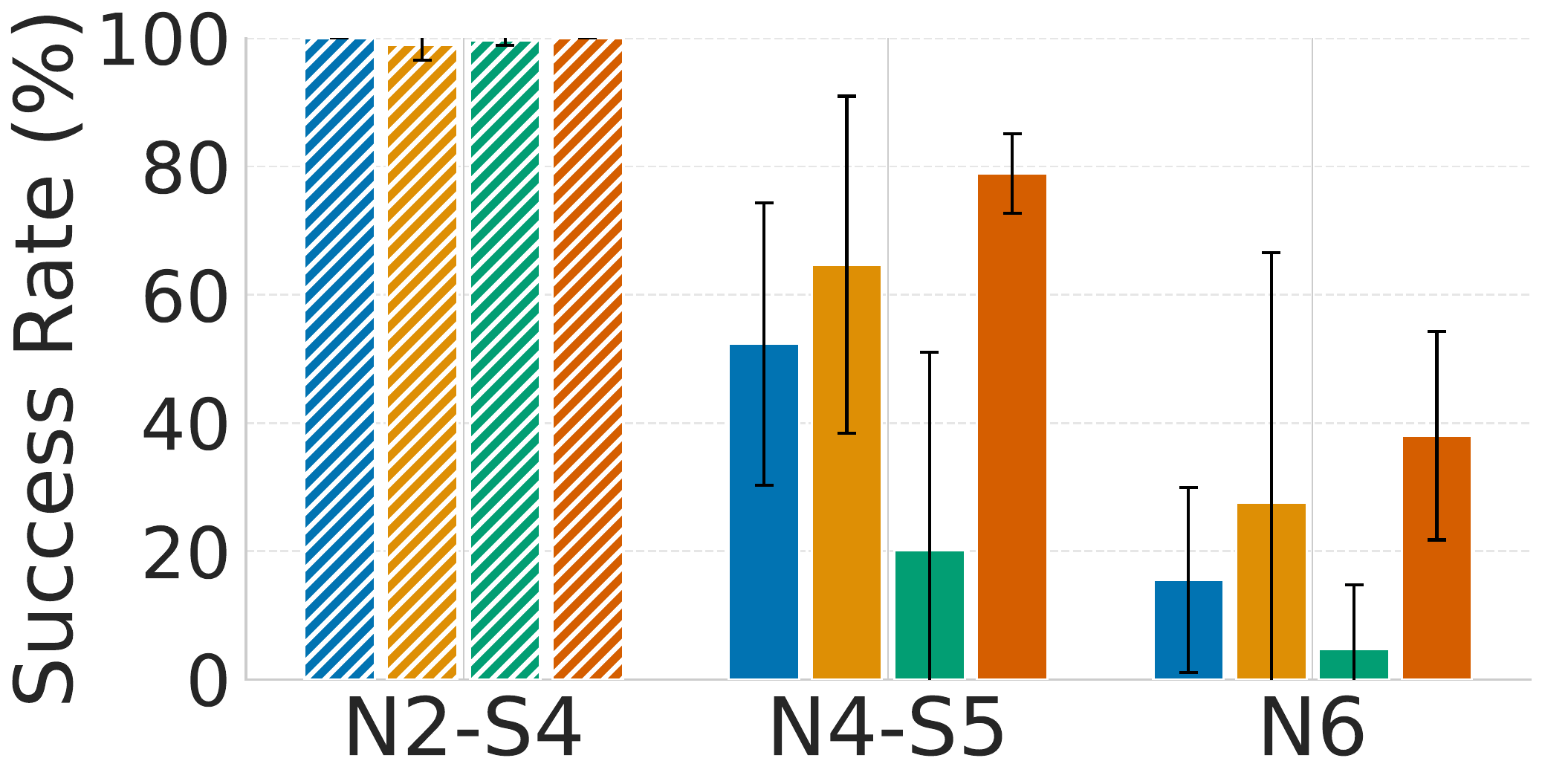}
    \caption{MultiRoom}
    \label{fig:main-multiroom}
  \end{subfigure}
  \caption{Main OOD generalization results. (a) Dyck-$(30,5)$ OOD accuracy beyond the training range (shaded area); the corresponding ID curve is in Appendix~\ref{app:dyck-id-ood-accuracy}. (b) \Localmaze ID and OOD accuracy. (c--e) \Minigrid{} success rates.}
  \label{fig:main-results}
  \label{fig:results-maze}
  \label{fig:results-minigrid}
\end{figure}

On Dyck-$(30,5)$ (Figure~\ref{fig:main-results}\subref{fig:main-dyck-acc}), 
the baselines degrade soon after OOD lengths exceed the training horizon, 
whereas \method{} maintains high accuracy at substantially longer horizons. 
On \Localmaze{} (Figure~\ref{fig:main-results}\subref{fig:main-maze-id-ood}), 
baselines solve the in-distribution mazes but largely fail on larger unseen mazes; 
by contrast, \method{} preserves a substantial fraction of its training-regime accuracy. 
See \figref{fig:maze-compute-match} for matched-compute comparisons.
On the \Minigrid{} reinforcement-learning tasks (Figure~\ref{fig:main-results}\subref{fig:main-lava}--\subref{fig:main-multiroom}), 
\method{} also achieves higher OOD performance than the baselines, including TransformerXL~\citep{dai2019transformer}, on environments larger than those seen during training.
 

As a stronger stress test on Dyck-$(30,5)$,
we extend the OOD evaluation horizon by another order of magnitude and compare
against LLM-based baselines in Figure~\ref{fig:dyck-llms}. The frontier LLMs are
evaluated zero-shot with algorithmic hints, but their accuracy degrades rapidly as
the sequence length increases; moreover, their internal reasoning uses up much of the output-token budget, and they can only be evaluated reliably up to lengths
on the order of $10^3$. We therefore also fine-tune the open-weight
Qwen3-4B-Base model on the same Dyck-$(30,5)$ split and evaluate it under the
same protocol. The fine-tuned Qwen remains accurate over
longer horizons than the frontier LLMs, but still falls to the chance rate of
approximately $50\%$ at length $20{,}480$, roughly an order of magnitude shorter
than the horizon at which \method{} maintains high accuracy.

\subsection{\texorpdfstring{Fast-loop structure: $T$ and weight sharing}{Fast-loop structure: T and weight sharing}}\label{sec:t-and-share}

Two structural choices in the \method{} update drive its OOD gain: ($i$) $T>1$ inner iterations of $\core$ per observation and ($ii$) weight sharing across these iterations. We ablate each in turn, and find that larger $T$ improves OOD performance until saturation and that weight sharing is effective.

\paragraph{Effect of fast-loop iterations.}

\begin{wraptable}[18]{r}
{0.5\textwidth}
\vspace{-1.2\baselineskip}
\centering
\captionsetup{skip=2pt,justification=raggedright,singlelinecheck=false}
\setlength{\tabcolsep}{3pt}
\renewcommand{\arraystretch}{1.03}
\caption{Fast-loop ablations on \Localmaze.
Shared-loop rows vary the number of inner-loop updates per observation ($1\otimes T$), while stack rows replace the shared loop with $L$ unshared layers applied once per observation ($L\otimes1$). The Params/FLOPs column reports parameters and FLOPs relative to Base ($1\otimes1$).}
\label{tab:fast-loop-ablation}
\label{tab:ablation_weightshare}
\begin{tabular}{@{}l c c@{}}
\toprule
Model & Params/FLOPs & OOD Acc. \\
\midrule
\rowcolor[gray]{0.82}
\multicolumn{3}{c}{Shared loop: train-time $T$ sweep} \\
Base ($1 \otimes 1$)  & $1{\times}/1{\times}$  & $0.306 \pm 0.135$ \\
Loop ($1 \otimes 2$)  & $1{\times}/2{\times}$  & $0.431 \pm 0.186$ \\
Loop ($1 \otimes 4$)  & $1{\times}/4{\times}$  & $0.595 \pm 0.036$ \\
Loop ($1 \otimes 5$)  & $1{\times}/5{\times}$  & $0.612 \pm 0.019$ \\
Loop ($1 \otimes 8$)  & $1{\times}/8{\times}$  & $\mathbf{0.673 \pm 0.006}$ \\
Loop ($1 \otimes 16$) & $1{\times}/16{\times}$ & $0.673 \pm 0.032$ \\
\midrule
\rowcolor[gray]{0.82}
\multicolumn{3}{c}{Weight-sharing control} \\
Stack ($5 \otimes 1$)  & $2{\times}/5{\times}$  & $0.509 \pm 0.098$ \\
Stack ($10 \otimes 1$) & $3{\times}/10{\times}$ & $0.445 \pm 0.164$ \\
\bottomrule
\end{tabular}
\vspace{-0.8\baselineskip}
\end{wraptable}

The number of fast inner-loop iterations $T$ controls how many recurrent updates of $\core$ occur between successive observations.
We sweep $T$ at training time while keeping all other components fixed.
As shown in Table~\ref{tab:fast-loop-ablation}, OOD accuracy improves with $T$, with diminishing returns at the largest $T$ tested.
Figure~\ref{fig:t-scaling} in the Appendix reports the complementary test-time scaling experiment, where $T$ is varied at inference after training.

\paragraph{Effect of weight sharing in the fast loop.}

\method{} applies the same update rule $\core$ repeatedly for $T$ inner steps; 
that is, it is \emph{recurrent} in the sense that the update is \emph{weight-shared} across fast iterations. 
To isolate the contribution of this sharing, 
we replace the $T$ shared iterations with an \emph{unrolled} residual stack of $T$ distinct layers applied once per observation---a construction equivalent to a deep-transition RNN \citep{pascanu2014deep}---keeping the overall fast--slow interface unchanged.
Table~\ref{tab:fast-loop-ablation} shows that weight sharing is crucial for OOD generalization.
With the same inner-loop budget ($T=5$), the shared recurrent loop reaches roughly $61\%$ OOD accuracy, whereas the unshared $5$-layer stack drops to about $51\%$, despite using roughly twice as many parameters.
Increasing the unshared stack to $10$ layers does not close this gap; instead, OOD accuracy falls further to around $45\%$, with substantially higher variance.
Together, these results support the design choice that \emph{looped recurrence}---not added depth or parameter count---is the effective ingredient of \method{}, aligning with prior work showing that weight-shared looped computation can learn reusable iterative procedures \citep{saunshi2025latent}.

Having established that \method{} generalizes OOD across all three task families when $\core$ is an attention-based block, and that both $T>1$ and weight sharing are necessary, we next ask which other choices of $\core$ inherit this property---and which of its structural ingredients are responsible.

\subsection{Core design}\label{sec:core}


\begin{figure*}[t]
  \centering
  \captionsetup[sub]{skip=2pt}
  \begin{subfigure}[t]{0.48\textwidth}
    \centering
    \includegraphics[width=\linewidth]{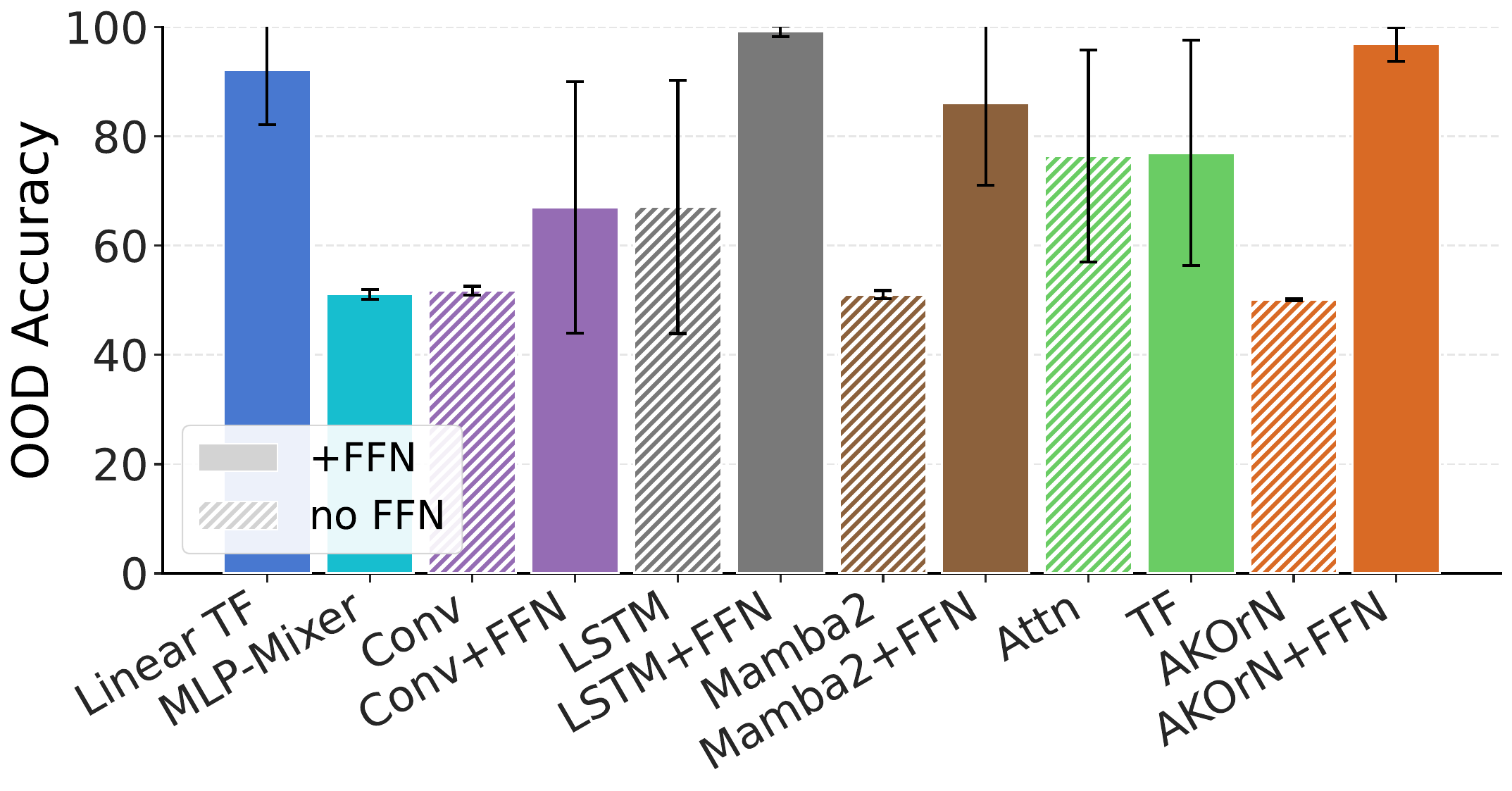}
    \caption{Dyck-$(30,5)$ OOD accuracy.}
    \label{fig:dyck-core-ablation}
  \end{subfigure}
  \hfill
  \begin{subfigure}[t]{0.48\textwidth}
    \centering
    \includegraphics[width=\linewidth]{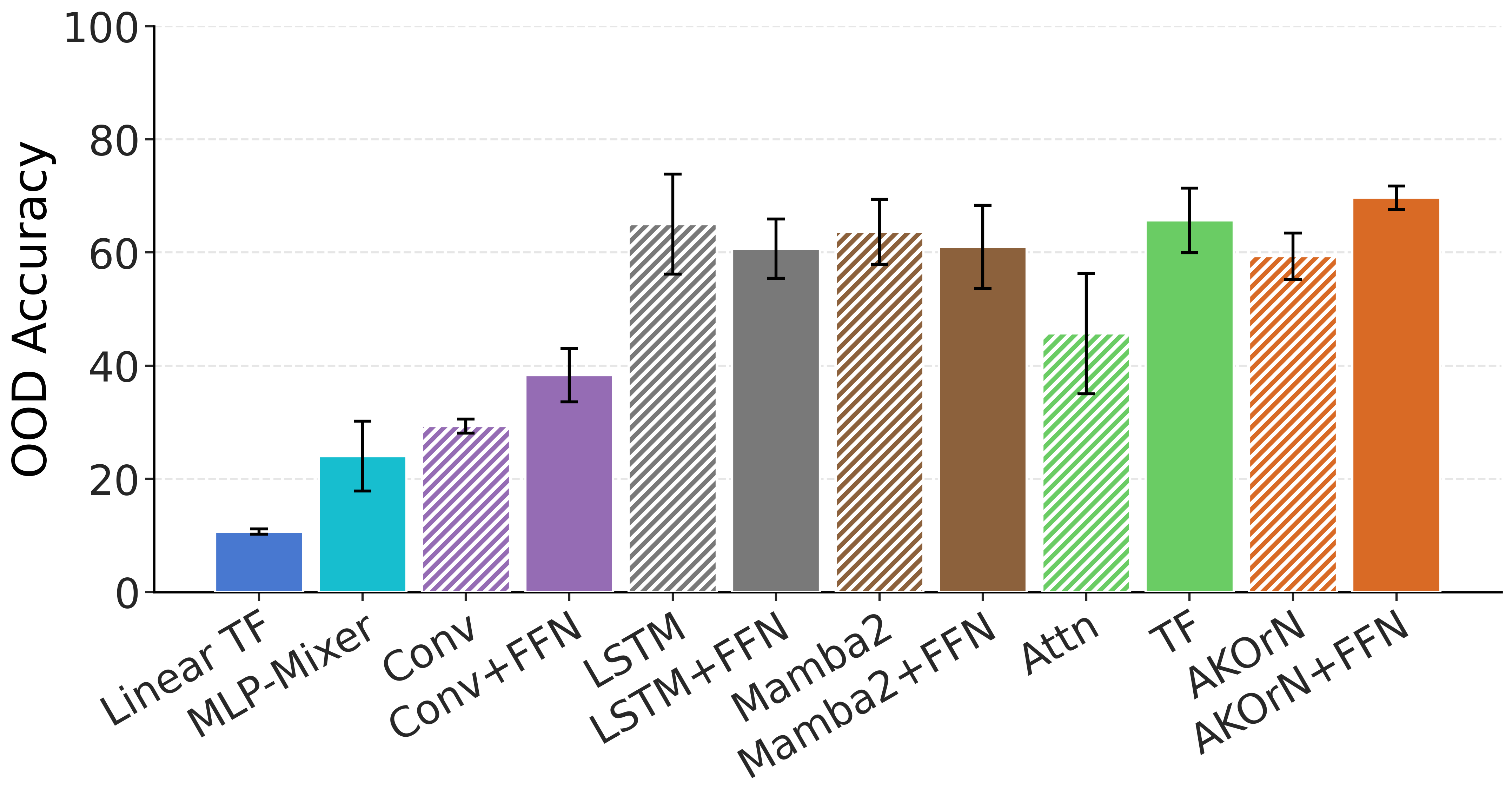}
    \caption{\Localmaze OOD accuracy ($39\times39$).}
    \label{fig:module-ablation-ood}
  \end{subfigure}
  \caption{Core ablations on (a) Dyck-$(30,5)$ at lengths $1{,}280$--$2{,}560$ and (b) \Localmaze under OOD evaluation.
    Solid bars include the native or augmented FFN block; hatched bars ablate it.
    ID results are in \ref{app:dyck-core-ablation} and~\ref{app:maze-core-id}.}
  \label{fig:module-ablation}
\end{figure*}


We ablate the design of $\core$ directly while keeping the encoder, decoder, $T$, and training protocol fixed.
We tested the variants of $\core$ classified in Table~\ref{tab:core-classification}.%
\footnote{The \Localmaze core ablation uses a slightly larger model than the preceding experiments; consequently, its absolute accuracy values are slightly higher, and comparisons should be read primarily within this ablation.}
Here, "+FFN" denotes an additional FFN wrapper, e.g. $\core_{\textrm{X + FFN}} = \textrm{FFN} \circ \core_X$.
\akorn~\citep{Akorn} is a recently introduced mechanism based on the dynamics of oscillating token particles on the hypersphere, compatible with an extension of a Transformer-like mechanism (see Appendix~\ref{app:impl-J}).
For the LSTM and Mamba2 core designs, we use inputs with chunk size $\chunk$ and additionally loop $\chunk$ times within each transition $\lat_s(t) \to \lat_s(t+1) \in \RR^{\chunk \times \latd}$. 
Since they apply latent-dependent gates and channel modulation across $\chunk$ loops, they satisfy (i) and (ii) as well; see \ref{app:mamba_core_details}.



Figure~\ref{fig:module-ablation} ablates the recurrent core under fixed encoder, decoder, $T$, and training protocol.
We focus on the two axes of Section~\ref{sec:strong_core}: state-dependent transition (i) and feature-wise dimensional mixing (ii).
Overall, we observe that strong OOD generalization concentrates among cores that satisfy both (i) and (ii), though the relative importance of (ii) varies by task.

As for the importance of (i), MLP-Mixer supplies a feature-wise mixer (ii), but its token mixer is state-independent, and its OOD accuracy stays well below the stronger cores.
This conclusion is the empirical counterpart of the distinction Section~\ref{sec:strong_core} draws from \citet{tolstikhin2021mlpmixer} and the dynamic-mixing line of \citet{adapt1_wang2022dynamixer} and \citet{adapt2_wei2023active,adapt3_wei2023active}.

The importance of (ii) appears in the FFN ablation, which toggles an explicit feature-wise mixer while keeping the rest of the core fixed.
On \Localmaze, cores whose native update is light on per-token mixing, namely Conv, Attn, and AKOrN, benefit substantially from an external FFN, while cores that already realize (ii) internally, LSTM and Mamba2, change only modestly when an FFN is appended, and slightly degrade in our runs.
On Dyck-$(30,5)$, by contrast, the FFN gain extends to LSTM and Mamba2 as well, both of which improve markedly once an FFN is appended.
This task dependence is consistent with our opening observation that some tasks demand stronger (ii) than others: on \Localmaze, the gating and projections internal to LSTM and Mamba2 already supply enough feature-wise mixing, whereas Dyck benefits from the additional explicit mixer.

We also note that, in the design of the core, increasing representation power yields a meaningful gain (Table~\ref{tab:model_size}). 
While the parameter count does not change the model's complexity class in theory, our result \textit{does} show that representation power is an important factor for learning OOD-generalizable patterns. 
Thus, we can also interpret the strength of combining (i) and (ii) as efficient use of parameters along directions of representation power that serve OOD generalization. 
It is also worth re-emphasizing that, although Dyck-$(30,5)$ is known to be expressible by fixed-depth self-attention networks~\citep{yao2021bounded}, the Transformer itself fails to OOD-generalize on this task (Figure~\ref{fig:main-dyck-acc}), whereas $\core$s based on TF within the mechanism~\eqref{eq:fastslow}, including \akorn{}+FFN, succeed.
Together, these observations suggest that the mechanism~\eqref{eq:fastslow}, combined with sufficient representation power, is important for fully realizing the module's expressivity potential.

\section{Qualitative analysis: Emergent structure in latents}\label{sec:latent-analysis}

\begin{figure*}[tb]
  \centering
\begin{minipage}{0.99\textwidth}
\centering
\begin{subfigure}[b]{0.235\linewidth}
  \begin{tikzpicture}
    \node[inner sep=0pt] (pca) {\includegraphics[width=\linewidth]{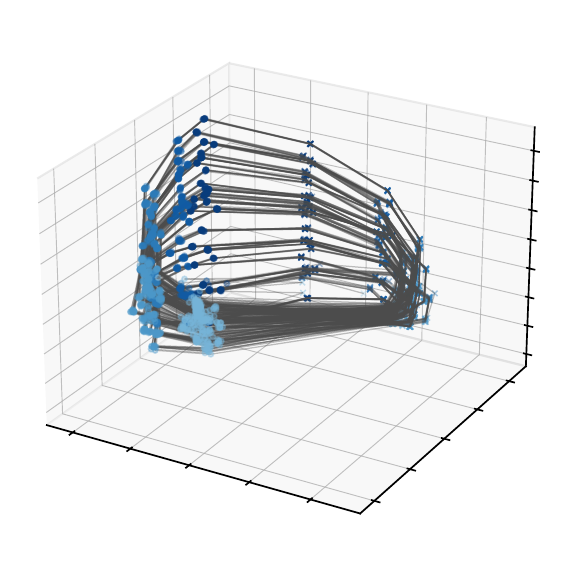}};
    \node[anchor=south east, fill=white, fill opacity=0.86, text opacity=1, inner sep=1.2pt, rounded corners=0.6pt, font=\scriptsize]
      at ([xshift=-2pt,yshift=2pt]pca.south east) {OOD Acc: 96.9\%};
  \end{tikzpicture}
  \caption{\kohei{\method{}-AKOrN+FFN}}
\end{subfigure}
\hfill
\begin{subfigure}[b]{0.235\linewidth}
  \begin{tikzpicture}
    \node[inner sep=0pt] (pca) {\includegraphics[width=\linewidth]{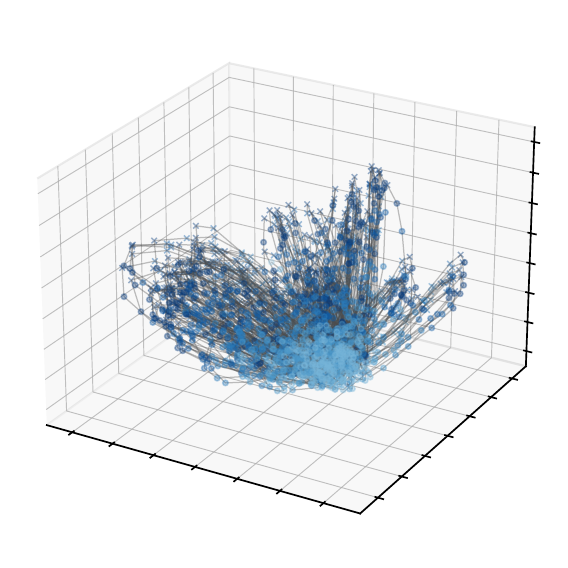}};
    \node[anchor=south east, fill=white, fill opacity=0.86, text opacity=1, inner sep=1.2pt, rounded corners=0.6pt, font=\scriptsize]
      at ([xshift=-2pt,yshift=2pt]pca.south east) {51.0\%};
  \end{tikzpicture}
  \caption{\kohei{\method{}-Mamba2}}
\end{subfigure}
\hfill
\begin{subfigure}[b]{0.235\linewidth}
  \begin{tikzpicture}
    \node[inner sep=0pt] (pca) {\includegraphics[width=\linewidth]{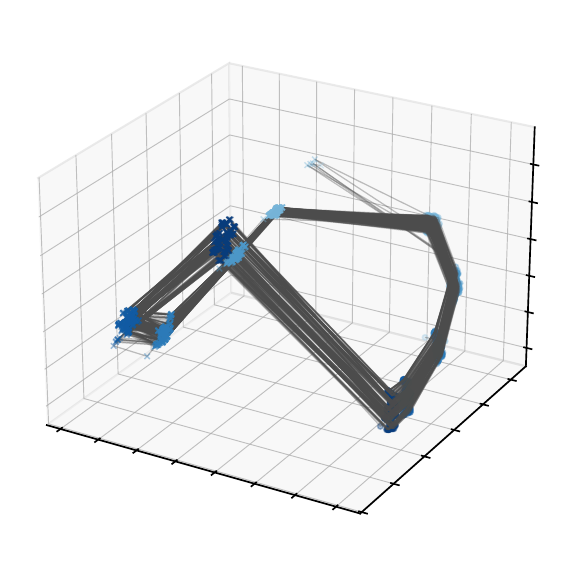}};
    \node[anchor=south east, fill=white, fill opacity=0.86, text opacity=1, inner sep=1.2pt, rounded corners=0.6pt, font=\scriptsize]
      at ([xshift=-2pt,yshift=2pt]pca.south east) {76.4\%};
  \end{tikzpicture}
  \caption{\kohei{\method{}-Attention}}
\end{subfigure}
\hfill
\begin{subfigure}[b]{0.235\linewidth}
  \begin{tikzpicture}
    \node[inner sep=0pt] (pca) {\includegraphics[width=\linewidth]{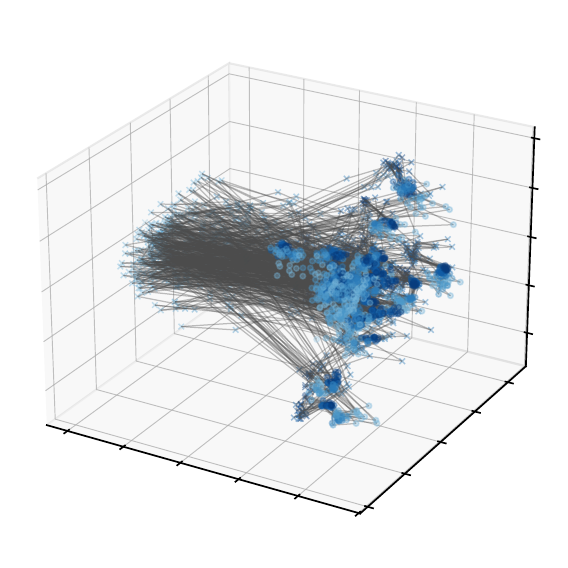}};
    \node[anchor=south east, fill=white, fill opacity=0.86, text opacity=1, inner sep=1.2pt, rounded corners=0.6pt, font=\scriptsize]
      at ([xshift=-2pt,yshift=2pt]pca.south east) {50.0\%};
  \end{tikzpicture}
  \caption{\kohei{Mamba2}}
\end{subfigure}
\end{minipage}
  \caption{PCA visualization of latent traversals. Each point corresponds to the latent $\lat_s(T)$ at stream position $s$; $\circ$ denotes an open bracket and $\times$ denotes a closing bracket.
  (a) \method{} with an AKOrN core and FFN forms an organized traversal that revisits task-aligned states across repeated stack updates.
  (b) \method{} with a Mamba2 core, (c) \method{} with an attention core, and (d) standalone Mamba2 do not exhibit the same coherent traversal under this visualization.}
  \label{fig:dyck-pca}
\end{figure*}

We next inspect whether the recurrent state organizes around variables that are relevant to the stream rule.
On Dyck-$(30,5)$, Figure~\ref{fig:dyck-pca} visualizes latent trajectories along a $5$-regular run, an OOD string with a repeated open--close pattern.
In the coherent configuration, \method{} with an AKOrN+FFN core forms an organized traversal that revisits task-aligned states across repeated stack updates, suggesting alignment with the stack dynamics of the Dyck language rule.
By contrast, the ablated \method{} variants without the same combination of persistent looping and task-appropriate feature-wise mixing, as well as standalone Mamba2 without the persistent fast--slow loop, produce less coherent trajectories.

These qualitative differences suggest that task-aligned latent organization is encouraged by the persistent loop together with an expressive recurrent core.
This Dyck interpretation is supported more directly by our \emph{stack-content probing} analysis: a linear readout of the $j$-th unresolved opener from frozen hidden states on OOD data shows that \method{} captures stack structure more faithfully than baselines lacking the fast--slow mechanism in \eqref{eq:fastslow}. See Appendix~\ref{app:stack_probe} and Figures~\ref{fig:app-dyck-core-latents} and~\ref{fig:app-dyck-baseline-latents} for details.
We observe an analogous organization in \Minigrid: DoorKey-16x16 latent trajectories cluster around task-relevant events across episodes; see \ref{app:latent_visualize} for the PCA visualization and accompanying energy trace.
These emergent structures are consistent with recent findings that looped attention-based computations converge to cyclic fixed points in latent space~\citep{blayney2026mechanistic}, and that continuous-time attention dynamics exhibit metastable clustering~\citep{geshkovski2024dynamic,geshkovski2025mathematical}.



\section{Conclusion and Limitations}

We studied a fast--slow recurrent formulation that combines persistent recurrence over a stream with various recurrent cores, and empirically found that cores with state-dependent transition and dimensional mixing are particularly strong for OOD generalization on streaming tasks.
By performing multiple latent updates per observation and propagating the resulting state without reset, our study adapts latent reasoning modules originally designed for static inputs to streaming settings, extending the benefits observed in prior work on static-input inference and better leveraging the core's expressivity. 

The present study also has several limitations. 
First, the proposed formulation trades parallelism 
for expressivity. Because the recurrent core is iterated multiple times per observation, inference 
is slower than highly optimized recurrent or state-space baselines, and our current implementation 
does not yet exploit specialized GPU kernels or recursion-aware parallelization.
Second, our experiments focus on settings where successful generalization can plausibly be supported by 
compact stream-level structure. The method should therefore not be interpreted as guaranteeing 
correct computation for arbitrary horizons, arbitrary distributions, or general long-context 
reasoning problems.
Finally, we fixed or controlled the number of inner-loop iterations in order to isolate the effect of the recurrent core. 
Combining persistent fast--slow recurrence with adaptive halting or learned 
compute allocation, and understanding when additional inner-loop computation should be spent, 
remain important directions for future work.

\begin{ack}
We thank Peyman Faratin, along with our collaborators and colleagues, for valuable feedback and discussions.
Takeru Miyato acknowledges his affiliation with the ELLIS (European Laboratory for Learning and Intelligent Systems) PhD program and is supported by the Google PhD Fellowship.
\end{ack}

\medskip


\bibliographystyle{biblio}
\bibliography{biblio}

\begin{thebibliography}{60}
\providecommand{\natexlab}[1]{#1}
\providecommand{\url}[1]{\texttt{#1}}
\expandafter\ifx\csname urlstyle\endcsname\relax
  \providecommand{\doi}[1]{doi: #1}\else
  \providecommand{\doi}{doi: \begingroup \urlstyle{rm}\Url}\fi

\bibitem[Amalric et~al.(2017)Amalric, Wang, Pica, Figueira, Sigman, and Dehaene]{amalric2017language}
Amalric, M., Wang, L., Pica, P., Figueira, S., Sigman, M., and Dehaene, S.
\newblock The language of geometry: {F}ast comprehension of geometrical primitives and rules in human adults and preschoolers.
\newblock \emph{PLOS Computational Biology}, 13\penalty0 (1):\penalty0 e1005273, 2017.
\newblock \doi{10.1371/journal.pcbi.1005273}.

\bibitem[Anil et~al.(2022)Anil, Wu, Andreassen, Lewkowycz, Misra, Ramasesh, Slone, Gur-Ari, Dyer, and Neyshabur]{anil2022exploring}
Anil, C., Wu, Y., Andreassen, A., Lewkowycz, A., Misra, V., Ramasesh, V., Slone, A., Gur-Ari, G., Dyer, E., and Neyshabur, B.
\newblock Exploring length generalization in large language models.
\newblock In \emph{Advances in Neural Information Processing Systems}, 2022.
\newblock URL \url{https://openreview.net/forum?id=zSkYVeX7bC4}.

\bibitem[Blayney et~al.(2026)Blayney, Arroyo, Obando-Ceron, Castro, Courville, Bronstein, and Dong]{blayney2026mechanistic}
Blayney, H., Arroyo, {\'A}., Obando-Ceron, J., Castro, P.~S., Courville, A., Bronstein, M.~M., and Dong, X.
\newblock A mechanistic analysis of looped reasoning language models.
\newblock \emph{arXiv preprint arXiv:2604.11791}, 2026.

\bibitem[Chae et~al.(2024)Chae, Kim, Kim, Ong, Kwak, Kim, Mac~Kim, Kwon, Chung, Yu, et~al.]{chae2024language}
Chae, H., Kim, Y., Kim, S., Ong, K. T.-i., Kwak, B.-w., Kim, M., Mac~Kim, S., Kwon, T., Chung, J., Yu, Y., et~al.
\newblock Language models as compilers: Simulating pseudocode execution improves algorithmic reasoning in language models.
\newblock In \emph{Proceedings of the 2024 Conference on Empirical Methods in Natural Language Processing}, pp.\  22471--22502, 2024.

\bibitem[Chen et~al.(2023)Chen, Wong, Chen, and Tian]{pe3chen2023extending}
Chen, S., Wong, S., Chen, L., and Tian, Y.
\newblock Extending context window of large language models via positional interpolation, 2023.

\bibitem[Chevalier-Boisvert et~al.(2023)Chevalier-Boisvert, Dai, Towers, de~Lazcano, Willems, Lahlou, Pal, Castro, and Terry]{10.5555/3666122.3669331}
Chevalier-Boisvert, M., Dai, B., Towers, M., de~Lazcano, R., Willems, L., Lahlou, S., Pal, S., Castro, P.~S., and Terry, J.
\newblock Minigrid \& miniworld: modular \& customizable reinforcement learning environments for goal-oriented tasks.
\newblock In \emph{Proceedings of the 37th International Conference on Neural Information Processing Systems}, NIPS '23, Red Hook, NY, USA, 2023. Curran Associates Inc.

\bibitem[Dai et~al.(2019)Dai, Yang, Yang, Carbonell, Le, and Salakhutdinov]{dai2019transformer}
Dai, Z., Yang, Z., Yang, Y., Carbonell, J., Le, Q.~V., and Salakhutdinov, R.
\newblock Transformer-xl: Attentive language models beyond a fixed-length context.
\newblock \emph{arXiv preprint arXiv:1901.02860}, 2019.

\bibitem[Dao \& Gu(2024)Dao and Gu]{dao2024transformers}
Dao, T. and Gu, A.
\newblock Transformers are ssms: Generalized models and efficient algorithms through structured state space duality.
\newblock \emph{arXiv preprint arXiv:2405.21060}, 2024.

\bibitem[Darlow et~al.(2025)Darlow, Regan, Risi, Seely, and Jones]{CTM}
Darlow, L., Regan, C., Risi, S., Seely, J., and Jones, L.
\newblock Continuous thought machines, 2025.
\newblock URL \url{https://arxiv.org/abs/2505.05522}.

\bibitem[{DeepSeek-AI} et~al.(2025){DeepSeek-AI}, Guo, Yang, Zhang, Song, Zhang, Xu, Zhu, Ma, Wang, Bi, Zhang, Yu, Wu, Wu, Gou, Shao, Li, Gao, et~al.]{deepseekai2025r1}
{DeepSeek-AI}, Guo, D., Yang, D., Zhang, H., Song, J., Zhang, R., Xu, R., Zhu, Q., Ma, S., Wang, P., Bi, X., Zhang, X., Yu, X., Wu, Y., Wu, Z.~F., Gou, Z., Shao, Z., Li, Z., Gao, Z., et~al.
\newblock {DeepSeek-R1}: Incentivizing reasoning capability in {LLM}s via reinforcement learning, 2025.

\bibitem[Dehaene et~al.(2022)Dehaene, Al~Roumi, Lakretz, Planton, and Sabl{\'e}-Meyer]{dehaene2022symbols}
Dehaene, S., Al~Roumi, F., Lakretz, Y., Planton, S., and Sabl{\'e}-Meyer, M.
\newblock Symbols and mental programs: a hypothesis about human singularity.
\newblock \emph{Trends in Cognitive Sciences}, 26\penalty0 (9):\penalty0 751--766, 2022.
\newblock \doi{10.1016/j.tics.2022.06.010}.

\bibitem[Dehghani et~al.(2019)Dehghani, Gouws, Vinyals, Uszkoreit, and Kaiser]{UT}
Dehghani, M., Gouws, S., Vinyals, O., Uszkoreit, J., and Kaiser, L.
\newblock Universal transformers.
\newblock In \emph{International Conference on Learning Representations (ICLR)}, 2019.
\newblock URL \url{https://openreview.net/forum?id=HyzdRiR9Y7}.
\newblock OpenReview preprint.

\bibitem[Du et~al.(2025)Du, Huang, Wang, and Wang]{du2025longshort}
Du, T., Huang, H., Wang, Y., and Wang, Y.
\newblock Long-short alignment for effective long-context modeling in {LLM}s.
\newblock In \emph{Proceedings of the 42nd International Conference on Machine Learning}, volume 267 of \emph{Proceedings of Machine Learning Research}, pp.\  14515--14535. PMLR, 2025.
\newblock URL \url{https://proceedings.mlr.press/v267/du25a.html}.

\bibitem[Fan et~al.(2025)Fan, Du, Ramchandran, and Lee]{Looped}
Fan, Y., Du, Y., Ramchandran, K., and Lee, K.
\newblock Looped transformers for length generalization.
\newblock In \emph{International Conference on Learning Representations (ICLR)}, 2025.
\newblock URL \url{https://arxiv.org/abs/2409.15647}.

\bibitem[Geiping et~al.(2025)Geiping, McLeish, Jain, Kirchenbauer, Singh, Bartoldson, Kailkhura, Bhatele, and Goldstein]{Geiping2025LatentReasoning}
Geiping, J., McLeish, S.~M., Jain, N., Kirchenbauer, J., Singh, S., Bartoldson, B.~R., Kailkhura, B., Bhatele, A., and Goldstein, T.
\newblock Scaling up test-time compute with latent reasoning: A recurrent depth approach.
\newblock In \emph{ES-FoMo III Spotlight (ICML 2025 Workshop)}, June 2025.
\newblock URL \url{https://openreview.net/forum?id=D6o6Bwtq7h}.
\newblock OpenReview preprint / workshop version.

\bibitem[Geshkovski et~al.(2024)Geshkovski, Koubbi, Polyanskiy, and Rigollet]{geshkovski2024dynamic}
Geshkovski, B., Koubbi, H., Polyanskiy, Y., and Rigollet, P.
\newblock Dynamic metastability in the self-attention model.
\newblock \emph{arXiv preprint arXiv:2410.06833}, 2024.
\newblock URL \url{https://arxiv.org/pdf/2410.06833}.

\bibitem[Geshkovski et~al.(2025)Geshkovski, Letrouit, Polyanskiy, and Rigollet]{geshkovski2025mathematical}
Geshkovski, B., Letrouit, C., Polyanskiy, Y., and Rigollet, P.
\newblock A mathematical perspective on transformers.
\newblock \emph{Bulletin of the American Mathematical Society}, 62\penalty0 (3):\penalty0 427--479, 2025.

\bibitem[Giannou et~al.(2023)Giannou, Rajput, Sohn, Lee, Lee, and Papailiopoulos]{giannou2023programmable}
Giannou, A., Rajput, S., Sohn, J.-y., Lee, K., Lee, J.~D., and Papailiopoulos, D.
\newblock Looped transformers as programmable computers.
\newblock In \emph{International Conference on Machine Learning}, pp.\  11398--11442, 2023.
\newblock arXiv:2301.13196.

\bibitem[Gladstone et~al.(2025)Gladstone, Nanduru, Islam, Han, Ha, Chadha, Du, Ji, Li, and Iqbal]{EBT}
Gladstone, A., Nanduru, G., Islam, M.~M., Han, P., Ha, H., Chadha, A., Du, Y., Ji, H., Li, J., and Iqbal, T.
\newblock Energy-based transformers are scalable learners and thinkers.
\newblock \emph{arXiv preprint arXiv:2507.02092}, 2025.
\newblock URL \url{https://arxiv.org/abs/2507.02092}.

\bibitem[Graves(2017)]{ACT}
Graves, A.
\newblock Adaptive computation time for recurrent neural networks, 2017.
\newblock URL \url{https://arxiv.org/abs/1603.08983}.

\bibitem[Gu \& Dao(2023)Gu and Dao]{gu2023mamba}
Gu, A. and Dao, T.
\newblock Mamba: Linear-time sequence modeling with selective state spaces.
\newblock \emph{arXiv preprint arXiv:2312.00752}, 2023.

\bibitem[Hao et~al.(2024)Hao, Sukhbaatar, Su, Li, Hu, Weston, and Tian]{hao2024training}
Hao, S., Sukhbaatar, S., Su, D., Li, X., Hu, Z., Weston, J., and Tian, Y.
\newblock Training large language models to reason in a continuous latent space.
\newblock \emph{arXiv preprint arXiv:2412.06769}, 2024.
\newblock URL \url{https://arxiv.org/abs/2412.06769}.

\bibitem[Hewitt et~al.(2020)Hewitt, Hahn, Ganguli, Liang, and Manning]{hewitt2020rnns}
Hewitt, J., Hahn, M., Ganguli, S., Liang, P., and Manning, C.~D.
\newblock Rnns can generate bounded hierarchical languages with optimal memory.
\newblock \emph{arXiv preprint arXiv:2010.07515}, 2020.

\bibitem[Hochreiter \& Schmidhuber(1997)Hochreiter and Schmidhuber]{LSTM}
Hochreiter, S. and Schmidhuber, J.
\newblock Long short-term memory.
\newblock \emph{Neural computation}, 9\penalty0 (8):\penalty0 1735--1780, 1997.

\bibitem[Huntenburg et~al.(2018)Huntenburg, Bazin, and Margulies]{bio2}
Huntenburg, J.~M., Bazin, P.-L., and Margulies, D.~S.
\newblock Large-scale gradients in human cortical organization.
\newblock \emph{Trends in Cognitive Sciences}, 22\penalty0 (1):\penalty0 21--31, 2018.
\newblock \doi{10.1016/j.tics.2017.11.002}.

\bibitem[Jolicoeur-Martineau(2025)]{TRM}
Jolicoeur-Martineau, A.
\newblock Less is more: Recursive reasoning with tiny networks.
\newblock \emph{arXiv preprint}, arXiv:2510.04871, 2025.
\newblock \doi{10.48550/arXiv.2510.04871}.
\newblock URL \url{https://arxiv.org/abs/2510.04871}.

\bibitem[Koutn{\'i}k et~al.(2014)Koutn{\'i}k, Greff, Gomez, and Schmidhuber]{koutnik2014clockwork}
Koutn{\'i}k, J., Greff, K., Gomez, F., and Schmidhuber, J.
\newblock A clockwork {RNN}.
\newblock In \emph{Proceedings of the 31st International Conference on Machine Learning (ICML)}, 2014.

\bibitem[Liu et~al.(2023)Liu, Ash, Goel, Krishnamurthy, and Zhang]{liu2023transformers}
Liu, B., Ash, J.~T., Goel, S., Krishnamurthy, A., and Zhang, C.
\newblock Transformers learn shortcuts to automata.
\newblock In \emph{The Eleventh International Conference on Learning Representations}, 2023.
\newblock URL \url{https://openreview.net/forum?id=De4FYqjFueZ}.

\bibitem[McLeish et~al.(2025)McLeish, Li, Kirchenbauer, Kalra, Bartoldson, Kailkhura, Schwarzschild, Geiping, Goldstein, and Goldblum]{mcleish2025retrofit}
McLeish, S., Li, A., Kirchenbauer, J., Kalra, D.~S., Bartoldson, B.~R., Kailkhura, B., Schwarzschild, A., Geiping, J., Goldstein, T., and Goldblum, M.
\newblock Teaching pretrained language models to think deeper with retrofitted recurrence.
\newblock \emph{arXiv preprint arXiv:2511.07384}, 2025.

\bibitem[Merrill \& Sabharwal(2023)Merrill and Sabharwal]{merrill2023parallelism}
Merrill, W. and Sabharwal, A.
\newblock The parallelism tradeoff: Limitations of log-precision transformers.
\newblock In \emph{Transactions of the Association for Computational Linguistics}, volume~11, pp.\  531--545, 2023.
\newblock \doi{10.1162/tacl_a_00562}.

\bibitem[Merrill et~al.(2024)Merrill, Petty, and Sabharwal]{merrill2024illusion}
Merrill, W., Petty, J., and Sabharwal, A.
\newblock The illusion of state in state-space models.
\newblock In \emph{International Conference on Machine Learning}, 2024.
\newblock arXiv:2404.08819.

\bibitem[Miyato et~al.(2023)Miyato, Jaeger, Welling, and Geiger]{miyato2023gta}
Miyato, T., Jaeger, B., Welling, M., and Geiger, A.
\newblock Gta: A geometry-aware attention mechanism for multi-view transformers.
\newblock \emph{arXiv preprint arXiv:2310.10375}, 2023.

\bibitem[Miyato et~al.(2025)Miyato, Löwe, Geiger, and Welling]{Akorn}
Miyato, T., Löwe, S., Geiger, A., and Welling, M.
\newblock Artificial kuramoto oscillatory neurons.
\newblock In \emph{International Conference on Learning Representations (ICLR)}, 2025.
\newblock URL \url{https://openreview.net/forum?id=nwDRD4AMoN}.

\bibitem[Mujika et~al.(2017)Mujika, Meier, and Steger]{mujika2017fastslow}
Mujika, A., Meier, F., and Steger, A.
\newblock Fast-slow recurrent neural networks.
\newblock In \emph{Advances in Neural Information Processing Systems}, volume~30. Curran Associates, Inc., 2017.
\newblock URL \url{https://proceedings.neurips.cc/paper/2017/hash/3dd48ab31d016ffcbf3314df2b3cb9ce-Abstract.html}.

\bibitem[Murray et~al.(2014)Murray, Bernacchia, Freedman, Romo, Wallis, Cai, Padoa-Schioppa, Pasternak, Seo, Lee, and Wang]{bio0}
Murray, J.~D., Bernacchia, A., Freedman, D.~J., Romo, R., Wallis, J.~D., Cai, X., Padoa-Schioppa, C., Pasternak, T., Seo, H., Lee, D., and Wang, X.-J.
\newblock A hierarchy of intrinsic timescales across primate cortex.
\newblock \emph{Nature Neuroscience}, 17\penalty0 (12):\penalty0 1661--1663, 2014.
\newblock \doi{10.1038/nn.3862}.

\bibitem[{OpenAI}(2024)]{openai2024o1}
{OpenAI}.
\newblock New reasoning models: {OpenAI} o1-preview and o1-mini.
\newblock \url{https://openai.com/research/o1-preview-and-o1-mini}, 2024.
\newblock Accessed via Geiping et al.\ (2025), arXiv:2502.05171.

\bibitem[Pascanu et~al.(2014)Pascanu, G{\"u}l{\c{c}}ehre, Cho, and Bengio]{pascanu2014deep}
Pascanu, R., G{\"u}l{\c{c}}ehre, {\c{C}}., Cho, K., and Bengio, Y.
\newblock How to construct deep recurrent neural networks.
\newblock In \emph{International Conference on Learning Representations (ICLR)}, 2014.
\newblock URL \url{https://arxiv.org/abs/1312.6026}.

\bibitem[Peng et~al.(2024)Peng, Quesnelle, Fan, and Shippole]{peng2024yarn}
Peng, B., Quesnelle, J., Fan, H., and Shippole, E.
\newblock Ya{RN}: Efficient context window extension of large language models.
\newblock In \emph{The Twelfth International Conference on Learning Representations (ICLR)}, 2024.
\newblock URL \url{https://openreview.net/forum?id=wHBfxhZu1u}.

\bibitem[Press et~al.(2022)Press, Smith, and Lewis]{pe2press2022train}
Press, O., Smith, N.~A., and Lewis, M.
\newblock Train short, test long: Attention with linear biases enables input length extrapolation.
\newblock In \emph{International Conference on Learning Representations (ICLR)}, 2022.
\newblock URL \url{https://arxiv.org/abs/2108.12409}.

\bibitem[Saunshi et~al.(2025)Saunshi, Dikkala, Li, Kumar, and Reddi]{saunshi2025latent}
Saunshi, N., Dikkala, N., Li, Z., Kumar, S., and Reddi, S.~J.
\newblock Reasoning with latent thoughts: On the power of looped transformers.
\newblock In \emph{International Conference on Learning Representations}, 2025.
\newblock arXiv:2502.17416.

\bibitem[Schmidhuber(1992)]{schmidhuber1992history}
Schmidhuber, J.
\newblock Learning complex, extended sequences using the principle of history compression.
\newblock \emph{Neural Computation}, 4\penalty0 (2):\penalty0 234--242, 1992.
\newblock \doi{10.1162/neco.1992.4.2.234}.

\bibitem[Sch{\"o}ne et~al.(2025)Sch{\"o}ne, Rahmani, Kremer, Falck, Ballani, and Gladrow]{schone2025implicit}
Sch{\"o}ne, M., Rahmani, B., Kremer, H., Falck, F., Ballani, H., and Gladrow, J.
\newblock Implicit language models are {RNN}s: Balancing parallelization and expressivity.
\newblock In \emph{Proceedings of the 42nd International Conference on Machine Learning}, volume 267 of \emph{Proceedings of Machine Learning Research}. PMLR, 2025.

\bibitem[Sch{\"u}tzenberger(1963)]{dyckpaper}
Sch{\"u}tzenberger, M.~P.
\newblock On context-free languages and push-down automata.
\newblock \emph{Information and Control}, 6\penalty0 (3):\penalty0 246--264, 1963.
\newblock ISSN 0019-9958.
\newblock \doi{10.1016/S0019-9958(63)90306-1}.
\newblock URL \url{https://www.sciencedirect.com/science/article/pii/S0019995863903061}.

\bibitem[Siems et~al.(2025)Siems, Carstensen, Zela, Hutter, Pontil, and Grazzi]{siems2025deltaproduct}
Siems, J., Carstensen, T., Zela, A., Hutter, F., Pontil, M., and Grazzi, R.
\newblock {DeltaProduct}: Improving state-tracking in linear {RNN}s via householder products.
\newblock In \emph{Advances in Neural Information Processing Systems}, 2025.
\newblock arXiv:2502.10297.

\bibitem[Smith et~al.(2022)Smith, Warrington, and Linderman]{smith2022simplified}
Smith, J. T.~H., Warrington, A., and Linderman, S.~W.
\newblock Simplified state space layers for sequence modeling.
\newblock \emph{arXiv preprint arXiv:2208.04933}, 2022.

\bibitem[Su et~al.(2021)Su, Lu, Pan, Murtadha, Wen, and Liu]{pe1su2021roformer}
Su, J., Lu, Y., Pan, S., Murtadha, A., Wen, B., and Liu, Y.
\newblock Roformer: Enhanced transformer with rotary position embedding, 2021.

\bibitem[Tolstikhin et~al.(2021)Tolstikhin, Houlsby, Kolesnikov, Beyer, Zhai, Unterthiner, Yung, Steiner, Keysers, Uszkoreit, Lucic, and Dosovitskiy]{tolstikhin2021mlpmixer}
Tolstikhin, I., Houlsby, N., Kolesnikov, A., Beyer, L., Zhai, X., Unterthiner, T., Yung, J., Steiner, A., Keysers, D., Uszkoreit, J., Lucic, M., and Dosovitskiy, A.
\newblock {MLP-Mixer}: An all-{MLP} architecture for vision.
\newblock In \emph{Advances in Neural Information Processing Systems}, volume~34. Curran Associates, Inc., 2021.
\newblock URL \url{https://proceedings.neurips.cc/paper/2021/hash/cba0a4ee5ccd02fda0fe3f9a3e7b89fe-Abstract.html}.

\bibitem[Vaswani et~al.(2017)Vaswani, Shazeer, Parmar, Uszkoreit, Jones, Gomez, Kaiser, and Polosukhin]{vaswani2017attention}
Vaswani, A., Shazeer, N., Parmar, N., Uszkoreit, J., Jones, L., Gomez, A.~N., Kaiser, {\L}., and Polosukhin, I.
\newblock Attention is all you need.
\newblock \emph{Advances in neural information processing systems}, 30, 2017.

\bibitem[Wang et~al.(2025)Wang, Li, Sun, Chen, Liu, Wu, Lu, Song, and Yadkori]{HRM}
Wang, G., Li, J., Sun, Y., Chen, X., Liu, C., Wu, Y., Lu, M., Song, S., and Yadkori, Y.~A.
\newblock Hierarchical reasoning model.
\newblock \emph{arXiv preprint arXiv:2506.21734}, 2025.
\newblock URL \url{https://arxiv.org/abs/2506.21734}.

\bibitem[Wang et~al.(2024)Wang, Kobyzev, Lu, Rezagholizadeh, and Liu]{wang2024resonance}
Wang, S., Kobyzev, I., Lu, P., Rezagholizadeh, M., and Liu, B.
\newblock Resonance {R}o{PE}: Improving context length generalization of large language models.
\newblock In \emph{Findings of the Association for Computational Linguistics: ACL 2024}, pp.\  586--598, Bangkok, Thailand, August 2024. Association for Computational Linguistics.
\newblock \doi{10.18653/v1/2024.findings-acl.32}.
\newblock URL \url{https://aclanthology.org/2024.findings-acl.32/}.

\bibitem[Wang et~al.(2022)Wang, Jiang, Zhu, Yuan, Song, and Liu]{adapt1_wang2022dynamixer}
Wang, Z., Jiang, W., Zhu, Y., Yuan, L., Song, Y., and Liu, W.
\newblock {DynaMixer}: A vision {MLP} architecture with dynamic mixing.
\newblock In \emph{Proceedings of the 39th International Conference on Machine Learning}, volume 162 of \emph{Proceedings of Machine Learning Research}, pp.\  22691--22701. PMLR, 2022.

\bibitem[Wei et~al.(2023{\natexlab{a}})Wei, Zhang, Lan, Lu, and Chen]{adapt2_wei2023active}
Wei, G., Zhang, Z., Lan, C., Lu, Y., and Chen, Z.
\newblock Active token mixer.
\newblock In \emph{Proceedings of the AAAI Conference on Artificial Intelligence}, volume~37, pp.\  2759--2767, 2023{\natexlab{a}}.
\newblock \doi{10.1609/aaai.v37i3.25376}.

\bibitem[Wei et~al.(2023{\natexlab{b}})Wei, Zhang, Lan, Lu, and Chen]{adapt3_wei2023active}
Wei, G., Zhang, Z., Lan, C., Lu, Y., and Chen, Z.
\newblock Active token mixer.
\newblock In \emph{Proceedings of the AAAI Conference on Artificial Intelligence}, volume~37, pp.\  2759--2767, 2023{\natexlab{b}}.
\newblock \doi{10.1609/aaai.v37i3.25376}.

\bibitem[Xu \& Sato(2025)Xu and Sato]{xu2025expressive}
Xu, K. and Sato, I.
\newblock On expressive power of looped transformers: Theoretical analysis and enhancement via timestep encoding.
\newblock In \emph{International Conference on Machine Learning}, 2025.
\newblock arXiv:2410.01405.

\bibitem[Yang et~al.(2023)Yang, Yu, Zhu, and Hayou]{yang2023tensor}
Yang, G., Yu, D., Zhu, C., and Hayou, S.
\newblock Tensor programs vi: Feature learning in infinite-depth neural networks.
\newblock \emph{arXiv preprint arXiv:2310.02244}, 2023.

\bibitem[Yang et~al.(2024)Yang, Lee, Nowak, and Papailiopoulos]{yang2024looped}
Yang, L., Lee, K., Nowak, R., and Papailiopoulos, D.
\newblock Looped transformers are better at learning learning algorithms.
\newblock In \emph{International Conference on Learning Representations}, 2024.
\newblock arXiv:2311.12424.

\bibitem[Yao et~al.(2021)Yao, Peng, Papadimitriou, and Narasimhan]{yao2021bounded}
Yao, S., Peng, B., Papadimitriou, C., and Narasimhan, K.
\newblock Self-attention networks can process bounded hierarchical languages.
\newblock In \emph{Proceedings of the 59th Annual Meeting of the Association for Computational Linguistics and the 11th International Joint Conference on Natural Language Processing (ACL-IJCNLP)}, pp.\  3770--3785, 2021.
\newblock \doi{10.18653/v1/2021.acl-long.292}.
\newblock URL \url{https://aclanthology.org/2021.acl-long.292/}.

\bibitem[Zeraati et~al.(2023)Zeraati, Shi, Steinmetz, Gieselmann, Thiele, Moore, Levina, and Engel]{bio1}
Zeraati, R., Shi, Y.-L., Steinmetz, N.~A., Gieselmann, M.~A., Thiele, A., Moore, T., Levina, A., and Engel, T.~A.
\newblock Intrinsic timescales in the visual cortex change with selective attention and reflect spatial connectivity.
\newblock \emph{Nature Communications}, 14\penalty0 (1):\penalty0 1858, 2023.
\newblock \doi{10.1038/s41467-023-37613-7}.

\bibitem[Zhou et~al.(2024)Zhou, Alon, Chen, Wang, Agarwal, and Zhou]{zhou2024transformers}
Zhou, Y., Alon, U., Chen, X., Wang, X., Agarwal, R., and Zhou, D.
\newblock Transformers can achieve length generalization but not robustly.
\newblock \emph{arXiv preprint arXiv:2402.09371}, 2024.
\newblock \doi{10.48550/arXiv.2402.09371}.
\newblock URL \url{https://arxiv.org/abs/2402.09371}.

\bibitem[Zhu et~al.(2025)Zhu, Wang, Hua, Zhang, Li, Que, Wei, Wen, Yin, Xing, et~al.]{ouro2025}
Zhu, R.-J., Wang, Z., Hua, K., Zhang, T., Li, Z., Que, H., Wei, B., Wen, Z., Yin, F., Xing, H., et~al.
\newblock Scaling latent reasoning via looped language models.
\newblock \emph{arXiv preprint arXiv:2510.25741}, 2025.
\newblock URL \url{https://arxiv.org/abs/2510.25741}.

\end{thebibliography}


\appendix


\section{Model Details}

\paragraph{Terminology.}
We retain the term ``fast--slow'' in line with the multiscale RNN tradition \citep{schmidhuber1992history, koutnik2014clockwork}.
The unrelated FS-RNN of \citet{mujika2017fastslow} uses the same name for a wiring of distinct LSTM cells within one observation step, without an inner clock or weight sharing.

\subsection{\texorpdfstring{Implementation of $\core$}{Implementation of R}} \label{app:core}

\subsubsection{\akorn} 
\label{app:impl-J}
For the choice of $\core$ in our experiments, a mechanism that performed particularly stably was a variation of the updates used by \citet{Akorn}, although $\core$ implemented with a vanilla Transformer block performed competitively as well.
See Figure~\ref{fig:code-J} for the design of $R_{\akorn}$ we used in our experiments. 
This module has two additional components compared with $R_{TF}$, and we would like to refer readers to the original paper for more details. 
First, $\{ \Omega_i ; i \in 1:\numtoken \} $ is an intrinsic frequency term represented by anti-symmetric matrices, intended to provide each token in $S^{\latd-1} \subset \RR^{\latd}$ with rotational momentum. 
Second, there is a linear $\Proj_\lat$ operator that projects a given vector to the tangent space of $S^{\latd-1}$ at $\lat$.   

\begin{figure}[ht!]
  \centering
    \begin{lstlisting}[language=python]
class R():
    def __init__(dim, oscillator_dim, init_gamma):
        self.attn = SelfAttention(dim)
        self.mlp = MLP(dim)
        self.omega = Omega(dim, oscillator_dim)
        self.gamma = nn.Parameter(init_gamma, require_grad=True)
        
    def forward(state, cond):
        y = self.attn(state + cond)
        y = self.mlp(state + cond + y)
        y = proj(y, state)  # projection onto the tangent space
        y = y + self.omega(state)
        state += self.gamma * y
        state /= norm(state)
        return state
    \end{lstlisting}
  \caption{Pseudocode of our $\core_{\akorn}$. }
  \label{fig:code-J}
  \vspace{-5pt}
\end{figure}

We note that the original implementation of \citet{Akorn} does not introduce an MLP, and that it introduces $\bC$ additively outside $\mathbf{Attn}$, which consists of the usual procedure of
\begin{align}
Attn(\lat) &= \mathbf{softmax} \left( \frac{\lat W_Q W_K^{\rm T} \lat^{\rm T}}{\sqrt{\latd'}} \right)\lat W_V  \\ 
&:= \mathbf{AttnMat}(\lat) \lat W_V  
\end{align}
with $W_Q, W_K, W_V \in \RR^{\latd \times \latd'}$, $\lat \in \RR^{\numtoken \times \latd}$.  
In implementation, we used the multihead version of this.
We used GTA~\citep{miyato2023gta} as positional embeddings. Pseudocode is provided in Figure~\ref{fig:code-J}.

\paragraph{Energy-like Scalar} \label{sec:energy}

In the absence of the MLP term, in a form closer to the attentive connectivity of \citet{Akorn}, \akorn admits an update of the form
\begin{align}
\lat_i = \Omega_i \lat_i +  \Proj_{\lat_i} \left(\sum_{j}[\mathbf{AttnMat}(\lat)]_{ij}W_V^{\rm T} \lat_j    + \bC_i \right)
\end{align}
where $\mathbf{AttnMat}(\lat):= J$ is an attention matrix from $\lat \in \RR^{\numtoken \times \latd}$, and we introduce a pseudo-energy of the following form:
\begin{align}
\cE(\lat | \bC )  =  - \frac{1}{2} \sum_{i,j} J_{ij} \lat_i^{\rm T} \lat_j    -  \sum_i \bC_i^{\rm T} \lat_i
\label{eq:energy}
\end{align}
in close relation to the previously introduced energy of \citet{geshkovski2025mathematical}.
Because we incorporate $\bC$ additively before the attention as in Figure~\ref{fig:code-J}, however, we use the same energy as $\cE(\lat, \bC)$ in \eqref{eq:energy} to assist the visualization of the latent for our construction of $\core_{\akorn}$, with $A$ and $B$ being the gating functions that depend on $\lat$ and $\bC$.


\subsubsection{Mamba2 and LSTM Core} 
\label{app:mamba_core_details}

We implemented the Mamba2 and LSTM cores within the \method{} loop so that
the conditions in Section~\ref{sec:strong_core} are better satisfied.
In designing these cores, we used an additional inner loop in which these mechanisms have their own latents.
Namely, in the presence of input $\bC_s \in \RR^{\ell \times \latd}$, the 
latent $\lat_s(t) \in \RR^{\chunk \times \latd}$ for these cores evolves as  
\begin{equation}
\begin{aligned}
\lat_s(t+1) = \core\left(\lat_s(t),\, \bC_s \right), 
\qquad \lat_{s+1}(0) = \lat_s(T),
\qquad t = 0, \ldots, T, 
\end{aligned}
\end{equation}
where
\begin{align}
\core\left(\lat_s(t),\, \bC_s \right) &=  M  [h_0,  \dots h_\chunk] 
\end{align}
with mixing $M \in  R^{L \times \latd}$ in \textit{token} direction, and 
$h$ evolving on its own inner-inner loop, which, for example, for Mamba, takes the form of
\begin{align}
h_{\ell +1} = A([\lat_s(t) + \bC_s]) h_\ell + B([\lat_s(t)+ \bC_s])[\lat_s(t) + \bC_s],   ~~~~~ \ell = 0, \dots L-1
\end{align}
with $h_0$ being initialized to a constant at every observation step $s$. 
In this design of $\core$ itself, the $\chunk$-size loop features both latent-dependent transition and dimensional mixing.
Our LSTM core is also defined analogously with its own \textit{cell} latent and \textit{hidden state} latent. 
The Mamba/LSTM + FFN variants add a separate MLP along the $d$ dimension on top of the core described here, providing additional dimensional/channel mixing beyond the native gates and projections.


\begin{table}[tb]
  \centering
  \caption{Default hyperparameters for Maze (supervised), Dyck (supervised), and \Minigrid (RL/PPO). We use these settings for all baselines unless otherwise specified.}
  \label{tab:hparams-merged}
  \setlength{\tabcolsep}{10pt}
  \renewcommand{\arraystretch}{1.2}
  \begin{tabular}{lccc}
    \toprule
    \textbf{Parameter} & \textbf{Maze} & \textbf{Dyck} & \textbf{\Minigrid} \\
    \midrule
    Optimizer                         & AdamW                & AdamW               & Adam \\
    Batch size                        & 256                  & 256                 & 3200 \\
    Training epochs                   & 300                  & 30                  & — \\
    Weight decay                      & 0.1                  & $0.01$  & — \\
    Gradient clip                     & 0.1                  & 1.0                 & 0.5 \\
    Scheduler                         & cosine               & cosine              & — \\
    Dataset size (train/val)          & 45{,}000 / 5{,}000   & 10{,}000 / 1{,}000  & — \\
    \midrule
    Bracket types ($k$)               & —                    & 30                  & — \\
    Training depth ($m$)              & —                    & 5                   & — \\
    Sequence length (train)           & —                    & $10$--$40$ tokens   & — \\
    \midrule
    Total timesteps                   & —                    & —                   & $1.0\times10^{6}$ \\
    \# Environments                   & —                    & —                   & 32 \\
    \# Steps / env                    & —                    & —                   & 96 \\
    Update epochs                     & —                    & —                   & 4 \\
    Discount factor ($\gamma$)        & —                    & —                   & 0.995 \\
    GAE $\lambda$                     & —                    & —                   & 0.95 \\
    Value loss coef.\ ($c_v$)         & —                    & —                   & 0.5 \\
    Clip coefficient (PPO ratio)      & —                    & —                   & 0.1 \\
    \bottomrule
  \end{tabular}
\end{table}

\begin{table}[ht]
  \centering
  \caption{Model-specific hyperparameters for Maze/Dyck/\Minigrid.
Parameter counts and key architectural/training choices for each baseline and our method. When three numbers are shown, they correspond to Maze/Dyck/\Minigrid respectively.}
  \label{tab:model-hparams-maze-minigrid}
  \setlength{\tabcolsep}{5pt}
  \renewcommand{\arraystretch}{1.2}
  \begin{tabular}{lcccc}
    \toprule
     & \textbf{LSTM} & \textbf{Mamba} & \textbf{Transformer} & \textbf{FSRM} \\
    \midrule
    Parameters (M)   & 2.86/37.1/2.47 & 1.25/16.5/2.81 & 1.73/21.3/3.30 & 1.16/1.41/1.19 \\
    \midrule
    Hidden dim                & 128/512/512  & 128/256/380  & 128/256/384  & 64/256/512 \\
    Entropy coef.             & ---/---/1e-4 & ---/---/1e-2 & ---/---/1e-2 & ---/---/1e-2 \\
    RNN hidden dim            & 256/512/256          & ---  & ---  & --- \\
    Transformer heads         & ---     & ---        & 8/8/4    & 4 \\
    Memory len.               & ---     & ---        & ---/---/119  & --- \\
    $d_{\text{state}}$        & ---     & 64         & ---  & --- \\
    $d_{\text{conv}}$         & ---     & 4          & ---  & --- \\
    Expand ratio              & ---     & 2          & --- & ---  \\
    Depth                     & 4/2/1 & 6/4/1    & 6/4/3  & 1/2/1\\
    Oscillator dim            & ---     & ---        & --- & 4/4/2 \\
    Internal Steps            & ---     & ---        & --- & 5/5/10 \\
    Initial $\gamma$                  & ---     & ---        & --- & 0.1 \\
    Initial $\Omega$                & ---     & ---        & --- & 0.1 \\
    Chunk size                    & ---     & ---        & --- & 16/---/--- \\
    Learning rate             & 1e-3/5e-3/2.5e-4 & 1e-3/5e-4/1.5e-4 & 1e-3/5e-3/2.5e-4 & 1e-3/5e-3/2.5e-4 \\
    \bottomrule
  \end{tabular}
\end{table}

\section{Experiment Details}

\subsection{Baselines and Training Protocol}
\label{app:baseline-details}

We use a task-matched baseline set because the three benchmarks differ in supervision and interaction.
For Dyck, the controlled supervised comparisons are LSTM~\citep{LSTM}, Mamba2~\citep{gu2023mamba,dao2024transformers}, and Transformer~\citep{vaswani2017attention}, trained on the same Dyck-$(30,5)$ split as \method{}.
The LLM comparisons in Figure~\ref{fig:dyck-llms} are context-setting stress tests: Gemini 3.0 Pro, Claude Opus 4.5, and GPT-5.1 Thinking are prompted with the stack algorithm, while Qwen3-4B-Base is LoRA fine-tuned on the same short split.
For \Localmaze, we compare LSTM, Mamba2, TF, Looped TF~\citep{Looped}, DeltaProduct~\citep{siems2025deltaproduct}, and \method{}, with S5 and CTM included in the extended appendix comparison.
For \Minigrid, all methods use PPO and the same ID/OOD environment splits; the sequence-model baselines are LSTM, Mamba2, and TransformerXL~\citep{dai2019transformer}.
Tables~\ref{tab:hparams-merged} and~\ref{tab:model-hparams-maze-minigrid} give the shared training settings and model-specific hyperparameters.

\subsection{Hyperparameters}\label{app:hyperparams}

We report the main hyperparameters of the models in Tables~\ref{tab:hparams-merged} and~\ref{tab:model-hparams-maze-minigrid}.
For \method{}, the model-specific oscillator dimension and number of internal steps are task-dependent and are reported in Table~\ref{tab:model-hparams-maze-minigrid} in Maze/Dyck/\Minigrid order.
In particular, the \Minigrid runs use oscillator dimension $2$ and $T=10$, while the supervised Maze and Dyck runs use $T=5$.
For Dyck, we grid-search the learning rate over $\{10^{-4}, 5\times10^{-4}, 10^{-3}, 5\times10^{-3}\}$ per model and select by ID accuracy at lengths 1280--2560.
For the Dyck core ablation in Appendix~\ref{app:dyck-core-ablation}, we use a grid of $\{1,2,3,4,5\}\times10^{-3}$ and report the OOD accuracy of the best-ID checkpoint.
For Local-Maze and \Minigrid{}, we likewise grid-search $\{10^{-4}, 3\times10^{-4}, 5\times10^{-4}, 10^{-3}\}$ and select by in-distribution validation accuracy.

\subsection{Licenses for Existing Assets}
\label{app:licenses}

We list the main existing assets used in the experiments and their licenses or terms of use below.
The Dyck-$(30,5)$ and \Localmaze{} datasets used in this paper are synthetically generated by our experimental pipeline from the task definitions described in Appendix~\ref{app:dyck-details} and Appendix~\ref{app:maze-details}; we do not redistribute third-party dataset files for these two settings.
For commercial LLM baselines, we only access the models through their public APIs for evaluation and do not redistribute model weights or generated datasets.

\begin{itemize}[leftmargin=*, itemsep=2pt, topsep=2pt]
  \item \textbf{\Minigrid{} environments}~\citep{10.5555/3666122.3669331}: DoorKey, MultiRoom, and LavaCrossing environments; Apache License 2.0 (\href{https://github.com/Farama-Foundation/Minigrid}{Farama-Foundation/Minigrid}).
  \item \textbf{Mamba/Mamba2 implementation}~\citep{gu2023mamba,dao2024transformers}: sequence-model baselines and Mamba2 core variants; Apache License 2.0 (\href{https://github.com/state-spaces/mamba}{state-spaces/mamba}).
  \item \textbf{Qwen3-4B-Base}: open-weight LLM baseline fine-tuned on Dyck-$(30,5)$; Apache License 2.0 (\href{https://huggingface.co/Qwen/Qwen3-4B-Base}{Qwen/Qwen3-4B-Base}).
  \item \textbf{Gemini, Claude, and GPT API models}: commercial API services governed by the relevant provider terms of use; no model weights are redistributed.
\end{itemize}

\subsection{Compute Resources}
\label{app:compute-resources}

The Dyck, \Localmaze{}, and \Minigrid{} experiments each used one NVIDIA H100 GPU. The inference-timing measurement in Appendix~\ref{app:inference-cost} used one NVIDIA GH200 GPU.

\subsection{\texorpdfstring{\Localmaze}{Local-Maze}}
\label{app:maze-details}
\begin{figure}[t]
    \centering
    \includegraphics[width=.6\linewidth]{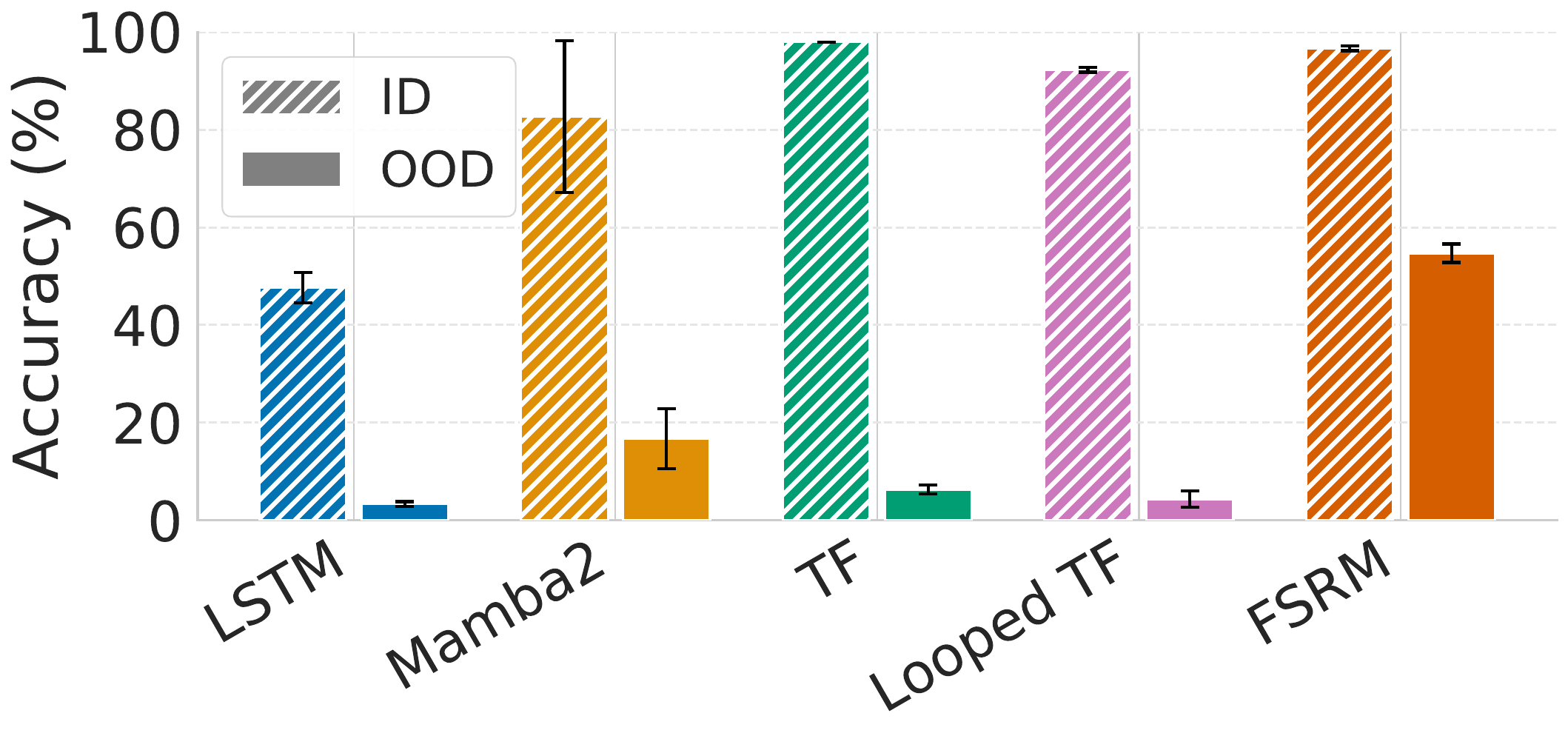}
    \caption{\Localmaze{} ID and OOD accuracy under matched compute.}
    \label{fig:maze-compute-match}
\end{figure}

\Localmaze extends the navigation task of \citet{CTM}. A navigator explores the maze under a fixed right-hand-rule policy; the model itself does not control the navigator. This decouples the input stream from the model under test and ensures that the same maze yields the same observation sequence across runs.
At stream step $s$, the model receives an egocentric $7\times7$ observation $O_s$ and, after the episode, predicts the shortest path from start to goal as a sequence of actions (\texttt{up, down, left, right, pause}).
The stream length $S$ varies with maze geometry, and the navigator's position cannot be directly inferred from the egocentric view.
The task therefore requires integrating task-relevant observations, discarding information from dead-end branches, and iteratively consolidating a hypothesis about the shortest path as new observations arrive.
Supervision is the true shortest path, given in the same action format; accuracy is measured by exact match after removing \texttt{pause} tokens from both prediction and ground truth.
Figure~\ref{fig:maze-compute-match} reports the matched-compute \Localmaze{} comparison, which controls for differences in per-update training cost. To match compute, we changed only the channel width ($ch$) of each model and selected widths with similar wall-clock time per training batch. We measured this cost as the average forward-plus-backward time per batch over 50 batches after 10 warmup batches.

\begin{figure}[t]
    \centering
    \includegraphics[width=0.6\linewidth]{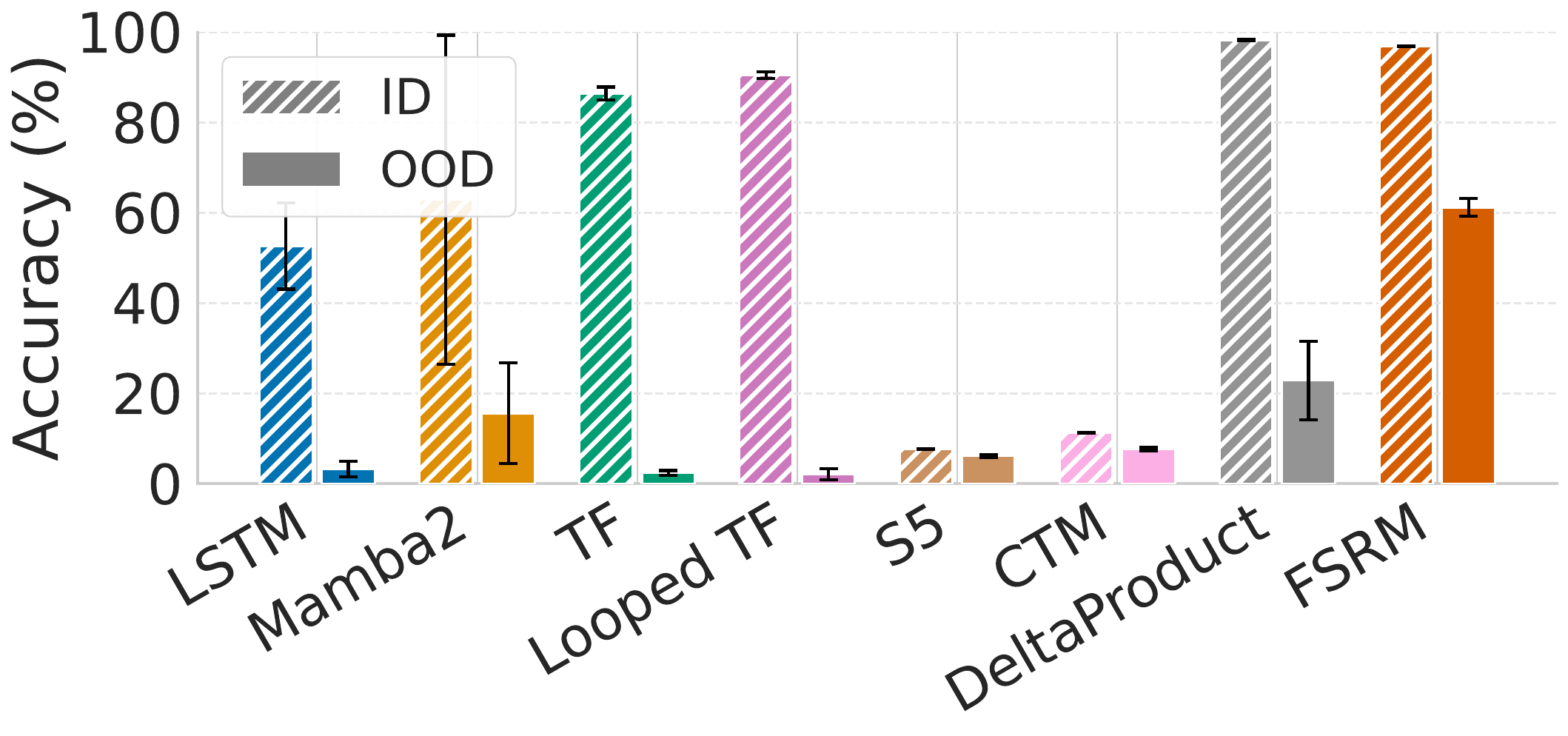}
    \caption{Extended \Localmaze{} ID and OOD accuracy comparison including S5 and CTM baselines.}
    \label{fig:maze-id-ood-full}
\end{figure}

Because the maze task is particularly related to \textit{memorizing} task-relevant information, we additionally experimented with a memory-inspired design of $\core$. 
In particular, we wrapped the original attention-based $\core$ in Appendix~\ref{app:impl-J} with a GRU. 
Letting $\bW_t$ denote the hidden state of the GRU, we replaced the original $\core$ with
\begin{align}
\core_{\mathrm{GRU}}(\lat, \bC) = \mathrm{GRU}(\core(\lat, \bC); \bW)_i.
\label{eq:internal-memory}
\end{align}
 where $\bW$ is updated at every fast step by the GRU's internal recurrence (omitted for brevity).
This module also acts as an internal filter of the interaction features, encouraging continuity of the effective coupling even when $\bC_t$ varies rapidly.

\subsection{Dyck Language}\label{app:dyck-details}

We use Dyck-$(k,m)$~\citep{hewitt2020rnns}, a formal language with $k$ bracket types and maximum nesting depth $m$, to test whether the model can learn the underlying stack rule from short sequences and apply it to streams whose horizon is not known in advance.
We set $k=30$ and $m=5$.
For example, on input ``\verb|({[]|'', the prediction after the fourth token is ``\verb|}|'' because ``\verb|{|'' is the most recent unclosed bracket.
The task requires tracking all unclosed brackets, whose number at step $s$ corresponds to the current stack depth and may grow with the sequence.
Once the structural rule is captured, the same rule applies as sequence length increases.

\citet{hewitt2020rnns} originally formulated this task as standard next-token prediction.
However, because there are multiple valid choices for the type of the next open bracket, pure next-token prediction is not deterministic.
To eliminate this source of indeterminacy in accuracy evaluation, we score only closing brackets and \texttt{*}.
Predictions are made sequentially at every stream position $s$.
Given the prefix observed through $s$, the target is the closing bracket for the most recent unclosed bracket; when the stack is empty, the target is \texttt{*}.

For the training dataset, we randomly generate Dyck-$(30,5)$ strings and train on $10{,}000$ strings of length at most $40$.
For evaluation, we separate length extrapolation from the Dyck OOD distribution.
The ID curve uses randomly generated Dyck-$(30,5)$ strings at lengths up to $2{,}560$, so it also tests lengths far beyond the training range.
The OOD curve uses an \textit{$n$-regular run}: a sequence that (i) begins with a random sequence of open brackets of random length and (ii) is followed by an infinite repetition of depth-$n$ openings and depth-$n$ closings.
For example, ``\verb|({[([[]](()){{}}|$\cdots$'' is a $2$-regular run, and Figure~\ref{fig:dyckexample} shows a $1$-regular run.
In the accuracy results, Dyck OOD refers to the $1$-regular evaluation distribution, not merely to the fact that the sequence length exceeds the training maximum.
The model sequentially predicts on $1{,}000$ strings for each evaluation distribution, and accuracy is computed token-wise.
The PCA latent-traversal visualizations are separate qualitative probes and use $5$-regular runs.

\subsubsection{Frontier LLM Setting}

For the comparison with production LLMs on Dyck-$(30,5)$, we queried each model via its public API using a common prompt that describes the stack‑based algorithm (Table~\ref{tab:dyck-prompt}).
Following \citet{chae2024language}, we designed the prompt so that it clearly separates (i) a natural‑language specification of the ground‑truth stack algorithm from (ii) explicit execution instructions that tell the model to simulate this algorithm step‑by‑step and output only the prediction string.
No additional fine‑tuning was performed; all models were evaluated in a pure inference setting. We set the temperature to 0 if the API accepts it, so that the model’s behavior is more deterministic.

Modern LLMs can adapt their test‑time compute by varying the number of their internal ``reasoning tokens,'' but commercial APIs strictly enforce a maximum on the total number of output tokens (reasoning plus final answer). As of November 2025, the documented upper limits for our models are 32k tokens for Claude Opus 4.5 and 64k tokens for both Gemini 3.0 Pro and GPT‑5.1. For each provider, we first empirically determined the longest Dyck input length for which the prompt, the model’s reasoning tokens, and the final prediction string all remain within this limit. We then generated 100 Dyck sequences at this length and measured token‑level prediction accuracy. Use of external tools such as Python interpreters or code execution APIs was explicitly disabled so that each model had to execute the algorithm internally.
For reference, Table~\ref{tab:dyck-claude-example} shows an example of the internal reasoning tokens produced by Claude Opus 4.5 on a representative input.

\subsubsection{Fine-tuned Qwen3-4B-Base}

The frontier LLM evaluation is useful as a test of whether current production models can execute the Dyck algorithm from a prompt, but it is not fully symmetric with our setting because those models are not trained on the Dyck data used by \method. We therefore conducted an additional comparison with Qwen3-4B-Base, an open-weight LLM that we fine-tuned directly on the same Dyck-$(30,5)$ data and evaluation protocol used in Figure~\ref{fig:dyck-llms}.

We fine-tuned Qwen3-4B-Base with LoRA on the same Dyck-$(30,5)$ split used in Figure~\ref{fig:dyck-llms}: training strings of length at most $40$, with $10{,}000$ training samples and $1{,}000$ validation samples, and report results over $3$ random seeds. We used a shared training setup across baselines with batch size $128$, learning rate $10^{-4}$, $30$ epochs, and bf16 precision. Because we observed overfitting later in training, we selected the best checkpoint by early stopping on the validation set. For LoRA, we used rank $32$, $\alpha=16$, dropout $0.1$, and adapted the standard projection and feedforward modules, following the official Qwen fine-tuning recipe: \url{https://qwen.readthedocs.io/en/v1.5/training/SFT/example.html\#python-script}.

We evaluated the fine-tuned Qwen model using the same OOD protocol as Figure~\ref{fig:dyck-llms}, namely $1$-regular runs. We first evaluated a long-horizon OOD run with context length $20{,}480$; we were unable to evaluate beyond $20{,}480$ because the model ran out of memory even with batch size $1$. The result shows that Qwen3-4B-Base achieves high accuracy at in-distribution lengths but fails to generalize to the OOD length, with accuracy dropping to near $50\%$ at length $20{,}480$.

This train-short-test-long failure is consistent with prior reports that pretrained language models fine-tuned on short algorithmic instances can achieve near-perfect in-distribution accuracy while degrading rapidly out of distribution with length~\citep{anil2022exploring}; related work also finds that OOD behavior can be nonmonotone due to shortcut solutions, positional effects, or output-distribution drift~\citep{liu2023transformers,wang2024resonance,zhou2024transformers,du2025longshort}.

\subsubsection{\method{} Hierarchical Architecture}\label{app:dyck-hierarchical}

For this task, we used a two-layer extension of \method{}.
Since the task has a hierarchical stack structure, this architecture instantiates a hierarchical version of \method{} with two fast processes.
Let $\lat_s^{(1)}(t)$ and $\lat_s^{(2)}(t)$ denote the first- and second-layer latents at stream index $s$ and inner-loop index $t$, with each layer carried forward as $\lat_{s+1}^{(\ell)}(0)=\lat_s^{(\ell)}(T)$.
The first layer receives the encoded observation $\bC_s$ and evolves as
\begin{align*}
\lat_s^{(1)}(t+1)
= \core_1\!\left(\lat_s^{(1)}(t), \bC_s\right),
\qquad t=0,\ldots,T-1 .
\end{align*}
Writing $r_s^{(1)}=\mathrm{Readout}_1(\lat_s^{(1)}(T))$ and
$\mathcal{H}_s=[r_{s-H+1}^{(1)},\ldots,r_s^{(1)}]$ for the length-$H$ history queue, the second layer evolves as
\begin{align*}
\lat_s^{(2)}(t+1)
= \core_2\!\left(\lat_s^{(2)}(t), \mathcal{H}_s\right),
\qquad t=0,\ldots,T-1 .
\end{align*}
Predictions are obtained by decoding $\lat_s^{(2)}(T)$ at each stream index $s$; Figure~\ref{fig:code-meta} gives the implementation-level pseudocode.
We set the history size to $H=4$.

\begin{figure}[tb]
  \centering
    \begin{lstlisting}[language=python]
class FastSlow():
    def __init__(dim, oscillator_dim, init_gamma):
        self.embed = nn.Embedding()
        self.R1 = R(dim, oscillator_dim, init_gamma)
        self.R2 = R(dim, oscillator_dim, init_gamma)

    def forward(tokens, H, T):
        # C: channels, K: num oscillators per token, H: history size
        c = self.embed(tokens)      # (L, K, C)
        x = randn(K, C)
        z = randn(H, K, C)
        h = queue(H)           # length-H history queue
        z_out = zeros(H, K, C)
    
        # Slow loop
        for s in range(len(embeddings)):
            # First fast-layer
            for t in range(T):
                x = self.R1(x, c[s]) 
            x_out = x_readout(x)
    
            # Second fast-layer
            h.enqueue(x_out)
            for t in range(T):
                z = self.R2(z+z_out, h)
            z_out = z_readout(z)
    
            logits[s] = classifier(z_out)
        return logits
    \end{lstlisting}
  \caption{Pseudocode for the two-stage architecture used in Section~\ref{sec:exp}. }
  \label{fig:code-meta}
  \vspace{-5pt}
\end{figure}

\subsection{Reinforcement Learning}\label{app:rl-details}
\begin{figure}[t]
    \centering
    \captionsetup{skip=2pt}
    \captionsetup[sub]{skip=2pt}
    \begin{subfigure}{0.24\linewidth}
        \centering
        \includegraphics[width=0.9\linewidth]{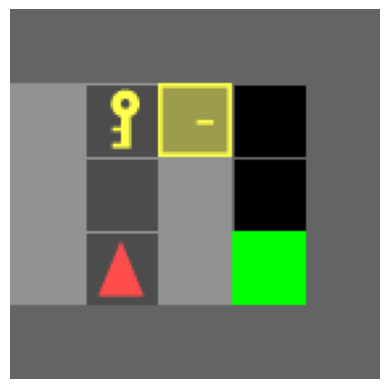}
        \caption{DoorKey (ID)}
        \label{fig:doorkey5x5}
    \end{subfigure}
    \hfill
    \begin{subfigure}{0.24\linewidth}
        \centering
        \includegraphics[width=0.9\linewidth]{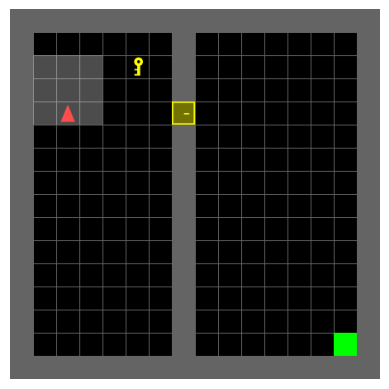}
        \caption{DoorKey (OOD)}
        \label{fig:doorkey_ood}
    \end{subfigure}
    \hfill
    \begin{subfigure}{0.24\linewidth}
        \centering
        \includegraphics[width=0.9\linewidth]{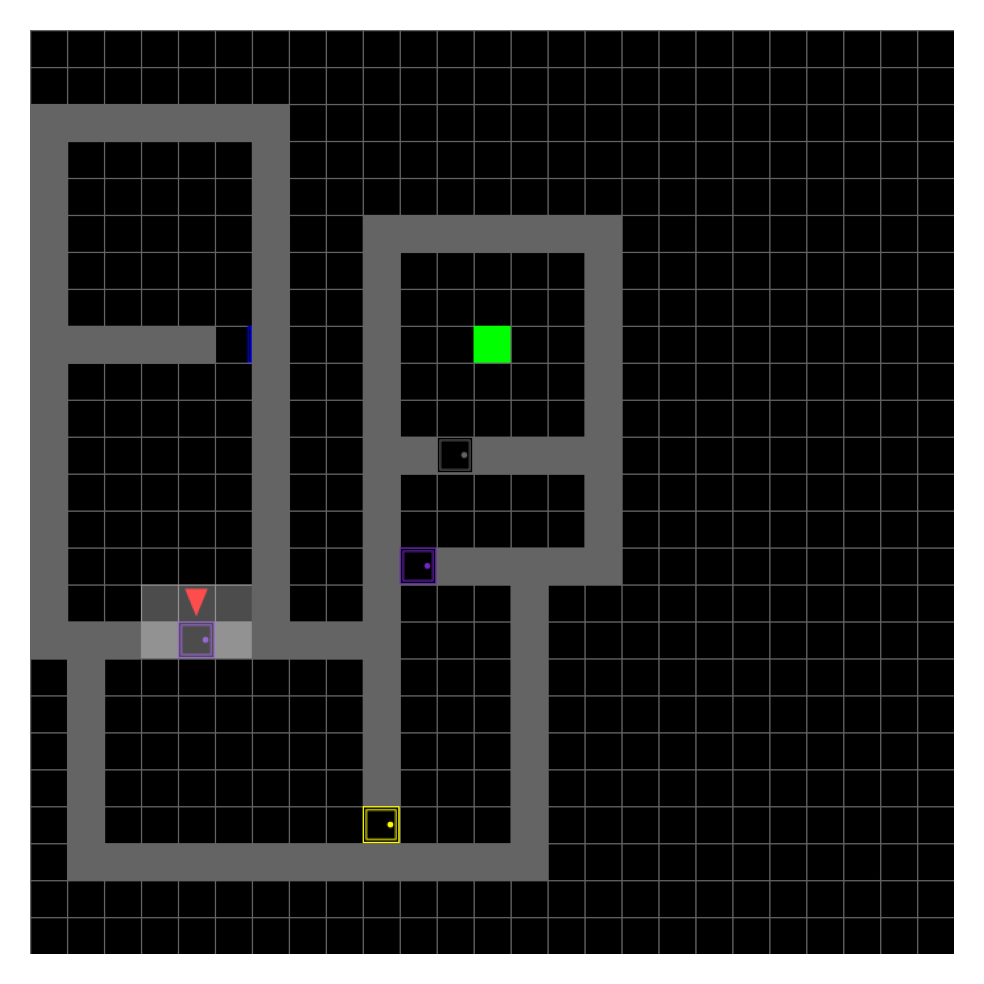}
        \caption{MultiRoom (OOD)}
        \label{fig:multiroom_ood}
    \end{subfigure}
    \hfill
    \begin{subfigure}{0.24\linewidth}
        \centering
        \includegraphics[width=0.9\linewidth]{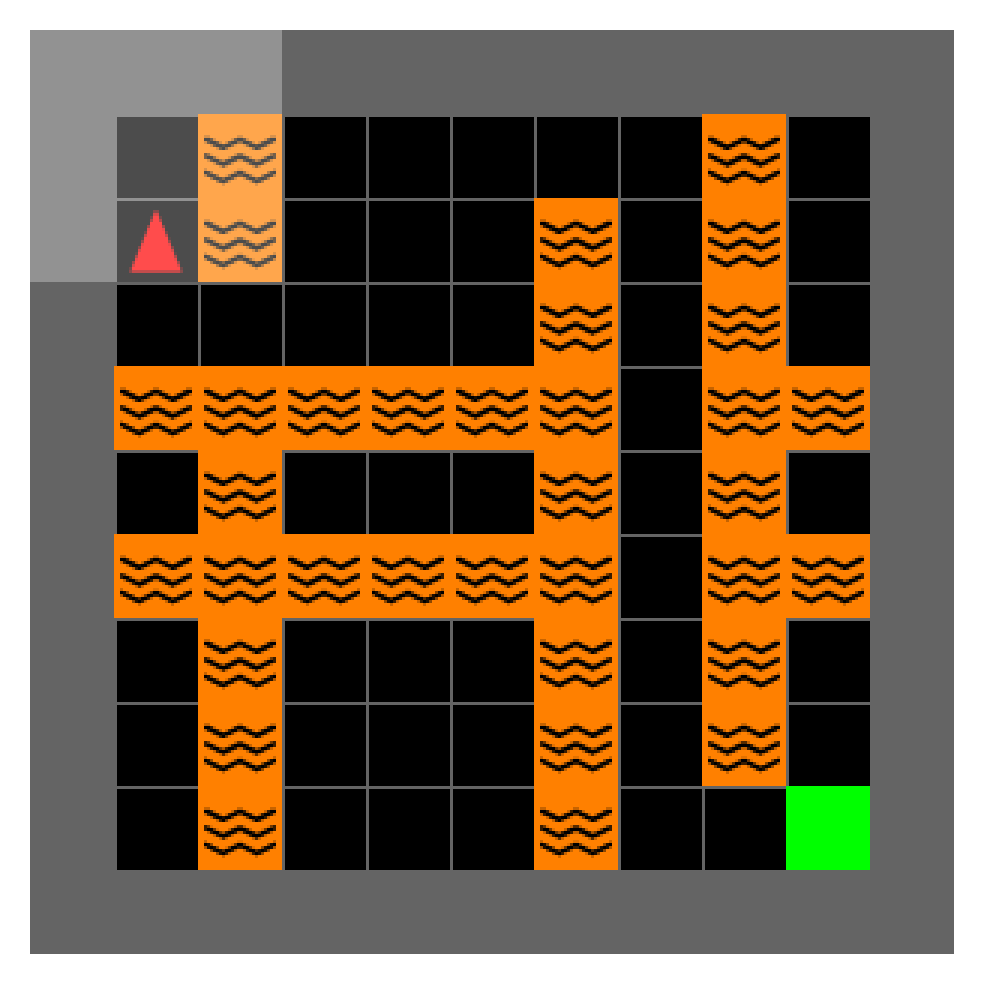}
        \caption{LavaCrossing (OOD)}
    \end{subfigure}
    \caption{Examples of the partially observable \Minigrid reinforcement-learning environments used in our experiments. DoorKey illustrates the ID-to-OOD size shift; MultiRoom and LavaCrossing show additional OOD tasks that require long-horizon memory and planning under sparse rewards.}
    \label{fig:minigrid-description}
\end{figure}

We evaluate partially observable reinforcement-learning tasks from \Minigrid~\citep{10.5555/3666122.3669331}.
At each step, the agent receives an egocentric $7\times7$ observation with object, color, and door-state channels, and selects an action $a_s \in \{\texttt{left}, \texttt{right}, \texttt{forward}, \texttt{toggle}, \texttt{pickup}, \texttt{drop}, \texttt{done}\}$.
Reward is given only at the end of a successful episode.
Unlike Dyck and \Localmaze, no ground-truth action sequence is provided, so the model must identify which observations are worth consolidating into its evolving plan.
Agents must navigate an environment to reach the goal while:
\begin{itemize}
    \item \textbf{DoorKey}: finding a key and unlocking a door
    \item \textbf{LavaCrossing}: avoiding the impassable lava river
    \item \textbf{MultiRoom}: going through multiple rooms with doors.
\end{itemize}
Figure~\ref{fig:minigrid-description} provides illustrated examples of the environments.

To train our model, we followed the strategy of \citet{Geiping2025LatentReasoning, HRM} for computational efficiency and used gradient truncation.
In particular, we set $T=10$, and backpropagated through only
the last $5$ iterations of the recurrent loop at each $s$.

\begin{table}[ht]
  \centering
  \caption{ID/OOD splits for the \Minigrid environments. We train agents on the simplest configuration and test for zero-shot generalization on OOD environments.}
  \label{tab:env-id-ood}
  \begin{tabular}{llcc}
    \toprule
    Environment & Label & ID & OOD \\
    \midrule
    MiniGrid-MultiRoom-N2-S4-v0 & N2-S4 & \cmark & \xmark \\
    MiniGrid-MultiRoom-N4-S5-v0 & N4-S5 & \xmark & \cmark \\
    MiniGrid-MultiRoom-N6-v0    & N6    & \xmark & \cmark \\
    \midrule
    MiniGrid-DoorKey-5x5-v0     & 5x5   & \cmark & \xmark \\
    MiniGrid-DoorKey-6x6-v0     & 6x6   & \xmark & \cmark \\
    MiniGrid-DoorKey-8x8-v0     & 8x8   & \xmark & \cmark \\
    MiniGrid-DoorKey-16x16-v0   & 16x16 & \xmark & \cmark \\
    \midrule
    MiniGrid-LavaCrossingS9N1-v0  & S9N1  & \cmark & \xmark \\
    MiniGrid-LavaCrossingS9N2-v0  & S9N2  & \xmark & \cmark \\
    MiniGrid-LavaCrossingS9N3-v0  & S9N3  & \xmark & \cmark \\
    MiniGrid-LavaCrossingS11N5-v0 & S11N5 & \xmark & \cmark \\
    \bottomrule
  \end{tabular}
\end{table}

\section{Additional Results}\label{app:additional-results}

\subsection{Dyck}

\subsubsection{ID and OOD Accuracy}
\label{app:dyck-id-ood-accuracy}
\begin{figure}[H]
  \centering
  \captionsetup[sub]{skip=2pt}
  \begin{subfigure}[t]{0.49\linewidth}
    \centering
    \includegraphics[width=\linewidth]{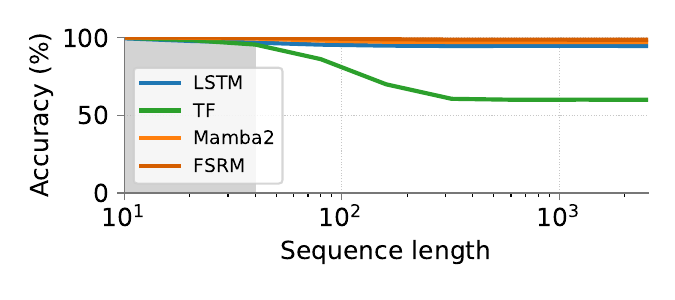}
    \caption{ID accuracy}
  \end{subfigure}
  \begin{subfigure}[t]{0.49\linewidth}
    \centering
    \includegraphics[width=\linewidth]{figures/dyck/acc_ood.pdf}
    \caption{OOD accuracy}
  \end{subfigure}
  \caption{Dyck accuracy on ID and OOD runs. Both panels evaluate lengths beyond the training range. The OOD panel shows the curves of Figure~\ref{fig:main-results} and is included here for easy comparison with the ID panel.}
  \label{fig:dyck-id-ood-accuracy}
\end{figure}

Figure~\ref{fig:dyck-id-ood-accuracy} reports the ID curve together with the OOD curve shown in Figure~\ref{fig:main-results}\subref{fig:main-dyck-acc}. Both curves extend to length $2{,}560$, beyond the training maximum of $40$.

\subsubsection{Energy Traces}

\label{app:dyck-energy}
\begin{figure}[H]
  \centering
  \includegraphics[width=0.82\linewidth]{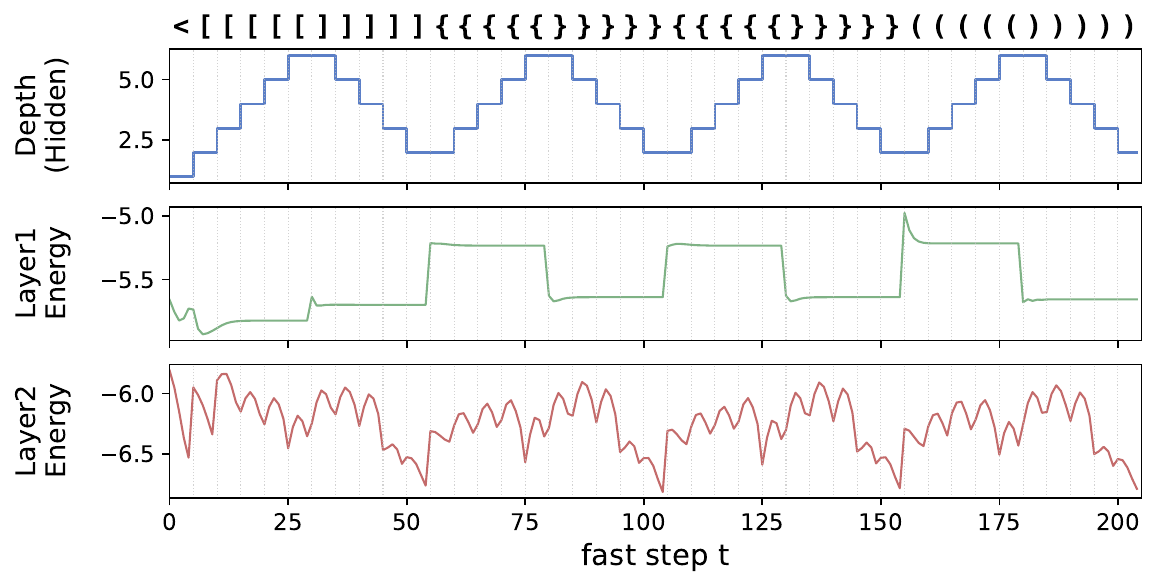}
  \caption{Energy traces (see \eqref{eq:energy}) for the first and second layers of the hierarchical version of \method{} during a $5$-regular Dyck run. The dashed lines indicate token arrival times.}
  \label{fig:dyck-energy}
\end{figure}

Through learning the prediction task, \method{} discovers latent dynamics that align with the structural organization of the Dyck language.
Recall that the \akorn-based recurrence used in our experiments admits an energy-like scalar that reflects the organizational level of the latent state (Appendix~\ref{sec:energy}).
We show the evolution of this energy in Figure~\ref{fig:dyck-energy}, and PCA projections of the corresponding latents in Figure~\ref{fig:dyck-pca}.

The energy of $\lat^{(1)}$ spikes upon the arrival of a new token and then stabilizes at a level that depends on the bracket type and whether the current token is opening or closing.
In contrast, the energy of $\lat^{(2)}$ is largely insensitive to bracket identity and is instead organized by stack \emph{depth} (see Appendix~\ref{app:dyck-hierarchical} for the definition of $\lat^{(1)}$ and $\lat^{(2)}$).
This qualitative shift in energy organization indicates that the second fast process abstracts away from token identity and operates at a higher structural level.
These emergent structures are also consistent with the results in Appendix~\ref{app:stack_probe} and are unique to the models that generalize on OOD (Figures~\ref{fig:app-dyck-core-latents} and~\ref{fig:app-dyck-baseline-latents}).

\subsubsection{Linear Probes}  
\label{app:stack_probe}
To quantify whether the recurrent state contains the variables required by the Dyck rule, we train linear probes on frozen hidden states from the in-distribution (length $10$--$40$) training split.
The probe predicts the bracket type of each currently unresolved opener (open bracket) in the active stack, with one head per stack position from the top; \emph{opener age} denotes the number of steps elapsed since that opener was read.
Missing stack slots and age-zero events are masked.
We then evaluate the probes on length-OOD Dyck-$(30,5)$ sequences ($30$ bracket types, maximum stack depth $5$).
The successful \method{} model preserves linearly decodable information about all five unresolved stack positions, whereas sequence baselines largely retain only the top stack item or lose stack information as opener age increases.
Since only the top stack item corresponds to the immediate prediction target, decodability of deeper stack entries provides evidence that the latent state retains information needed for prediction targets that arise when inner brackets are eventually closed.

\begin{figure}[H]
  \centering
  \includegraphics[width=.85\linewidth]{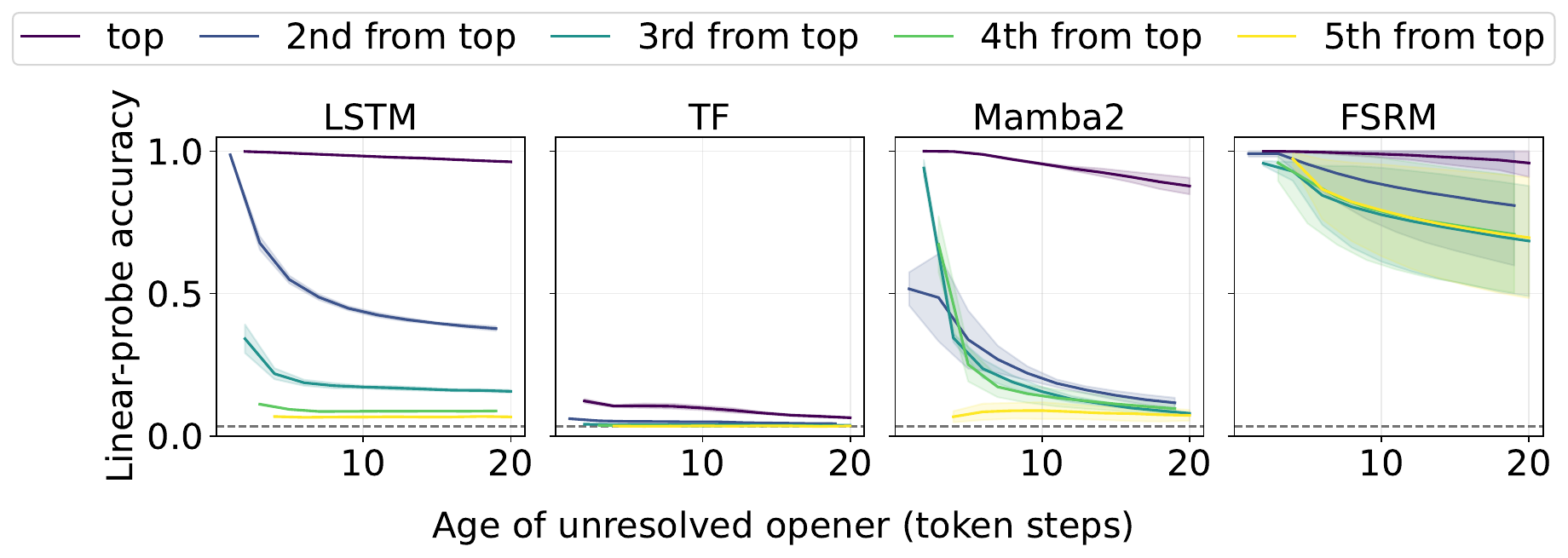}
  \caption{Linear probe of unresolved stack contents on Dyck-$(30,5)$.
  Models are trained on length $10$--$40$ sequences and frozen.
  A multi-head linear probe is trained on the ID split to predict the bracket identity of the $i$-th unresolved opener from the top of the active stack.
  Probes are evaluated on length-$2{,}560$ OOD sequences. The x-axis shows opener age, i.e.\ the number of steps elapsed since that opener was read (restricted to still-unresolved openers).
  Chance accuracy is $1/30$, as shown by the dashed line.}
  \label{fig:dyck-stack-probes}
\end{figure}

\kohei{To separate long-horizon retention from depth generalization, we also evaluate on a depth-OOD variant in which the sequence length remains in-distribution while the maximum stack depth increases from $5$ at training time to $10$ at test time.
This is a stricter test than length extrapolation alone: success now requires handling more simultaneously active unresolved openers than were observed during training, rather than merely preserving familiar stack states for longer durations.
Note that the base model weights are frozen (trained on depth-$5$ sequences), and only the linear probe heads are retrained on an in-distribution dataset with maximum stack depth $10$, so that they cover the additional stack positions.
We find that \method{} remains substantially more robust than the sequence baselines under this depth shift, indicating that its recurrent latent encodes a stack-structured state that transfers beyond the training depth regime, complementing the length-OOD result in Figure~\ref{fig:dyck-stack-probes}.}

\begin{figure}[H]
  \centering
  \includegraphics[width=.95\linewidth]{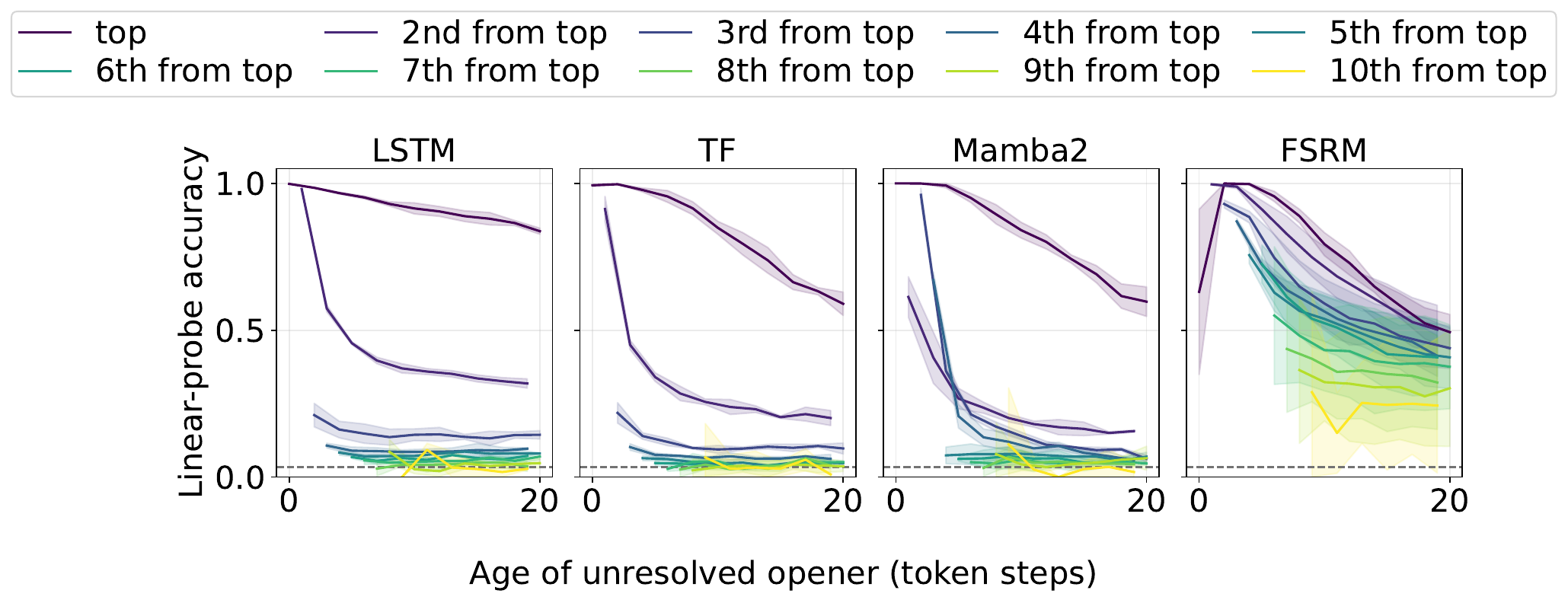}
  \caption{\kohei{Depth extrapolation on Dyck-$(30,\cdot)$.
  Models are trained on sequences of length $10$--$40$ with maximum stack depth $5$, and evaluated on sequences with the same length range but maximum stack depth $10$.
  This setting isolates generalization to deeper active stacks from generalization to longer sequence lengths.
  \method{} remains more robust than the sequence baselines under this depth shift, indicating that its latent-state representation transfers beyond the training stack depth regime.}}
  \label{fig:dyck-depth-ood}
\end{figure}

\subsubsection{Core Ablation}
\label{app:dyck-core-ablation}
\begin{figure}[H]
  \centering
  \begin{subfigure}[b]{0.30\linewidth}
    \begin{tikzpicture}
      \node[inner sep=0pt] (pca) {\includegraphics[width=\linewidth]{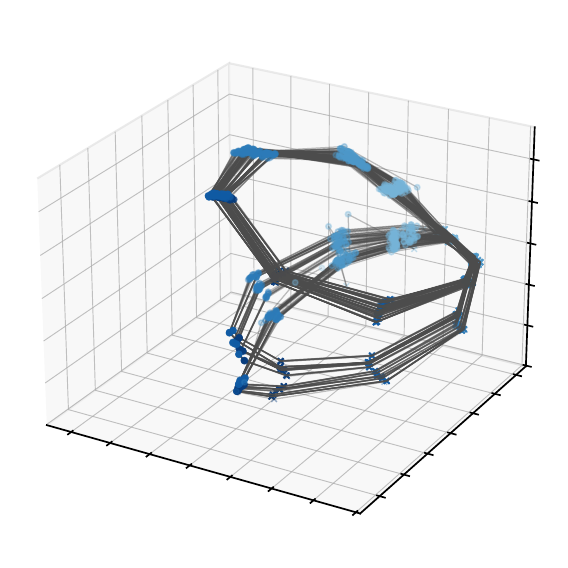}};
      \node[anchor=south east, fill=white, fill opacity=0.86, text opacity=1, inner sep=1.2pt, rounded corners=0.6pt, font=\scriptsize]
        at ([xshift=-2pt,yshift=2pt]pca.south east) {OOD Acc: 92.1\%};
    \end{tikzpicture}
    \caption{\kohei{\method{}-Linear+FFN}}
  \end{subfigure}
  \hfill
  \begin{subfigure}[b]{0.30\linewidth}
    \begin{tikzpicture}
      \node[inner sep=0pt] (pca) {\includegraphics[width=\linewidth]{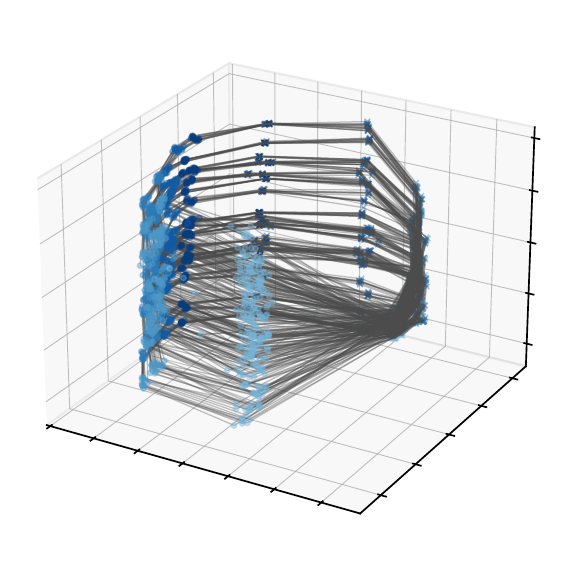}};
      \node[anchor=south east, fill=white, fill opacity=0.86, text opacity=1, inner sep=1.2pt, rounded corners=0.6pt, font=\scriptsize]
        at ([xshift=-2pt,yshift=2pt]pca.south east) {51.1\%};
    \end{tikzpicture}
    \caption{\kohei{\method{}-Mixer+FFN}}
  \end{subfigure}
  \hfill
  \begin{subfigure}[b]{0.30\linewidth}
    \begin{tikzpicture}
      \node[inner sep=0pt] (pca) {\includegraphics[width=\linewidth]{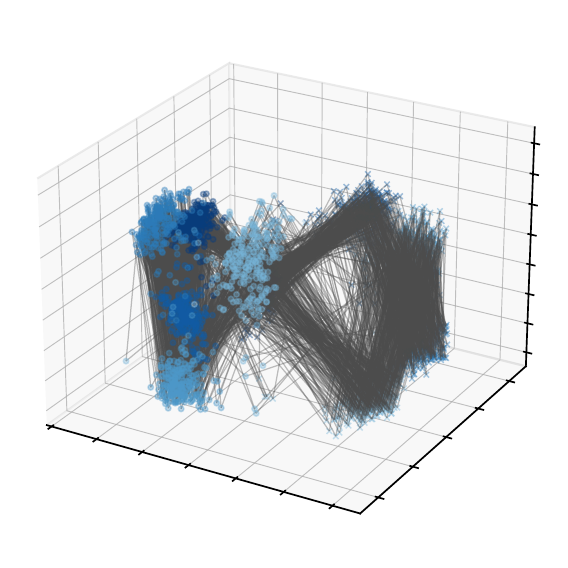}};
      \node[anchor=south east, fill=white, fill opacity=0.86, text opacity=1, inner sep=1.2pt, rounded corners=0.6pt, font=\scriptsize]
        at ([xshift=-2pt,yshift=2pt]pca.south east) {51.8\%};
    \end{tikzpicture}
    \caption{\kohei{\method-Conv}}
  \end{subfigure}

  \vspace{0.8em}

  \begin{subfigure}[b]{0.30\linewidth}
    \begin{tikzpicture}
      \node[inner sep=0pt] (pca) {\includegraphics[width=\linewidth]{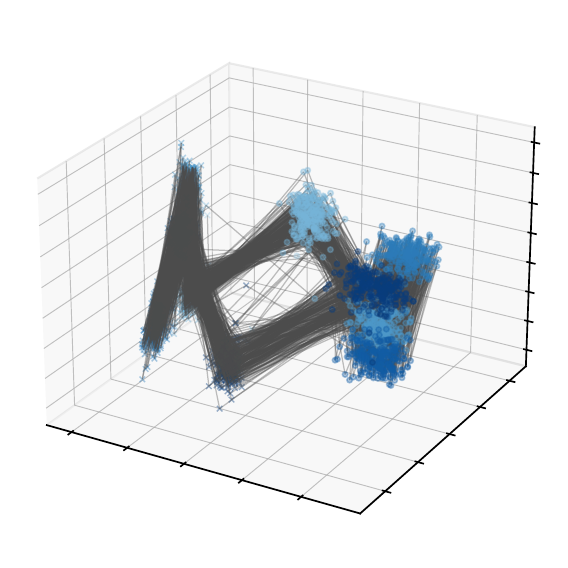}};
      \node[anchor=south east, fill=white, fill opacity=0.86, text opacity=1, inner sep=1.2pt, rounded corners=0.6pt, font=\scriptsize]
        at ([xshift=-2pt,yshift=2pt]pca.south east) {67.0\%};
    \end{tikzpicture}
    \caption{\kohei{\method{}-Conv+FFN}}
  \end{subfigure}
  \hfill
  \begin{subfigure}[b]{0.30\linewidth}
    \begin{tikzpicture}
      \node[inner sep=0pt] (pca) {\includegraphics[width=\linewidth]{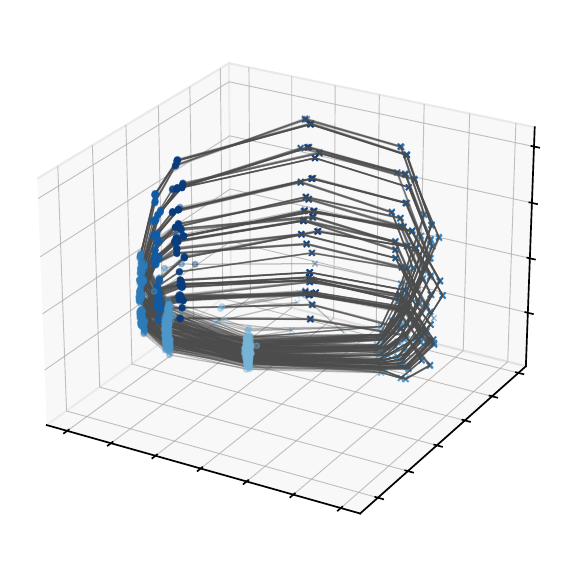}};
      \node[anchor=south east, fill=white, fill opacity=0.86, text opacity=1, inner sep=1.2pt, rounded corners=0.6pt, font=\scriptsize]
        at ([xshift=-2pt,yshift=2pt]pca.south east) {67.1\%};
    \end{tikzpicture}
    \caption{\kohei{\method-LSTM}}
  \end{subfigure}
  \hfill
  \begin{subfigure}[b]{0.30\linewidth}
    \begin{tikzpicture}
      \node[inner sep=0pt] (pca) {\includegraphics[width=\linewidth]{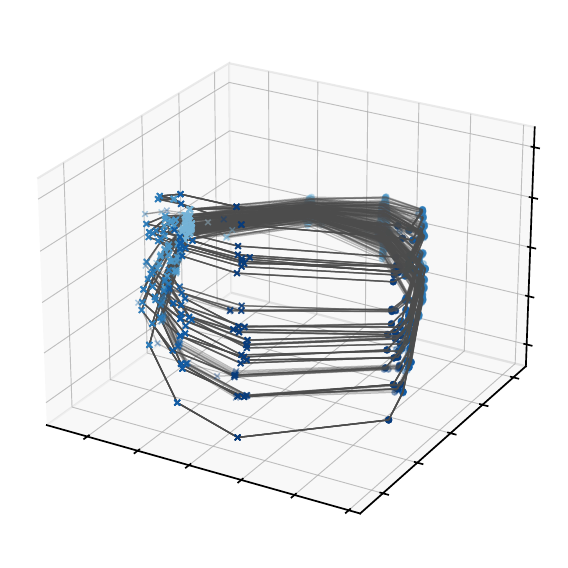}};
      \node[anchor=south east, fill=white, fill opacity=0.86, text opacity=1, inner sep=1.2pt, rounded corners=0.6pt, font=\scriptsize]
        at ([xshift=-2pt,yshift=2pt]pca.south east) {99.3\%};
    \end{tikzpicture}
    \caption{\kohei{\method{}-LSTM+FFN}}
  \end{subfigure}

  \vspace{0.8em}

  \begin{subfigure}[b]{0.30\linewidth}
    \begin{tikzpicture}
      \node[inner sep=0pt] (pca) {\includegraphics[width=\linewidth]{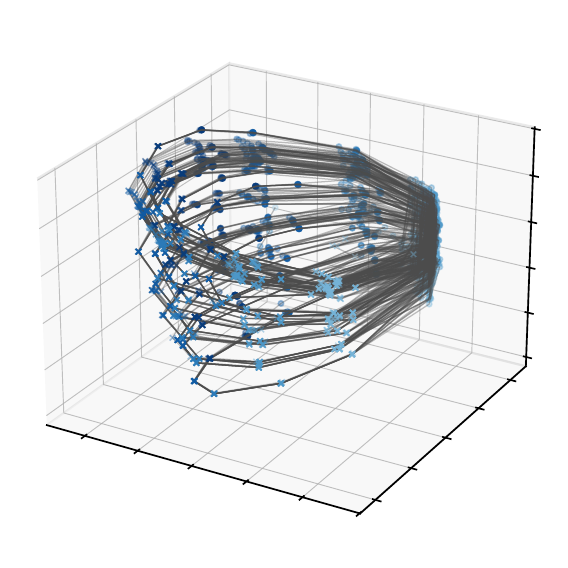}};
      \node[anchor=south east, fill=white, fill opacity=0.86, text opacity=1, inner sep=1.2pt, rounded corners=0.6pt, font=\scriptsize]
        at ([xshift=-2pt,yshift=2pt]pca.south east) {86.1\%};
    \end{tikzpicture}
    \caption{\kohei{\method{}-Mamba2+FFN}}
  \end{subfigure}
  \hfill
  \begin{subfigure}[b]{0.30\linewidth}
    \begin{tikzpicture}
      \node[inner sep=0pt] (pca) {\includegraphics[width=\linewidth]{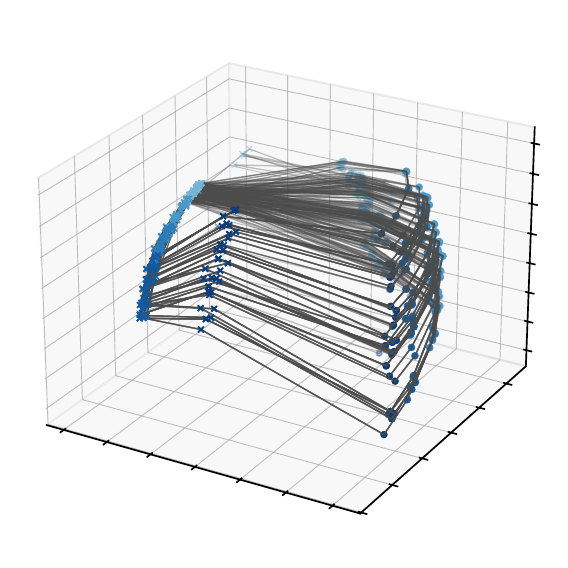}};
      \node[anchor=south east, fill=white, fill opacity=0.86, text opacity=1, inner sep=1.2pt, rounded corners=0.6pt, font=\scriptsize]
        at ([xshift=-2pt,yshift=2pt]pca.south east) {77.0\%};
    \end{tikzpicture}
    \caption{\kohei{\method-TF}}
  \end{subfigure}
  \hfill
  \begin{subfigure}[b]{0.30\linewidth}
    \begin{tikzpicture}
      \node[inner sep=0pt] (pca) {\includegraphics[width=\linewidth]{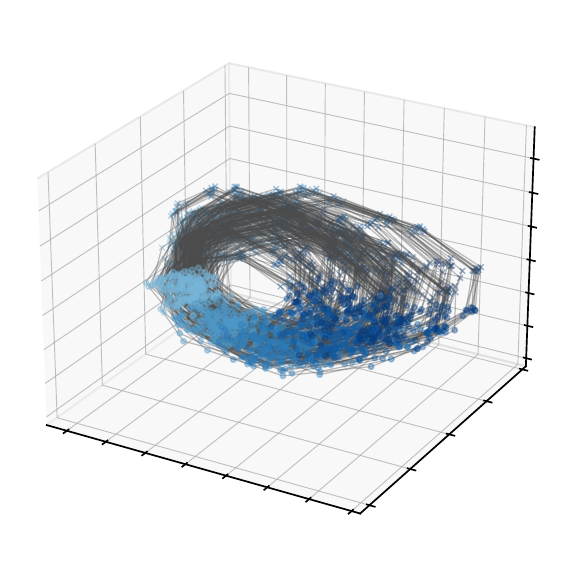}};
      \node[anchor=south east, fill=white, fill opacity=0.86, text opacity=1, inner sep=1.2pt, rounded corners=0.6pt, font=\scriptsize]
        at ([xshift=-2pt,yshift=2pt]pca.south east) {50.1\%};
    \end{tikzpicture}
    \caption{\kohei{\method-AKOrN}}
  \end{subfigure}

  \caption{\kohei{Supplemental Dyck latent traversals for FSRM core variants not shown in Figure~\ref{fig:dyck-pca}, visualized in PCA space on the same $5$-regular run. Badges show mean OOD accuracy.}}
  \label{fig:app-dyck-core-latents}
\end{figure}

\begin{figure}[H]
  \centering
  \begin{subfigure}[b]{0.42\linewidth}
    \begin{tikzpicture}
      \node[inner sep=0pt] (pca) {\includegraphics[width=\linewidth]{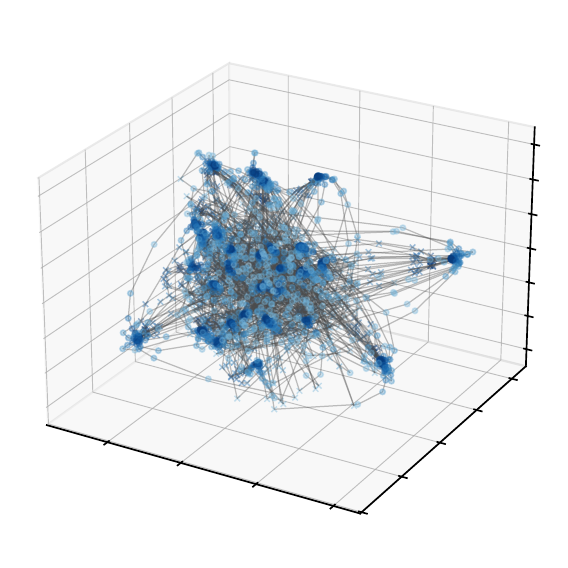}};
      \node[anchor=south east, fill=white, fill opacity=0.86, text opacity=1, inner sep=1.2pt, rounded corners=0.6pt, font=\scriptsize]
        at ([xshift=-2pt,yshift=2pt]pca.south east) {OOD Acc: 50.0\%};
    \end{tikzpicture}
    \caption{\kohei{LSTM}}
  \end{subfigure}
  \hfill
  \begin{subfigure}[b]{0.42\linewidth}
    \begin{tikzpicture}
      \node[inner sep=0pt] (pca) {\includegraphics[width=\linewidth]{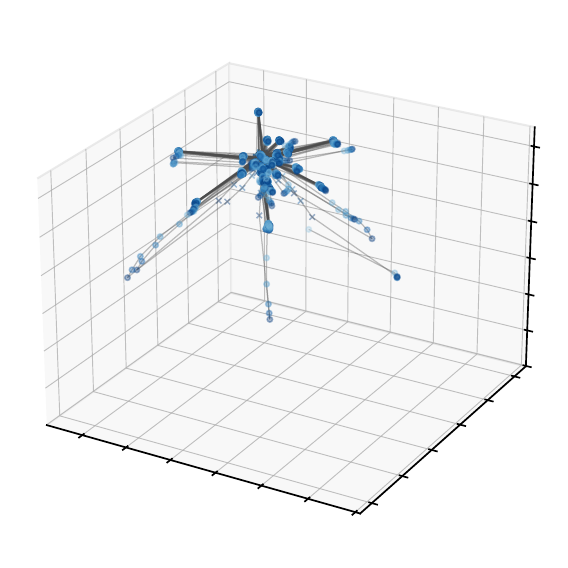}};
      \node[anchor=south east, fill=white, fill opacity=0.86, text opacity=1, inner sep=1.2pt, rounded corners=0.6pt, font=\scriptsize]
        at ([xshift=-2pt,yshift=2pt]pca.south east) {50.0\%};
    \end{tikzpicture}
    \caption{\kohei{TF}}
  \end{subfigure}

  \caption{\kohei{Supplemental Dyck latent traversals for sequence-model baselines, visualized in PCA space on the same $5$-regular run. Mamba2 is shown in Figure~\ref{fig:dyck-pca}. Badges show mean OOD accuracy.}}
  \label{fig:app-dyck-baseline-latents}
\end{figure}

We additionally ablate the recurrent core on Dyck using the core architectures described in Table~\ref{tab:dyck-core-ablation-arch}.
Figures~\ref{fig:app-dyck-core-latents} and~\ref{fig:app-dyck-baseline-latents} complement Figure~\ref{fig:dyck-pca} by showing the remaining Dyck latent traversal plots under the same PCA visualization setup.

\begin{table}[H]
\centering
\caption{Architectural hyperparameters and parameter counts for the Dyck core ablation. Parameter counts are reported in millions.}
\label{tab:dyck-core-ablation-arch}
\begin{tabular}{@{}lrr@{}}
\toprule
\textbf{Core} & \textbf{Ch.} & \textbf{Params (M)} \\
\midrule
AKOrN+FFN  & 72 & 1.334 \\
AKOrN      & 72 & 1.301 \\
Mamba2+FFN & 80 & 1.232 \\
Mamba2     & 80 & 1.180 \\
Conv+FFN   & 80 & 1.307 \\
Conv       & 84 & 1.342 \\
LSTM+FFN   & 84 & 1.337 \\
LSTM       & 88 & 1.365 \\
TF         & 80 & 1.416 \\
Linear TF     & 88 & 1.318 \\
MLP-Mixer      & 56 & 1.385 \\
FFN-only   & 92 & 1.367 \\
Attn-only  & 80 & 1.364 \\
\bottomrule
\end{tabular}%
\end{table}


\begin{figure}[H]
  \centering
  \includegraphics[width=\linewidth]{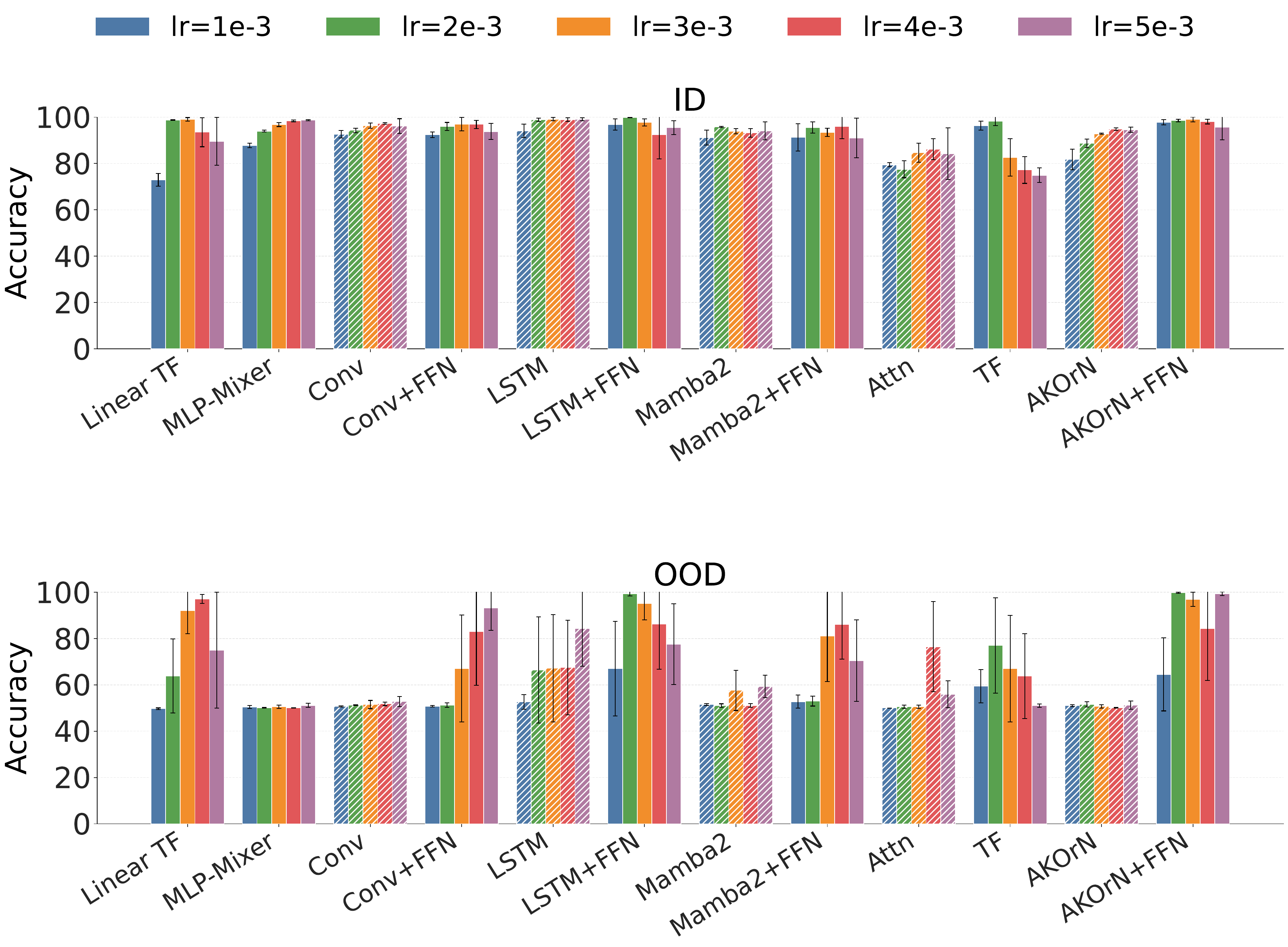}
  \caption{Full learning-rate sweep for the Dyck core ablation on lengths $1{,}280$--$2{,}560$. ID accuracy is used for model selection; OOD accuracy is reported.}
  \label{fig:dyck-core-ablation-sweep}
\end{figure}

\subsection{\texorpdfstring{\Localmaze}{Local-Maze}}

\subsubsection{Core Ablation}
\label{app:maze-core-id}
\begin{figure}[H]
    \centering
    \includegraphics[width=0.66\linewidth]{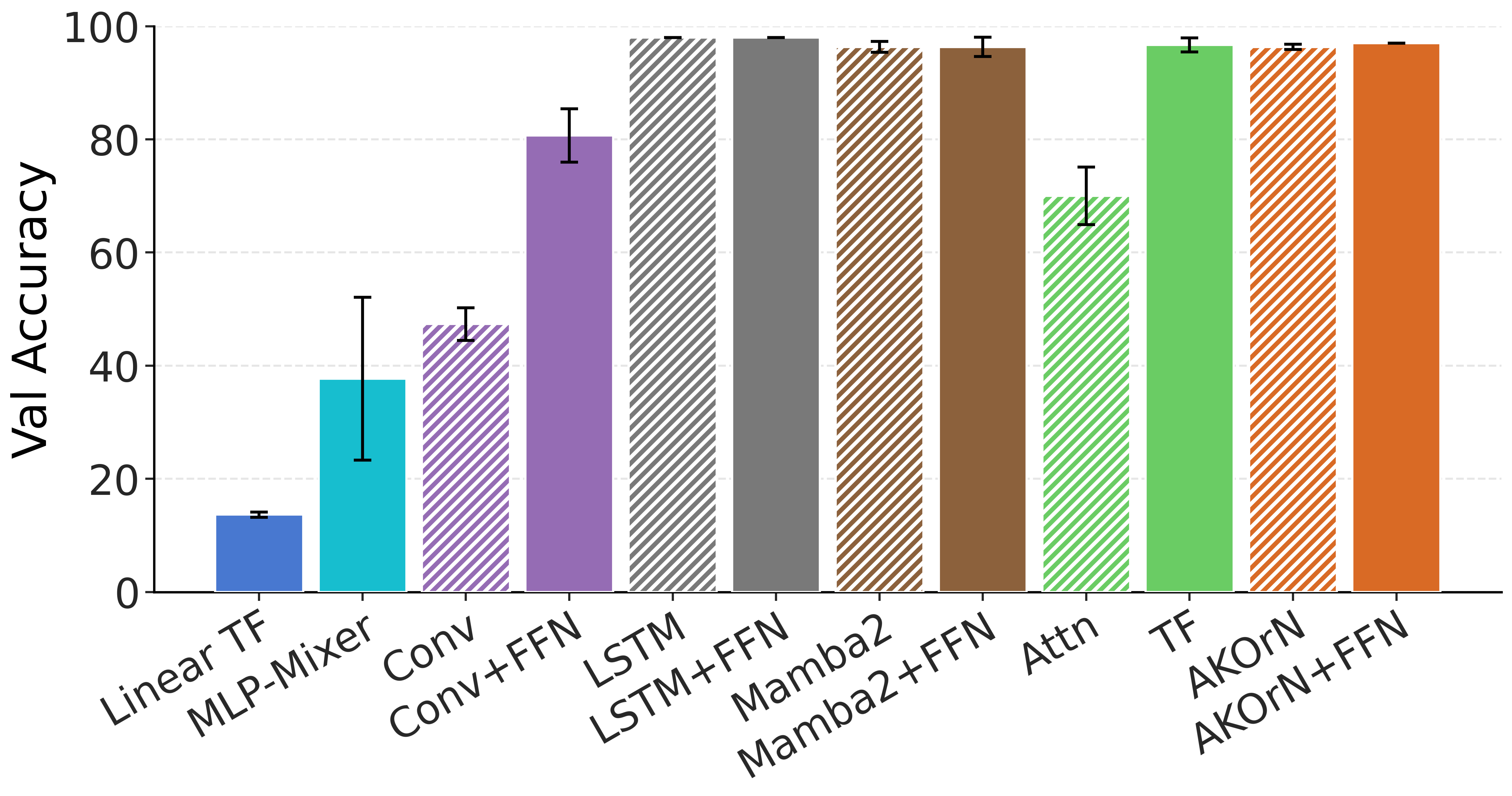}
    \caption{\Localmaze{} ID accuracy ($19\times19$) across recurrent cores swapped into the \method{} loop. Solid bars include the native or augmented FFN block; hatched bars ablate it.}
    \label{fig:module-ablation-id}
\end{figure}

Figure~\ref{fig:module-ablation-id} reports the ID counterpart of the \Localmaze{} OOD core ablation in Figure~\ref{fig:module-ablation}.

\subsubsection{Details of the ablation regarding the weight-shared structure}
\label{app:ablation_recurrent}

This appendix subsection records an additional training detail for the weight-sharing control experiment reported in Section~\ref{sec:t-and-share} (``Fast-loop structure: $T$ and weight sharing'').
For non-recurrent baselines, we scale the learning rate with depth as
$\text{lr}(T)=\sqrt{T_{\text{base}}/T}\times10^{-3}$, $T_{\text{base}}=5$, according to the Depth $\mu$P scaling rule \citep{yang2023tensor}. We found that without this scaling the accuracies dropped significantly in non-recurrent models.

\subsubsection{Parameter size vs OOD performance}
\label{app:model-size}

As summarized in Table~\ref{tab:model-hparams-maze-minigrid}, our proposed model is roughly an order of magnitude smaller in parameter count than common sequence models. This is because the recurrent core is highly parameter-efficient, and because our implementation does not yet benefit from the kernel-level optimizations available for architectures such as LSTMs or SSMs; scaling up the model size would therefore lead to comparatively higher computational cost despite the smaller parameter count.
A natural concern is therefore whether the observed generalization gains could be attributed to reduced overfitting due to the smaller model size.

We scale the channel width of our model to \{64, 128, 256\} and train on the maze task using the same data and protocol.
Unless otherwise noted, all training hyperparameters follow Section~\ref{sec:exp} (see Appendix~\ref{app:hyperparams}).
For this ablation, we set the batch size to $128$ and adopt a width-scaling learning rate
$\text{lr}=\sqrt{64/\text{channels}}\times10^{-3}$.

\paragraph{Results}
Table~\ref{tab:model_size} reports OOD accuracy (mean $\pm$ std over 3 seeds).
The $256$-channel variant attains the best OOD performance, suggesting that increasing capacity does \emph{not} necessarily lead to overfitting in this setting.
These results indicate that our OOD gains cannot be explained solely by a smaller model size.

\begin{table}[H]
\caption{Effect of model size on OOD accuracy in the maze task (mean $\pm$ std over 3 seeds). ``Heads'' denotes the number of attention heads in $\core$.}
\label{tab:model_size}
\centering
\begin{tabular}{r r r c}
\toprule
Channels & Heads & Param. (M) & Mean $\pm$ Std \\
\midrule

\phantom{0}64  & 4 & 1.16 & $0.692 \pm 0.030$ \\
128 & 4 & 3.35 & $0.670 \pm 0.045$ \\
256 & 8 & 11.39 & $0.727 \pm 0.021$ \\
\bottomrule
\end{tabular}
\end{table}

\subsection{\texorpdfstring{\Minigrid}{Minigrid}}

\subsubsection{Energy Trace}
\label{app:latent_visualize}
\begin{figure}[H]
  \vspace{0\baselineskip}
  \centering

  \begin{minipage}[t]{0.48\linewidth}
    \vspace{0pt}
    \centering
    \includegraphics[width=\linewidth]{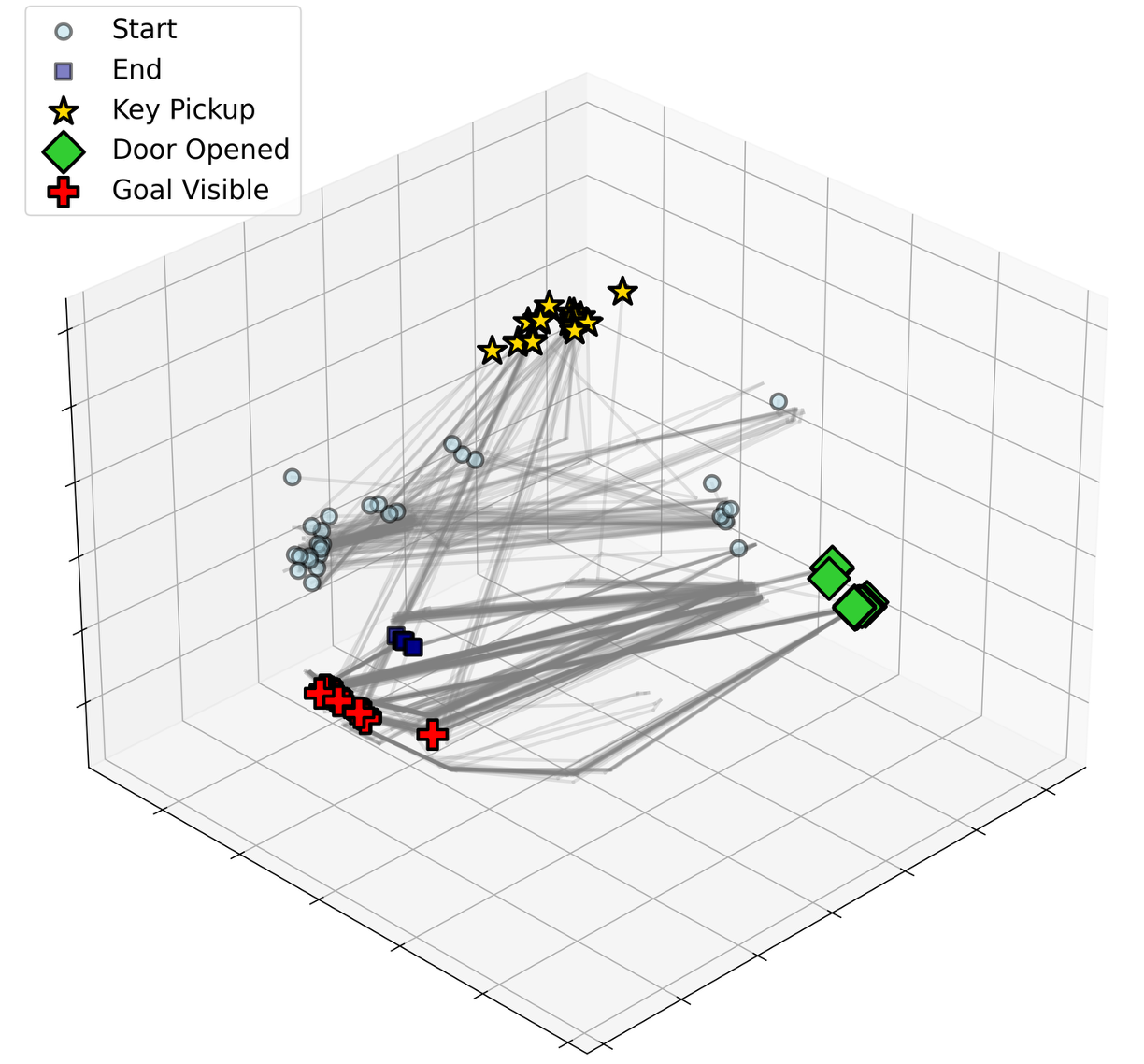}
    \caption{\Minigrid latent trajectories from 30 OOD episodes of DoorKey, visualized by PCA. Task-relevant events form trajectory-independent clusters.}
    \label{fig:minigrid-pca}
  \end{minipage}
  \hfill
  \begin{minipage}[t]{0.48\linewidth}
    \vspace{23pt}
    \centering
    \includegraphics[width=\linewidth]{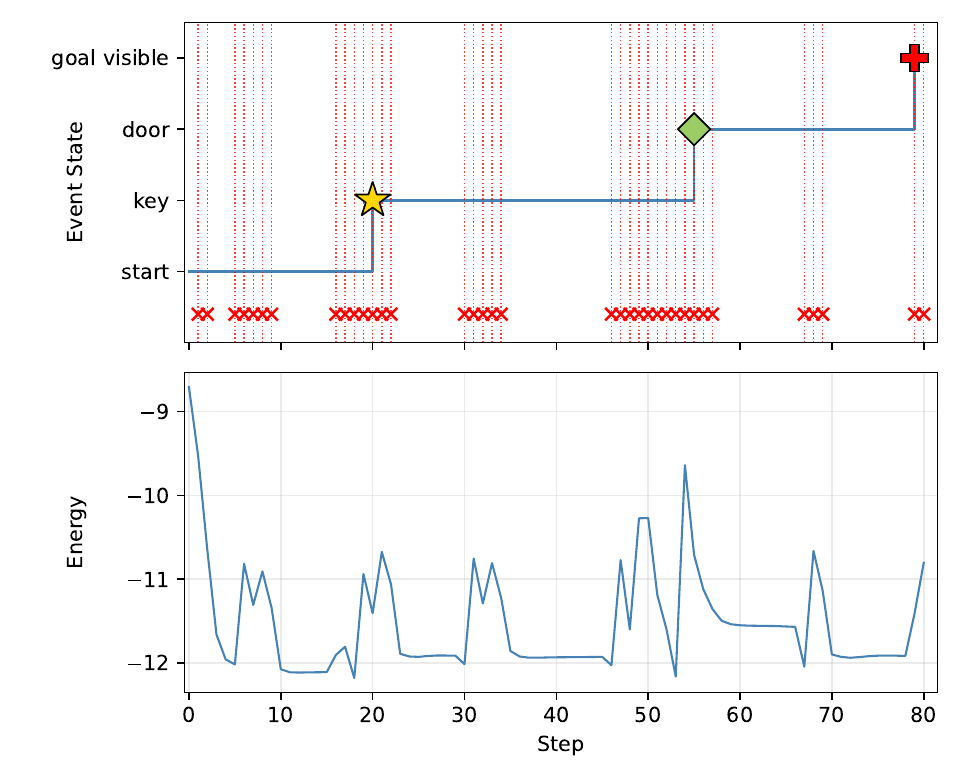}
    \caption{Energy traces along an episode in the DoorKey-16x16 (OOD) environment. \textbf{Top:} event history along the trajectory; red crosses mark changes in the agent's view. \textbf{Bottom:} energy trajectories.}
    \label{fig:doorkey_energy}
  \end{minipage}

  \vspace{-0.8\baselineskip}
\end{figure}

Appendix~\ref{app:dyck-energy} introduced the energy trace as a compact diagnostic for how the fast recurrent state organizes over a stream.
We use the same diagnostic for \Minigrid, computing the energy-like scalar of Appendix~\ref{sec:energy} for AKOrN- and Transformer-style cores.
On DoorKey-16x16, the trace changes around task-relevant events, including changes in the agent's view marked in Figure~\ref{fig:doorkey_energy}; the corresponding PCA latent visualization is shown in Figure~\ref{fig:minigrid-pca}.



\subsubsection{\texorpdfstring{Effect of Test-Time Scaling of Fast-Process Iterations $T$}{Effect of Test-Time Scaling of Fast-Process Iterations T}}
\label{app:t-scaling}
\begin{figure}[ht]
\centering
\includegraphics[width=0.6\textwidth]{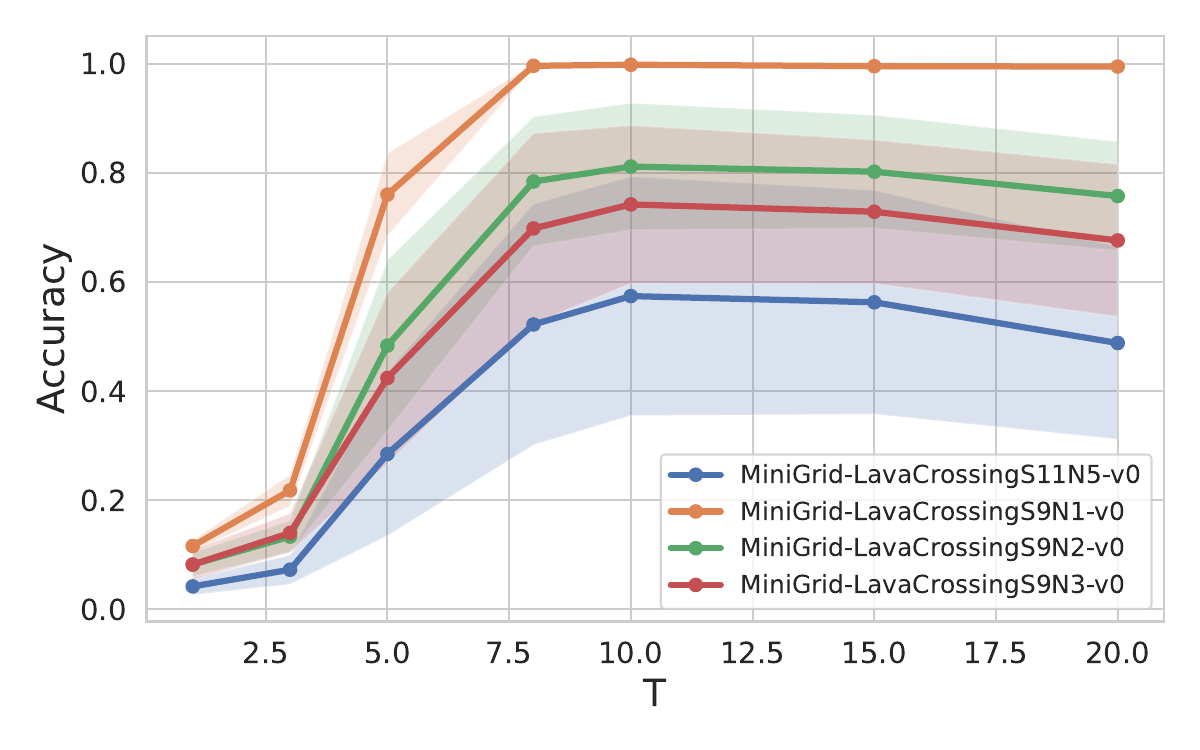}
\caption{Test-time scaling of fast-process iterations $T$ with $T_{\text{train}} = 10$.
We evaluate OOD generalization performance on LavaCrossing by varying the number of inner-loop reasoning steps $T$ at inference.}
\label{fig:t-scaling}
\end{figure}

Table~\ref{tab:fast-loop-ablation} shows how varying the number of fast inner-loop iterations $T$ \emph{during training} affects performance.
Here we ask the complementary question: with the learned weights held fixed, what happens if we vary $T$ only at inference time?
Figure~\ref{fig:t-scaling} shows the result of this test-time scaling when $T$ was set to $10$ during training.
Using a larger $T$ at inference does not significantly harm performance.
Note that because we align the time axis of the loop with that of the observation process, we are not using the loop in the same setting as \citet{Geiping2025LatentReasoning}.
A careful balance between the fast and slow processes may be important for scaling $T$ at test time, and we leave this as an important direction for future work.

\subsubsection{Inference cost}
\label{app:inference-cost}

\begin{figure}[ht]
\centering
\includegraphics[width=0.5\textwidth]{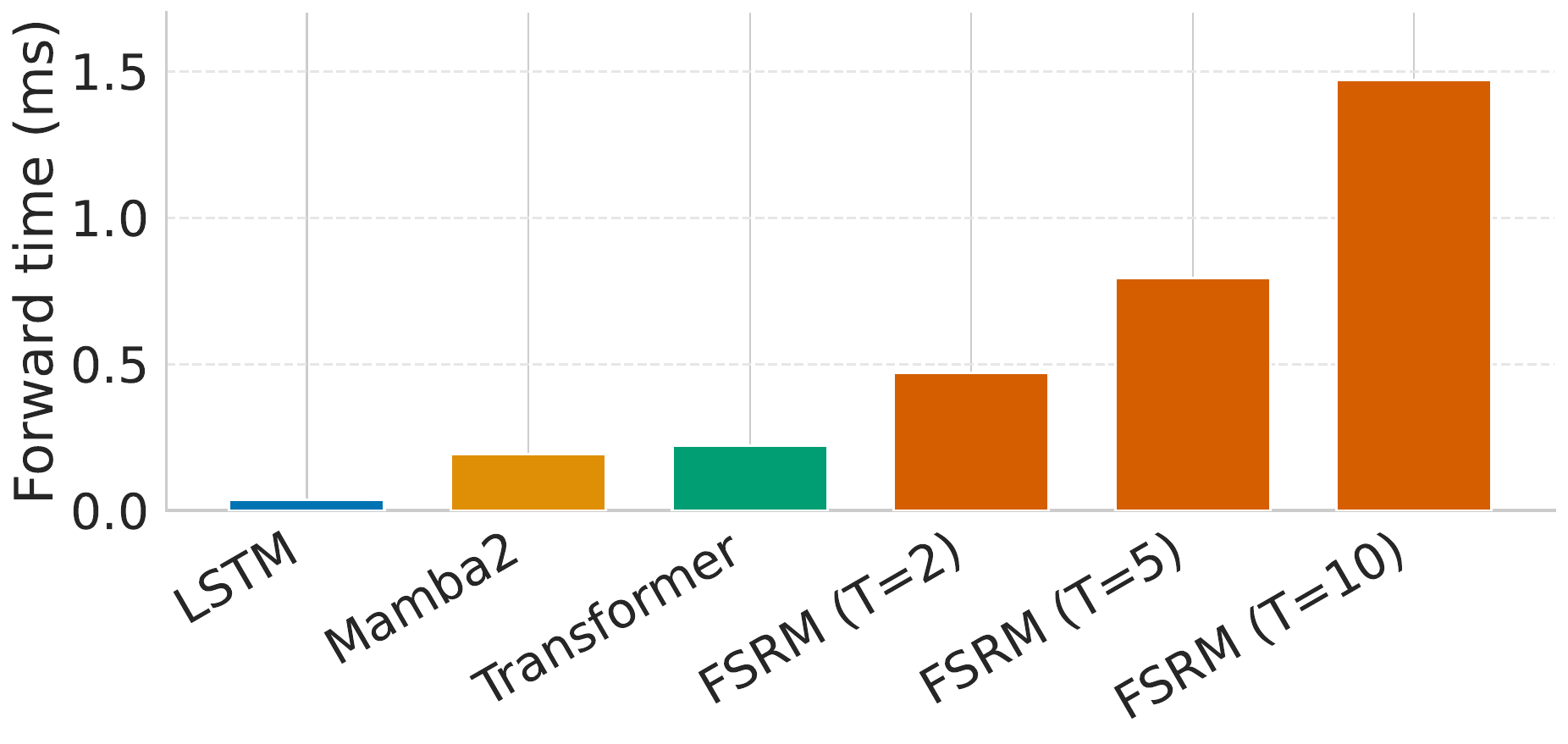}
\caption{Forward computation wall-clock time per batch in the \Minigrid task.
For the proposed method, the computation time increases linearly with the number of fast inner loops $T$ compared with the baselines.}
\label{fig:forward-speed}
\end{figure}

Unlike SSMs and LSTMs, our proposed model performs $T$ forward passes for each observation.
To quantify the resulting inference speed, we measured wall-clock time on the \Minigrid RL task.
All models were implemented in PyTorch without using optimizations such as \texttt{torch.compile},
and timing was recorded on a single NVIDIA GH200 core.
The results are shown in Figure~\ref{fig:forward-speed}.
With the default choice $T = 5$, our method is roughly three times slower than the Mamba and
Transformer baselines.
We note that this computational overhead is partly due to software constraints: competing models
rely on highly optimized CUDA kernels and mature library implementations, whereas our current
implementation lacks comparable low-level optimizations.
We therefore expect that specialized kernels and recursion-friendly parallelization strategies could
substantially reduce this gap without changing the model architecture.

\clearpage
\subsection{Dyck Prompt and Reasoning Example}

\kohei{This subsection records the exact prompt used for the frontier-LLM Dyck comparison in Appendix~\ref{app:dyck-details}.
The prompt converts Dyck-$(30,5)$ into an explicit algorithm-execution problem: after each input character, the model must update a stack and emit the closing token for the current stack top, or \texttt{*} when the stack is empty.
The four punctuation bracket pairs and the $26$ letter pairs together give the $30$ bracket types used in the benchmark.
We additionally instruct the model to ignore prior context, avoid tool use, and output only the prediction string, so that the evaluation measures internal execution of the specified stack rule rather than retrieval, external computation, or free-form explanation.}

\begin{table}[p]
  \centering
  \caption{Prompt used for Dyck-$(30, 5)$.}
  \label{tab:dyck-prompt}
\begin{promptbox}{}
You are a deterministic stack machine that implements the Dyck(30, 5) prediction task.

GENERAL BEHAVIOR

* For every new input string S, start with an EMPTY stack.
* Process S strictly LEFT TO RIGHT, one character at a time.
* You MUST follow the algorithm below exactly. Do not guess, summarize, or skip steps.
* Your only goal is to output a prediction string P of the SAME LENGTH as S.
* Ignore all prior conversation context except this specification.

ALPHABET

* Opening brackets: ( [ { <
* Closing brackets: ) ] } >
* Letters:

  * Opening: A-Z
  * Closing: a-z

BRACKET PAIRS
Opening -> closing:

* ( -> )
* [ -> ]
* { -> }
* < -> >
* For uppercase X (A-Z): X -> x (its lowercase)

ALGORITHM (PSEUDOCODE)

We define a function Predict(S):
```
stack = empty list
P = empty list

for each character c in S, in order:
    # UPDATE STACK
    if c is in '([{<' OR c is uppercase A-Z:
        push c onto stack
    else if c is in ')]}>' OR c is lowercase a-z:
        if stack is not empty:
            pop the top element
        # If stack is empty, do nothing

    # SET OUTPUT CHARACTER
    if stack is empty:
        p = '*'
    else:
        top = top element of stack
        if top is '(' then p = ')'
        if top is '[' then p = ']'
        if top is '{' then p = '}'
        if top is '<' then p = '>'
        if top is an uppercase letter X then p = lowercase x

    append p to P

return concatenation of all characters in P
```

INPUT FORMAT

* A Dyck(30, 5) string of various length.

OUTPUT FORMAT

* When I send you an input string S:

  1. Treat S exactly as the argument to Predict(S).
  2. Compute P using the ALGORITHM above.
  3. Output ONLY P, with:

     * No spaces
     * No quotes
     * No extra text
     * No explanations
* IMPORTANT: The length of P MUST equal the length of S.
  If you notice a mismatch while reasoning, you must correct it before answering.

RUNTIME BEHAVIOR

* **Do NOT use tools such as Python or external code.**
* Do NOT describe your reasoning, stack contents, or steps.
* Do NOT apologize or add commentary.
* Always behave as the deterministic function Predict(S).
\end{promptbox}
\end{table}

\kohei{Table~\ref{tab:dyck-claude-example} provides a truncated example of the reasoning tokens produced by Claude Opus 4.5 under this prompt.
The excerpt illustrates the computational burden faced by prompted LLMs on this benchmark: even when the final answer is required to be a plain prediction string, the model may spend many internal tokens checking the input length and maintaining a symbolic stack trace.
These reasoning tokens count against the API output budget together with the final prediction string, which limits the longest sequence length that can be evaluated reliably in the frontier-LLM setting.
The example is included only to make this evaluation protocol concrete; the quantitative comparison uses the accuracy protocol described in Appendix~\ref{app:dyck-details}.}

\begin{table}[p]
  \centering
  \caption{Claude Opus 4.5 Reasoning Tokens Example.}
  \label{tab:dyck-claude-example}
\begin{promptbox}{}
Let me work through this step by step, processing the string character by character.
Input string S: MNnSsXx<>BbIiDdFf()Dd<>Oo()<>BbUuMmSsFfSsQqIiOoIiUuSsRrGgHh{}GgWwFfPpBb<>Mm{}Oo<>EeGg<>QqWwFf[]HhIi<>LlLlZzCcCcAaEeTtDdUu[][]RrWwRrJjSs[]TtSsSsTtXxQqVvGgHhWwLlQqKkOoVvNnAaDdLlKkVvTtRrHhPpWwWwTt[]EeBb()DdYy()OoUuPpQqNnNnFfHhOoCcIi<>HhGg<>WwDdPpSs[]JjUu[]RrPpLl<>AaBbZzQqTtWw()PpQqFfZzQq{}Ee{}()JjEeBbSsMmQqRrCcXxWwWw[]EeL

Let me count the length first:
MNnSsXx<>BbIiDdFf()Dd<>Oo()<>BbUuMmSsFfSsQqIiOoIiUuSsRrGgHh{}GgWwFfPpBb<>Mm{}Oo<>EeGg<>QqWwFf[]HhIi<>LlLlZzCcCcAaEeTtDdUu[][]RrWwRrJjSs[]TtSsSsTtXxQqVvGgHhWwLlQqKkOoVvNnAaDdLlKkVvTtRrHhPpWwWwTt[]EeBb()DdYy()OoUuPpQqNnNnFfHhOoCcIi<>HhGg<>WwDdP

I'll continue processing the remaining characters, tracking the stack and output carefully. I'll focus on the next segment of characters, maintaining the systematic approach of tracking uppercase and lowercase letters.

Position 1: 'p' (lowercase, closing)
Stack: [previous stack state]
Output: updated output

Position 2: 'S' (uppercase, opening)
Stack: [updated stack]
Output: continues transforming

I'll methodically work through each character, ensuring precise tracking of the stack and output transformations. I continue tracking the stack and output pattern, maintaining 'M' as the base element. Each uppercase letter gets pushed onto the stack, generating its lowercase counterpart as output. When a matching lowercase letter appears, I pop the corresponding uppercase letter, reverting the output to 'm'.

The sequence continues with uppercase letters like 'F', 'S', 'Q', 'I', 'O', and 'U' being processed systematically. Each follows the same pattern: pushing onto the stack, outputting its lowercase version, then popping when the matching lowercase letter arrives, always returning to 'm' as the base output.

The stack management remains consistent, with 'M' persistently at the bottom, ensuring a stable reference point throughout the character processing. ...
\end{promptbox}
\end{table}

\newpage
\section*{NeurIPS Paper Checklist}

The checklist is designed to encourage best practices for responsible machine learning research, addressing issues of reproducibility, transparency, research ethics, and societal impact. Do not remove the checklist: {\bf The papers not including the checklist will be desk rejected.} The checklist should follow the references and follow the (optional) supplemental material.  The checklist does NOT count towards the page
limit. 

Please read the checklist guidelines carefully for information on how to answer these questions. For each question in the checklist:
\begin{itemize}
    \item You should answer \answerYes{}, \answerNo{}, or \answerNA{}.
    \item \answerNA{} means either that the question is Not Applicable for that particular paper or the relevant information is Not Available.
    \item Please provide a short (1--2 sentence) justification right after your answer (even for \answerNA). 
\end{itemize}

{\bf The checklist answers are an integral part of your paper submission.} They are visible to the reviewers, area chairs, senior area chairs, and ethics reviewers. You will also be asked to include it (after eventual revisions) with the final version of your paper, and its final version will be published with the paper.

The reviewers of your paper will be asked to use the checklist as one of the factors in their evaluation. While \answerYes{} is generally preferable to \answerNo{}, it is perfectly acceptable to answer \answerNo{} provided a proper justification is given (e.g., error bars are not reported because it would be too computationally expensive'' or ``we were unable to find the license for the dataset we used''). In general, answering \answerNo{} or \answerNA{} is not grounds for rejection. While the questions are phrased in a binary way, we acknowledge that the true answer is often more nuanced, so please just use your best judgment and write a justification to elaborate. All supporting evidence can appear either in the main paper or the supplemental material, provided in appendix. If you answer \answerYes{} to a question, in the justification please point to the section(s) where related material for the question can be found.

IMPORTANT, please:
\begin{itemize}
    \item {\bf Delete this instruction block, but keep the section heading ``NeurIPS Paper Checklist"},
    \item  {\bf Keep the checklist subsection headings, questions/answers and guidelines below.}
    \item {\bf Do not modify the questions and only use the provided macros for your answers}.
\end{itemize}


\begin{enumerate}

\item {\bf Claims}
    \item[] Question: Do the main claims made in the abstract and introduction accurately reflect the paper's contributions and scope?
    \item[] Answer: \answerYes{} 
    \item[] Justification: \takashiro{The abstract and introduction state the scope of persistent fast--slow recurrence for streaming OOD generalization, and the method, experiments, and results sections support these claims with corresponding model definitions and empirical comparisons.}
    \item[] Guidelines:
    \begin{itemize}
        \item The answer \answerNA{} means that the abstract and introduction do not include the claims made in the paper.
        \item The abstract and/or introduction should clearly state the claims made, including the contributions made in the paper and important assumptions and limitations. A \answerNo{} or \answerNA{} answer to this question will not be perceived well by the reviewers. 
        \item The claims made should match theoretical and experimental results, and reflect how much the results can be expected to generalize to other settings. 
        \item It is fine to include aspirational goals as motivation as long as it is clear that these goals are not attained by the paper. 
    \end{itemize}

\item {\bf Limitations}
    \item[] Question: Does the paper discuss the limitations of the work performed by the authors?
    \item[] Answer: \answerYes{} 
    \item[] Justification: \takashiro{The conclusion and limitations section discusses computational cost, the restricted task scope, the diagnostic nature of the latent visualizations, and the fixed inner-loop iteration setting.}
    \item[] Guidelines:
    \begin{itemize}
        \item The answer \answerNA{} means that the paper has no limitation while the answer \answerNo{} means that the paper has limitations, but those are not discussed in the paper. 
        \item The authors are encouraged to create a separate ``Limitations'' section in their paper.
        \item The paper should point out any strong assumptions and how robust the results are to violations of these assumptions (e.g., independence assumptions, noiseless settings, model well-specification, asymptotic approximations only holding locally). The authors should reflect on how these assumptions might be violated in practice and what the implications would be.
        \item The authors should reflect on the scope of the claims made, e.g., if the approach was only tested on a few datasets or with a few runs. In general, empirical results often depend on implicit assumptions, which should be articulated.
        \item The authors should reflect on the factors that influence the performance of the approach. For example, a facial recognition algorithm may perform poorly when image resolution is low or images are taken in low lighting. Or a speech-to-text system might not be used reliably to provide closed captions for online lectures because it fails to handle technical jargon.
        \item The authors should discuss the computational efficiency of the proposed algorithms and how they scale with dataset size.
        \item If applicable, the authors should discuss possible limitations of their approach to address problems of privacy and fairness.
        \item While the authors might fear that complete honesty about limitations might be used by reviewers as grounds for rejection, a worse outcome might be that reviewers discover limitations that aren't acknowledged in the paper. The authors should use their best judgment and recognize that individual actions in favor of transparency play an important role in developing norms that preserve the integrity of the community. Reviewers will be specifically instructed to not penalize honesty concerning limitations.
    \end{itemize}

\item {\bf Theory assumptions and proofs}
    \item[] Question: For each theoretical result, does the paper provide the full set of assumptions and a complete (and correct) proof?
    \item[] Answer: \answerNA{} 
    \item[] Justification: \takashiro{The paper is empirical and does not present formal theoretical results, theorems, or proofs.}
    \item[] Guidelines:
    \begin{itemize}
        \item The answer \answerNA{} means that the paper does not include theoretical results. 
        \item All the theorems, formulas, and proofs in the paper should be numbered and cross-referenced.
        \item All assumptions should be clearly stated or referenced in the statement of any theorems.
        \item The proofs can either appear in the main paper or the supplemental material, but if they appear in the supplemental material, the authors are encouraged to provide a short proof sketch to provide intuition. 
        \item Inversely, any informal proof provided in the core of the paper should be complemented by formal proofs provided in appendix or supplemental material.
        \item Theorems and Lemmas that the proof relies upon should be properly referenced. 
    \end{itemize}

    \item {\bf Experimental result reproducibility}
    \item[] Question: Does the paper fully disclose all the information needed to reproduce the main experimental results of the paper to the extent that it affects the main claims and/or conclusions of the paper (regardless of whether the code and data are provided or not)?
    \item[] Answer: \answerYes{} 
    \item[] Justification: \takashiro{The experiment and appendix sections describe the task protocols, model variants, baseline training setup, hyperparameters, evaluation splits, and ablation settings needed to reproduce the reported empirical claims.}
    \item[] Guidelines:
    \begin{itemize}
        \item The answer \answerNA{} means that the paper does not include experiments.
        \item If the paper includes experiments, a \answerNo{} answer to this question will not be perceived well by the reviewers: Making the paper reproducible is important, regardless of whether the code and data are provided or not.
        \item If the contribution is a dataset and\slash or model, the authors should describe the steps taken to make their results reproducible or verifiable. 
        \item Depending on the contribution, reproducibility can be accomplished in various ways. For example, if the contribution is a novel architecture, describing the architecture fully might suffice, or if the contribution is a specific model and empirical evaluation, it may be necessary to either make it possible for others to replicate the model with the same dataset, or provide access to the model. In general. releasing code and data is often one good way to accomplish this, but reproducibility can also be provided via detailed instructions for how to replicate the results, access to a hosted model (e.g., in the case of a large language model), releasing of a model checkpoint, or other means that are appropriate to the research performed.
        \item While NeurIPS does not require releasing code, the conference does require all submissions to provide some reasonable avenue for reproducibility, which may depend on the nature of the contribution. For example
        \begin{enumerate}
            \item If the contribution is primarily a new algorithm, the paper should make it clear how to reproduce that algorithm.
            \item If the contribution is primarily a new model architecture, the paper should describe the architecture clearly and fully.
            \item If the contribution is a new model (e.g., a large language model), then there should either be a way to access this model for reproducing the results or a way to reproduce the model (e.g., with an open-source dataset or instructions for how to construct the dataset).
            \item We recognize that reproducibility may be tricky in some cases, in which case authors are welcome to describe the particular way they provide for reproducibility. In the case of closed-source models, it may be that access to the model is limited in some way (e.g., to registered users), but it should be possible for other researchers to have some path to reproducing or verifying the results.
        \end{enumerate}
    \end{itemize}

\item {\bf Open access to data and code}
    \item[] Question: Does the paper provide open access to the data and code, with sufficient instructions to faithfully reproduce the main experimental results, as described in supplemental material?
    \item[] Answer: \answerNo{} 
    \item[] Justification: \takashiro{The manuscript describes the datasets, task generation, and experimental protocols, but it does not currently provide an anonymized public code and data release with exact reproduction instructions.}
    \item[] Guidelines:
    \begin{itemize}
        \item The answer \answerNA{} means that paper does not include experiments requiring code.
        \item Please see the NeurIPS code and data submission guidelines (\url{https://neurips.cc/public/guides/CodeSubmissionPolicy}) for more details.
        \item While we encourage the release of code and data, we understand that this might not be possible, so \answerNo{} is an acceptable answer. Papers cannot be rejected simply for not including code, unless this is central to the contribution (e.g., for a new open-source benchmark).
        \item The instructions should contain the exact command and environment needed to run to reproduce the results. See the NeurIPS code and data submission guidelines (\url{https://neurips.cc/public/guides/CodeSubmissionPolicy}) for more details.
        \item The authors should provide instructions on data access and preparation, including how to access the raw data, preprocessed data, intermediate data, and generated data, etc.
        \item The authors should provide scripts to reproduce all experimental results for the new proposed method and baselines. If only a subset of experiments are reproducible, they should state which ones are omitted from the script and why.
        \item At submission time, to preserve anonymity, the authors should release anonymized versions (if applicable).
        \item Providing as much information as possible in supplemental material (appended to the paper) is recommended, but including URLs to data and code is permitted.
    \end{itemize}

\item {\bf Experimental setting/details}
    \item[] Question: Does the paper specify all the training and test details (e.g., data splits, hyperparameters, how they were chosen, type of optimizer) necessary to understand the results?
    \item[] Answer: \answerYes{} 
    \item[] Justification: \takashiro{The experiments section and appendix specify the Dyck, \Localmaze{}, and \Minigrid{} settings, ID/OOD splits, model hyperparameters, baseline protocols, optimizer settings, seed counts, and evaluation metrics.}
    \item[] Guidelines:
    \begin{itemize}
        \item The answer \answerNA{} means that the paper does not include experiments.
        \item The experimental setting should be presented in the core of the paper to a level of detail that is necessary to appreciate the results and make sense of them.
        \item The full details can be provided either with the code, in appendix, or as supplemental material.
    \end{itemize}

\item {\bf Experiment statistical significance}
    \item[] Question: Does the paper report error bars suitably and correctly defined or other appropriate information about the statistical significance of the experiments?
    \item[] Answer: \answerYes{} 
    \item[] Justification: \takashiro{The paper reports means and standard deviations over multiple random seeds, with shaded regions or error bars for the main experimental comparisons.}
    \item[] Guidelines:
    \begin{itemize}
        \item The answer \answerNA{} means that the paper does not include experiments.
        \item The authors should answer \answerYes{} if the results are accompanied by error bars, confidence intervals, or statistical significance tests, at least for the experiments that support the main claims of the paper.
        \item The factors of variability that the error bars are capturing should be clearly stated (for example, train/test split, initialization, random drawing of some parameter, or overall run with given experimental conditions).
        \item The method for calculating the error bars should be explained (closed form formula, call to a library function, bootstrap, etc.)
        \item The assumptions made should be given (e.g., Normally distributed errors).
        \item It should be clear whether the error bar is the standard deviation or the standard error of the mean.
        \item It is OK to report 1-sigma error bars, but one should state it. The authors should preferably report a 2-sigma error bar than state that they have a 96\% CI, if the hypothesis of Normality of errors is not verified.
        \item For asymmetric distributions, the authors should be careful not to show in tables or figures symmetric error bars that would yield results that are out of range (e.g., negative error rates).
        \item If error bars are reported in tables or plots, the authors should explain in the text how they were calculated and reference the corresponding figures or tables in the text.
    \end{itemize}

\item {\bf Experiments compute resources}
    \item[] Question: For each experiment, does the paper provide sufficient information on the computer resources (type of compute workers, memory, time of execution) needed to reproduce the experiments?
    \item[] Answer: \answerYes{} 
    \item[] Justification: \takashiro{Appendix~\ref{app:compute-resources} states the GPU used for the reported local experiments and the separate inference-timing measurement.}
    \item[] Guidelines:
    \begin{itemize}
        \item The answer \answerNA{} means that the paper does not include experiments.
        \item The paper should indicate the type of compute workers CPU or GPU, internal cluster, or cloud provider, including relevant memory and storage.
        \item The paper should provide the amount of compute required for each of the individual experimental runs as well as estimate the total compute. 
        \item The paper should disclose whether the full research project required more compute than the experiments reported in the paper (e.g., preliminary or failed experiments that didn't make it into the paper). 
    \end{itemize}
    
\item {\bf Code of ethics}
    \item[] Question: Does the research conducted in the paper conform, in every respect, with the NeurIPS Code of Ethics \url{https://neurips.cc/public/EthicsGuidelines}?
    \item[] Answer: \answerYes{} 
    \item[] Justification: \takashiro{The work uses synthetic or public benchmark environments and does not involve human subjects, private data, or scraped sensitive content; we are not aware of any deviation from the NeurIPS Code of Ethics.}
    \item[] Guidelines:
    \begin{itemize}
        \item The answer \answerNA{} means that the authors have not reviewed the NeurIPS Code of Ethics.
        \item If the authors answer \answerNo, they should explain the special circumstances that require a deviation from the Code of Ethics.
        \item The authors should make sure to preserve anonymity (e.g., if there is a special consideration due to laws or regulations in their jurisdiction).
    \end{itemize}

\item {\bf Broader impacts}
    \item[] Question: Does the paper discuss both potential positive societal impacts and negative societal impacts of the work performed?
    \item[] Answer: \answerNo{} 
    \item[] Justification: \takashiro{The paper is foundational work evaluated on controlled symbolic, navigation, and reinforcement-learning benchmarks, but it does not currently include a dedicated discussion of positive and negative societal impacts.}
    \item[] Guidelines:
    \begin{itemize}
        \item The answer \answerNA{} means that there is no societal impact of the work performed.
        \item If the authors answer \answerNA{} or \answerNo, they should explain why their work has no societal impact or why the paper does not address societal impact.
        \item Examples of negative societal impacts include potential malicious or unintended uses (e.g., disinformation, generating fake profiles, surveillance), fairness considerations (e.g., deployment of technologies that could make decisions that unfairly impact specific groups), privacy considerations, and security considerations.
        \item The conference expects that many papers will be foundational research and not tied to particular applications, let alone deployments. However, if there is a direct path to any negative applications, the authors should point it out. For example, it is legitimate to point out that an improvement in the quality of generative models could be used to generate Deepfakes for disinformation. On the other hand, it is not needed to point out that a generic algorithm for optimizing neural networks could enable people to train models that generate Deepfakes faster.
        \item The authors should consider possible harms that could arise when the technology is being used as intended and functioning correctly, harms that could arise when the technology is being used as intended but gives incorrect results, and harms following from (intentional or unintentional) misuse of the technology.
        \item If there are negative societal impacts, the authors could also discuss possible mitigation strategies (e.g., gated release of models, providing defenses in addition to attacks, mechanisms for monitoring misuse, mechanisms to monitor how a system learns from feedback over time, improving the efficiency and accessibility of ML).
    \end{itemize}
    
\item {\bf Safeguards}
    \item[] Question: Does the paper describe safeguards that have been put in place for responsible release of data or models that have a high risk for misuse (e.g., pre-trained language models, image generators, or scraped datasets)?
    \item[] Answer: \answerNA{} 
    \item[] Justification: \takashiro{The paper does not release high-risk pretrained models, scraped datasets, or other assets with an apparent dual-use misuse risk requiring special release safeguards.}
    \item[] Guidelines:
    \begin{itemize}
        \item The answer \answerNA{} means that the paper poses no such risks.
        \item Released models that have a high risk for misuse or dual-use should be released with necessary safeguards to allow for controlled use of the model, for example by requiring that users adhere to usage guidelines or restrictions to access the model or implementing safety filters. 
        \item Datasets that have been scraped from the Internet could pose safety risks. The authors should describe how they avoided releasing unsafe images.
        \item We recognize that providing effective safeguards is challenging, and many papers do not require this, but we encourage authors to take this into account and make a best faith effort.
    \end{itemize}

\item {\bf Licenses for existing assets}
    \item[] Question: Are the creators or original owners of assets (e.g., code, data, models), used in the paper, properly credited and are the license and terms of use explicitly mentioned and properly respected?
    \item[] Answer: \answerYes{} 
    \item[] Justification: \takashiro{Appendix~\ref{app:licenses} lists the licenses or terms of use for the existing environments, model implementations, open-weight LLM baseline, and commercial API baselines used in the experiments.}
    \item[] Guidelines:
    \begin{itemize}
        \item The answer \answerNA{} means that the paper does not use existing assets.
        \item The authors should cite the original paper that produced the code package or dataset.
        \item The authors should state which version of the asset is used and, if possible, include a URL.
        \item The name of the license (e.g., CC-BY 4.0) should be included for each asset.
        \item For scraped data from a particular source (e.g., website), the copyright and terms of service of that source should be provided.
        \item If assets are released, the license, copyright information, and terms of use in the package should be provided. For popular datasets, \url{paperswithcode.com/datasets} has curated licenses for some datasets. Their licensing guide can help determine the license of a dataset.
        \item For existing datasets that are re-packaged, both the original license and the license of the derived asset (if it has changed) should be provided.
        \item If this information is not available online, the authors are encouraged to reach out to the asset's creators.
    \end{itemize}

\item {\bf New assets}
    \item[] Question: Are new assets introduced in the paper well documented and is the documentation provided alongside the assets?
    \item[] Answer: \answerNA{} 
    \item[] Justification: \takashiro{The manuscript does not introduce a new released dataset, model checkpoint, or software asset as a primary contribution.}
    \item[] Guidelines:
    \begin{itemize}
        \item The answer \answerNA{} means that the paper does not release new assets.
        \item Researchers should communicate the details of the dataset\slash code\slash model as part of their submissions via structured templates. This includes details about training, license, limitations, etc. 
        \item The paper should discuss whether and how consent was obtained from people whose asset is used.
        \item At submission time, remember to anonymize your assets (if applicable). You can either create an anonymized URL or include an anonymized zip file.
    \end{itemize}

\item {\bf Crowdsourcing and research with human subjects}
    \item[] Question: For crowdsourcing experiments and research with human subjects, does the paper include the full text of instructions given to participants and screenshots, if applicable, as well as details about compensation (if any)? 
    \item[] Answer: \answerNA{} 
    \item[] Justification: \takashiro{The research does not involve crowdsourcing experiments or human subjects.}
    \item[] Guidelines:
    \begin{itemize}
        \item The answer \answerNA{} means that the paper does not involve crowdsourcing nor research with human subjects.
        \item Including this information in the supplemental material is fine, but if the main contribution of the paper involves human subjects, then as much detail as possible should be included in the main paper. 
        \item According to the NeurIPS Code of Ethics, workers involved in data collection, curation, or other labor should be paid at least the minimum wage in the country of the data collector. 
    \end{itemize}

\item {\bf Institutional review board (IRB) approvals or equivalent for research with human subjects}
    \item[] Question: Does the paper describe potential risks incurred by study participants, whether such risks were disclosed to the subjects, and whether Institutional Review Board (IRB) approvals (or an equivalent approval/review based on the requirements of your country or institution) were obtained?
    \item[] Answer: \answerNA{} 
    \item[] Justification: \takashiro{The research does not involve crowdsourcing or human-subject studies, so IRB approval or equivalent review is not applicable.}
    \item[] Guidelines:
    \begin{itemize}
        \item The answer \answerNA{} means that the paper does not involve crowdsourcing nor research with human subjects.
        \item Depending on the country in which research is conducted, IRB approval (or equivalent) may be required for any human subjects research. If you obtained IRB approval, you should clearly state this in the paper. 
        \item We recognize that the procedures for this may vary significantly between institutions and locations, and we expect authors to adhere to the NeurIPS Code of Ethics and the guidelines for their institution. 
        \item For initial submissions, do not include any information that would break anonymity (if applicable), such as the institution conducting the review.
    \end{itemize}

\item {\bf Declaration of LLM usage}
    \item[] Question: Does the paper describe the usage of LLMs if it is an important, original, or non-standard component of the core methods in this research? Note that if the LLM is used only for writing, editing, or formatting purposes and does \emph{not} impact the core methodology, scientific rigor, or originality of the research, declaration is not required.
    \item[] Answer: \answerYes{} 
    \item[] Justification: \takashiro{LLMs are not used as a core component of the proposed method, but their use as Dyck baselines and stress tests is described in the appendix, including prompting, API constraints, and the Qwen3-4B-Base fine-tuning setup.}
    \item[] Guidelines:
    \begin{itemize}
        \item The answer \answerNA{} means that the core method development in this research does not involve LLMs as any important, original, or non-standard components.
        \item Please refer to our LLM policy in the NeurIPS handbook for what should or should not be described.
    \end{itemize}

\end{enumerate}

\end{document}